\documentclass[10pt,twocolumn,letterpaper]{article}

\usepackage{cvpr}              %
\usepackage[accsupp]{axessibility} %
\usepackage[dvipsnames]{xcolor}
\definecolor{cvprblue}{rgb}{0.21,0.49,0.74}
\usepackage[pagebackref,breaklinks,colorlinks,citecolor=cvprblue]{hyperref}
\usepackage{amsmath}
\usepackage{amssymb}
\usepackage{graphicx, fleqn, tabularx, wrapfig}
\usepackage{textcomp}
\usepackage{enumerate}
\usepackage{subcaption}
\usepackage{booktabs}
\usepackage{makecell}
\usepackage{multirow}
\usepackage{nicefrac,xfrac}
\usepackage{listings}
\usepackage{xcolor}
\usepackage{pgfplots}
\usepackage{stackengine}
\usepackage{xspace}
\usepackage{bbm}
\usepackage{bbding}
\usepackage{mathtools}
\usepackage{cancel}
\usepackage{tabularray}

\definecolor{codegreen}{rgb}{0,0.6,0}
\definecolor{codegray}{rgb}{0.5,0.5,0.5}
\definecolor{codepurple}{rgb}{0.58,0,0.82}
\definecolor{backcolour}{rgb}{0.95,0.95,0.92}

\lstdefinestyle{mystyle}{
    commentstyle=\color{codegreen},
    keywordstyle=\color{magenta},
    numberstyle=\tiny\color{codegray},
    stringstyle=\color{codepurple},
    basicstyle=\ttfamily\scriptsize,
    breakatwhitespace=false,
    breaklines=true,
    captionpos=b,
    keepspaces=true,
    numbersep=5pt,
    showspaces=false,
    showstringspaces=false,
    showtabs=false,
    tabsize=2,
    frame=single
}
\makeatletter
\renewcommand{\paragraph}{%
  \@startsection{paragraph}{4}%
  {\z@}{0.4ex \@plus 1ex \@minus .2ex}{-1em}%
  {\normalfont\normalsize\bfseries}%
}
\makeatother

\newcolumntype{L}[1]{>{\raggedright\let\newline\\\arraybackslash\hspace{0pt}}m{#1}}
\newcolumntype{C}[1]{>{\centering\let\newline\\\arraybackslash\hspace{0pt}}m{#1}}
\newcolumntype{R}[1]{>{\raggedleft\let\newline\\\arraybackslash\hspace{0pt}}m{#1}}
\newcolumntype{Y}{>{\centering\arraybackslash}X}

\definecolor{mypurple}{RGB}{223, 185, 226}
\definecolor{myblue}{RGB}{166, 189, 218}

\def\paperID{8642} %
\def\confName{CVPR}
\def\confYear{2024}

\title{EscherNet: A Generative Model for Scalable View Synthesis}

\author{
  Xin Kong$^1$\thanks{Corresponding Authors: {\tt \{x.kong21,shikun.liu17\}@imperial.ac.uk}. } \quad Shikun Liu$^{1}$\footnotemark[1]\quad  Xiaoyang Lyu$^2$ \quad Marwan Taher$^1$\\\medskip Xiaojuan Qi$^2$ \quad Andrew J. Davison$^1$ \\
 \normalsize $^{1}$Dyson Robotics Lab, Imperial College London \qquad 
 \normalsize $^{2}$The University of Hong Kong \\
 \small $^{\ast}$Corresponding Authors: {\tt \{x.kong21,shikun.liu17\}@imperial.ac.uk}
}

\begin{document}

\twocolumn[{%
    \renewcommand\twocolumn[1][]{#1}%
  \maketitle
  \vspace{-0.4cm}
  \includegraphics[width=\textwidth]{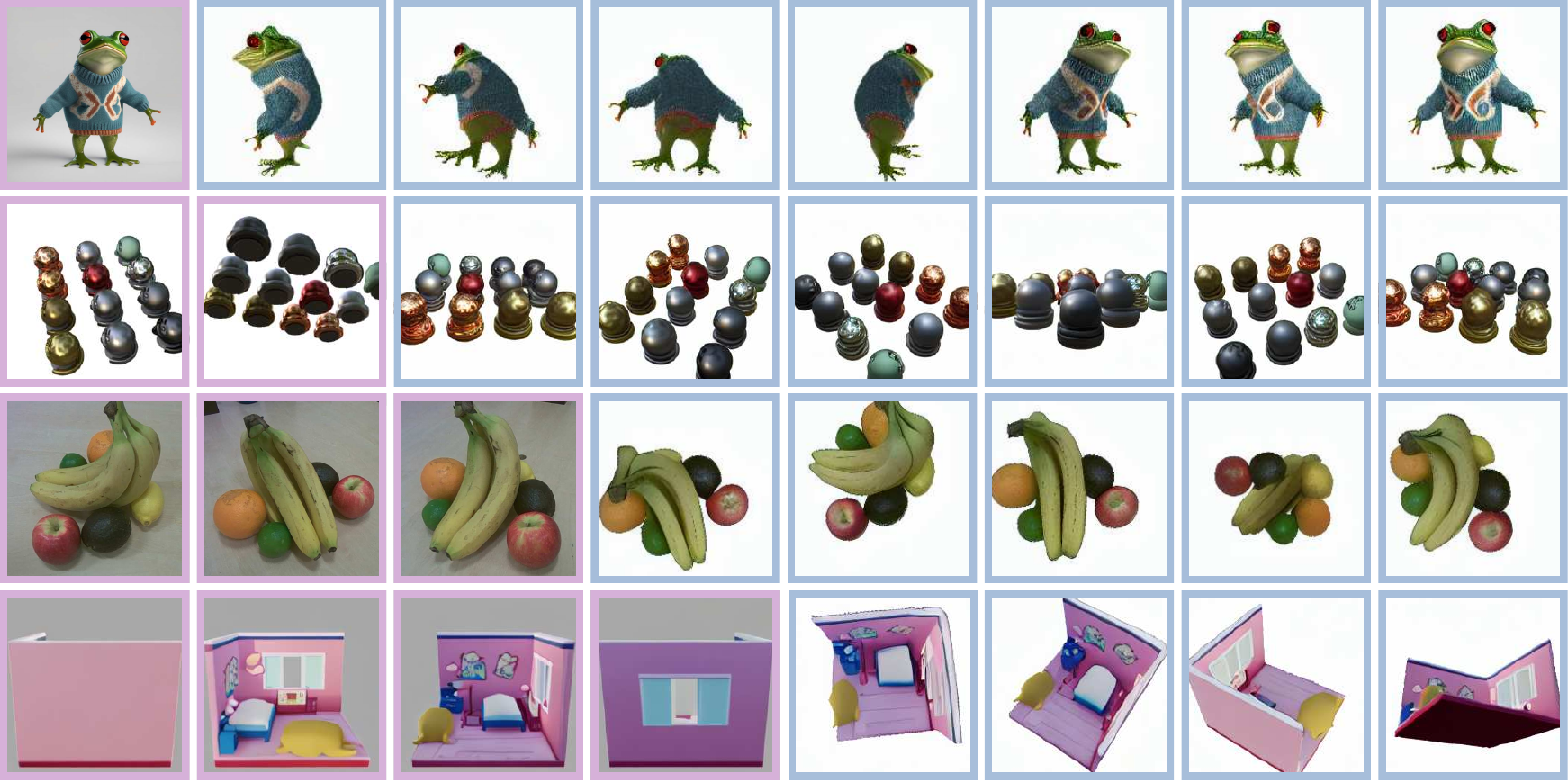}
  \captionof{figure}{We introduce EscherNet, a diffusion model that can generate a flexible number of consistent target views (highlighted in \textcolor{myblue}{\bf blue}) with arbitrary camera poses, based on a flexible number of reference views (highlighted in \textcolor{mypurple}{\bf purple}). EscherNet demonstrates remarkable precision in camera control and robust generalisation across synthetic and real-world images featuring multiple objects and rich textures.}
  \label{fig:teaser}
  \vspace{2em}
}]

\begin{abstract}
  We introduce EscherNet, a multi-view conditioned diffusion model for view synthesis. EscherNet learns implicit and generative 3D representations coupled with a specialised camera positional encoding, allowing precise and continuous relative control of the camera transformation between an arbitrary number of reference and target views. EscherNet offers exceptional generality, flexibility, and scalability in view synthesis --- it can generate more than 100 consistent target views simultaneously on a single consumer-grade GPU, despite being trained with a fixed number of 3 reference views to 3 target views. As a result, EscherNet not only addresses zero-shot novel view synthesis, but also naturally unifies single- and multi-image 3D reconstruction, combining these diverse tasks into a single, cohesive framework. Our extensive experiments demonstrate that EscherNet achieves state-of-the-art performance in multiple benchmarks, even when compared to methods specifically tailored for each individual problem. This remarkable versatility opens up new directions for designing scalable neural architectures for 3D vision. Project page: \url{https://kxhit.github.io/EscherNet}.
\end{abstract}
\vspace{-0.2cm}

\section{Introduction}
View synthesis stands as a fundamental task in computer vision and computer graphics. By allowing the re-rendering of a scene from arbitrary viewpoints based on a set of reference viewpoints, this mimics the adaptability observed in human vision. This ability is not only crucial for practical everyday tasks like object manipulation and navigation, but also plays a pivotal role in fostering human creativity, enabling us to envision and craft objects with depth, perspective, and a sense of immersion.

In this paper, we revisit the problem of view synthesis and ask: {\it How can we learn a general 3D representation to facilitate scalable view synthesis?} We attempt to investigate this question from the following two observations:

{\bf i)} Up until now, recent advances in view synthesis have predominantly focused on training speed and/or rendering efficiency~\cite{muller2022instant,kerbl20233gaussian,sun2022direct,garbin2021fastnerf}. Notably, these advancements all share a common reliance on volumetric rendering for scene optimisation. Thus, all these view synthesis methods are inherently {\it scene-specific}, coupled with global 3D spatial coordinates. In contrast, we advocate for a paradigm shift where a 3D representation relies solely on scene colours and geometries, learning implicit representations without the need for ground-truth 3D geometry, while also maintaining independence from any specific coordinate system. This distinction is crucial for achieving scalability to overcome the constraints imposed by scene-specific encoding.

{\bf ii)} View synthesis, by nature, is more suitable to be cast as a {\it conditional generative modelling problem}, similar to generative image in-painting~\cite{yu2018deepfill,lugmayr2022repaint}. When given only a sparse set of reference views, a desired model should provide multiple plausible predictions, leveraging the inherent stochasticity within the generative formulation and drawing insights from natural image statistics and semantic priors learned from other images and objects. As the available information increases, the generated scene becomes more constrained, gradually converging closer to the ground-truth representation. Notably, existing 3D generative models currently only support a single reference view~\cite{liu2023zero,shi2023zero123++,liu2023syncdreamer,long2023wonder3d,liu2023one}. We argue that a more desirable generative formulation should flexibly accommodate varying levels of input information.

Building upon these insights, we introduce EscherNet, an image-to-image conditional diffusion model for view synthesis. EscherNet leverages a transformer architecture~\cite{vaswani2017transformer}, employing dot-product self-attention to capture the intricate relation between both reference-to-target and target-to-target views consistencies. A key innovation within EscherNet is the design of camera positional encoding (CaPE), dedicated to representing both 4 DoF (object-centric) and 6 DoF camera poses. This encoding incorporates spatial structures into the tokens, enabling the model to compute self-attention between query and key solely based on their relative camera transformation. In summary, EscherNet exhibits these remarkable characteristics:
\begin{itemize}
    \item {\bf Consistency}: EscherNet inherently integrates view consistency thanks to the design of camera positional encoding, encouraging both {\it reference-to-target and target-to-target view consistencies}.
      \item {\bf Scalability}: Unlike many existing neural rendering methods that are constrained by scene-specific optimisation, EscherNet decouples itself from any specific coordinate system and the need for ground-truth 3D geometry, without any expensive 3D operations ({\it e.g.} 3D convolutions or volumetric rendering), making it easier to {\it scale with everyday posed 2D image data}.
    \item {\bf Generalisation}: Despite being trained on only a fixed number of 3 reference to 3 target views, EscherNet exhibits the capability to generate {\it any number of target views, with any camera poses, based on any number of reference views}. Notably, EscherNet exhibits improved generation quality with an increased number of reference views, aligning seamlessly with our original design goal.
\end{itemize}

We conduct a comprehensive evaluation across both novel view synthesis and single/multi-image 3D reconstruction benchmarks. Our findings demonstrate that EscherNet not only outperforms all 3D diffusion models in terms of generation quality but also can generate plausible view synthesis given very limited views. This stands in contrast to these scene-specific neural rendering methods such as InstantNGP~\cite{muller2022instant} and Gaussian Splatting~\cite{kerbl20233gaussian}, which often struggle to generate meaningful content under such constraints. This underscores the effectiveness of our method's simple yet scalable design, offering a promising avenue for advancing view synthesis and 3D vision as a whole.

\section{Related Work}

\paragraph{Neural 3D Representations} 
Early works in neural 3D representation learning focused on directly optimising on 3D data, using representations such as voxels~\cite{maturana2015voxnet} and point clouds~\cite{qi2017pointnet, qi2017pointnet++}, for explicit 3D representation learning. Alternatively, another line of works focused on training neural networks to map 3D spatial coordinates to signed distance functions~\cite{park2019deepsdf} or occupancies~\cite{mescheder2019occupancy, Peng2020ECCV}, for implicit 3D representation learning. However, all these methods heavily rely on ground-truth 3D geometry, limiting their applicability to small-scale synthetic 3D data~\cite{wu20153dshapenet,chang2015shapenet}.

To accommodate a broader range of data sources, differentiable rendering functions~\cite{niemeyer2020differentiable, sitzmann2019scene} have been introduced to optimise neural implicit shape representations with multi-view posed images. More recently, NeRF~\cite{mildenhall2020nerf} paved the way to a significant enhancement in rendering quality compared to these methods by optimising MLPs to encode 5D radiance fields. In contrast to tightly coupling 3D scenes with spatial coordinates, we introduce EscherNet as an alternative for 3D representation learning by optimising a neural network to learn the interaction between multi-view posed images, independent of any coordinate system.

\paragraph{Novel View Synthesis}
The success of NeRF has sparked a wave of follow-up methods that address faster training and/or rendering efficiency, by incorporating different variants of space discretisation~\cite{garbin2021fastnerf,hedman2021baking,chen2022tensorf},  codebooks~\cite{takikawa2022variable}, and encodings using hash tables~\cite{muller2022instant} or Gaussians~\cite{kerbl20233gaussian}.

To enhance NeRF's generalisation ability across diverse scenes and in a few-shot setting, PixelNeRF~\cite{yu2021pixelnerf} attempts to learn a scene prior by jointly optimising multiple scenes, but it is constrained by the high computational demands required by volumetric rendering. Various other approaches have addressed this issue by introducing regularisation techniques, such as incorporating low-level priors from local patches~\cite{niemeyer2022regnerf}, ensuring semantic consistency~\cite{jain2021dietnerf}, considering adjacent ray frequency~\cite{yang2023freenerf}, and incorporating depth signals~\cite{deng2022depth}. In contrast, EscherNet encodes scenes directly through the image space, enabling the learning of more generalised scene priors through large-scale datasets.

\begin{figure}[t!]
  \centering
  \begin{subfigure}[b]{0.31\linewidth}
    \centering
    \includegraphics[trim={12cm 0cm 12cm 5cm}, clip, width=\linewidth]{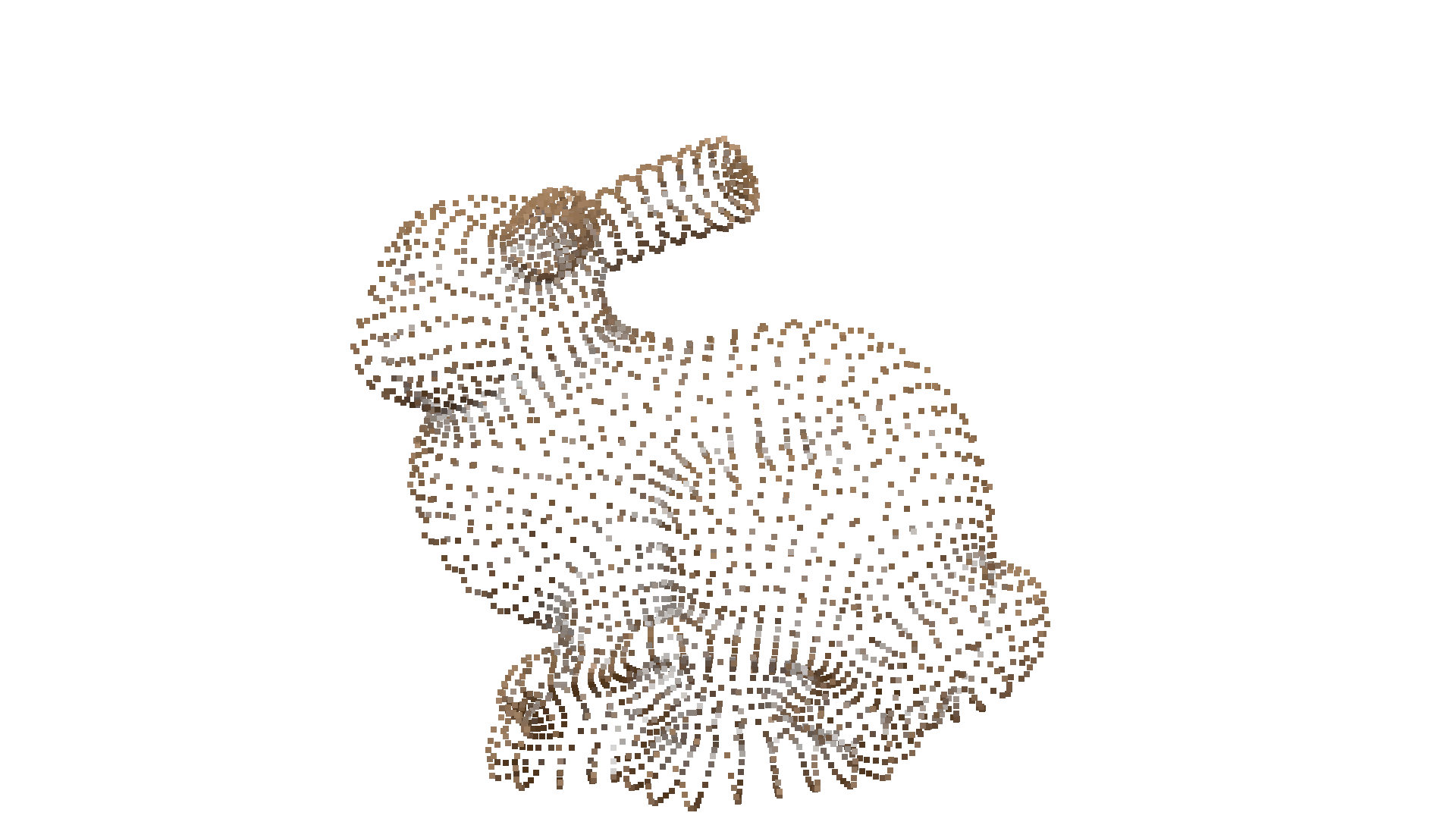}
    \caption*{Points}
  \end{subfigure}\hfill
  \begin{subfigure}[b]{0.31\linewidth}
    \centering
    \includegraphics[trim={12cm 0cm 12cm 5cm}, clip, width=\linewidth]{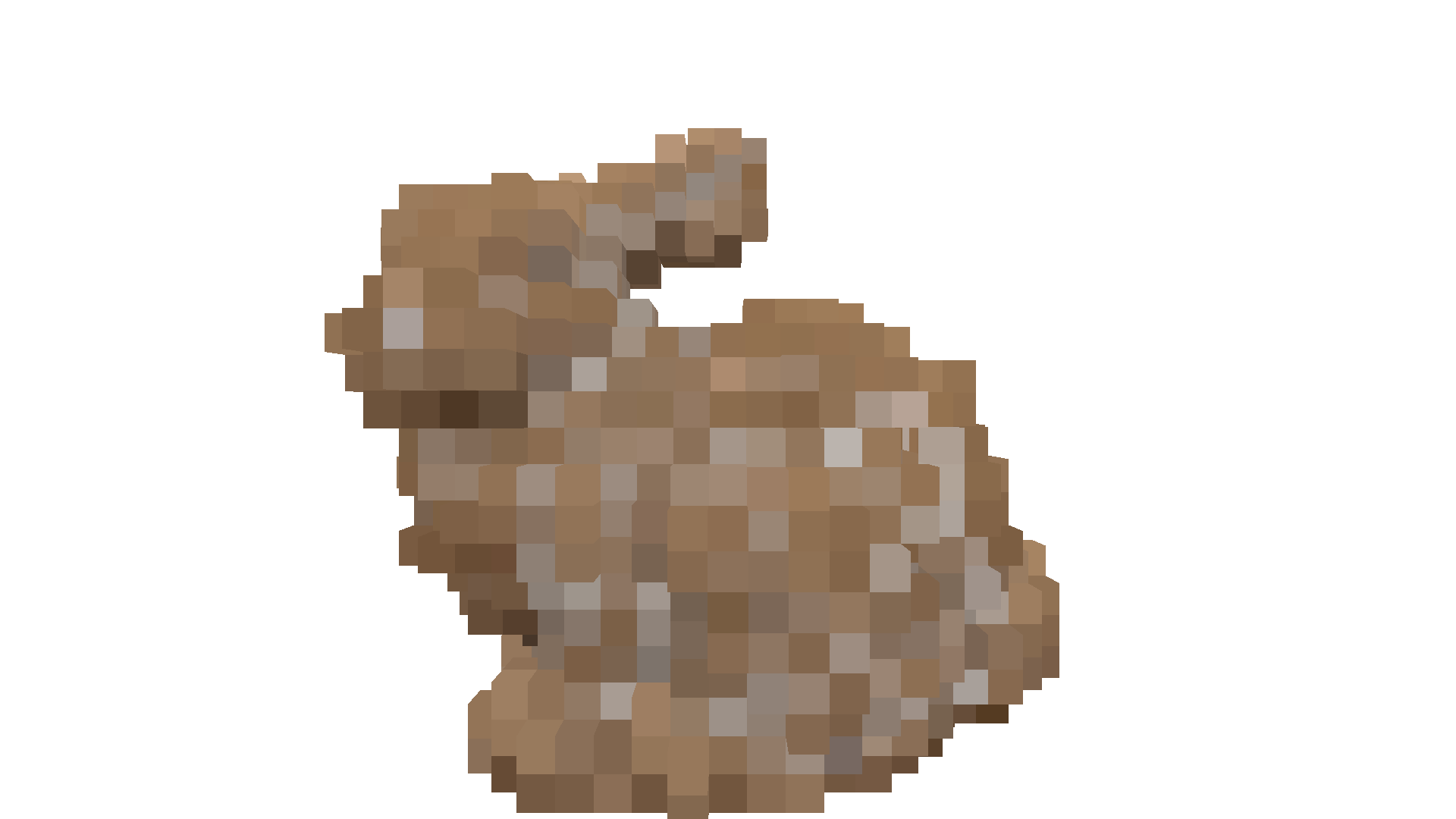}
    \caption*{Voxels}
  \end{subfigure}\hfill
  \begin{subfigure}[b]{0.31\linewidth}
    \centering
    \includegraphics[width=0.82\linewidth]{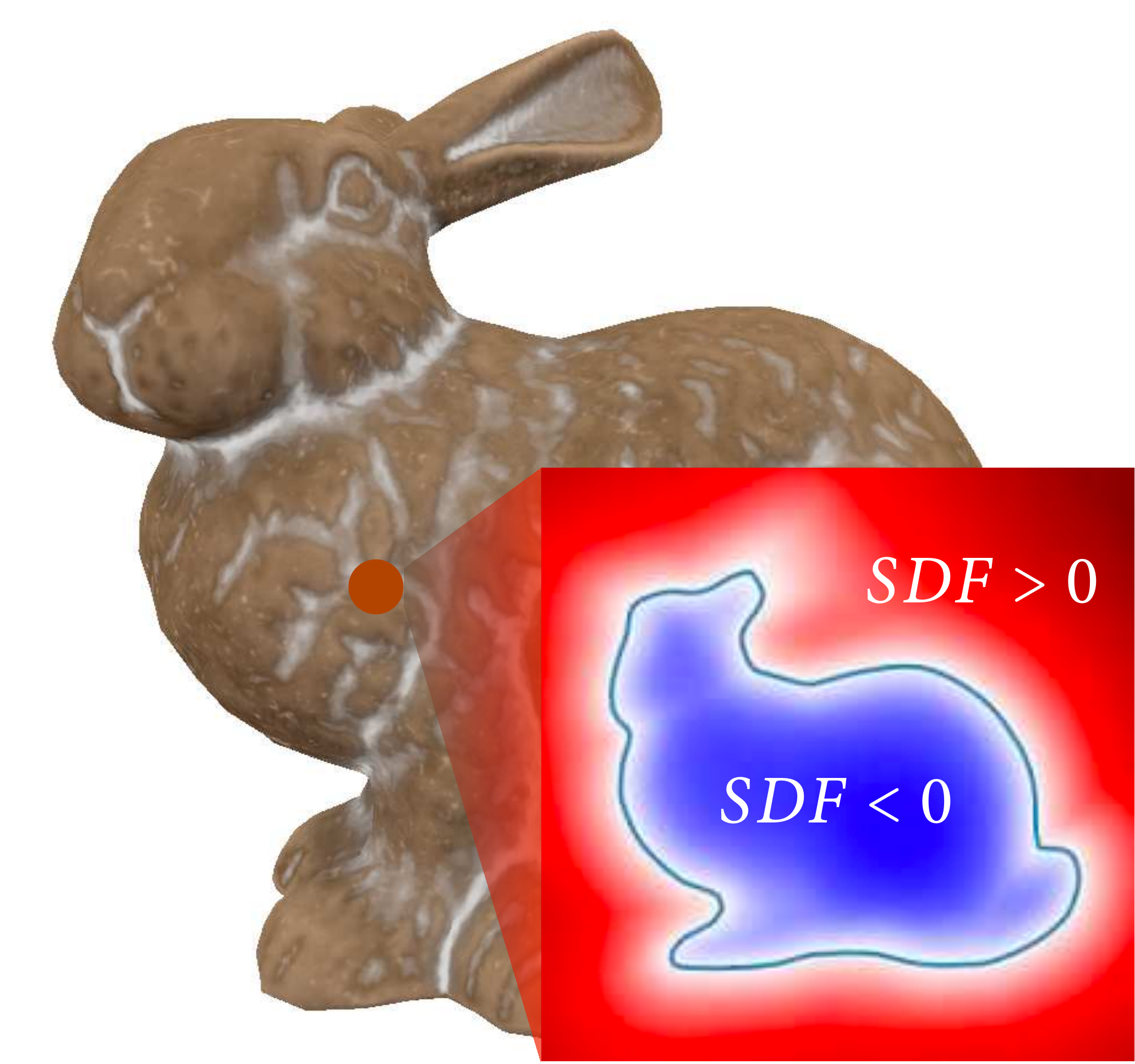}
    \caption*{SDF}
  \end{subfigure}\\
  \begin{subfigure}[b]{0.33\linewidth}
    \centering
    \includegraphics[height=0.95\linewidth]{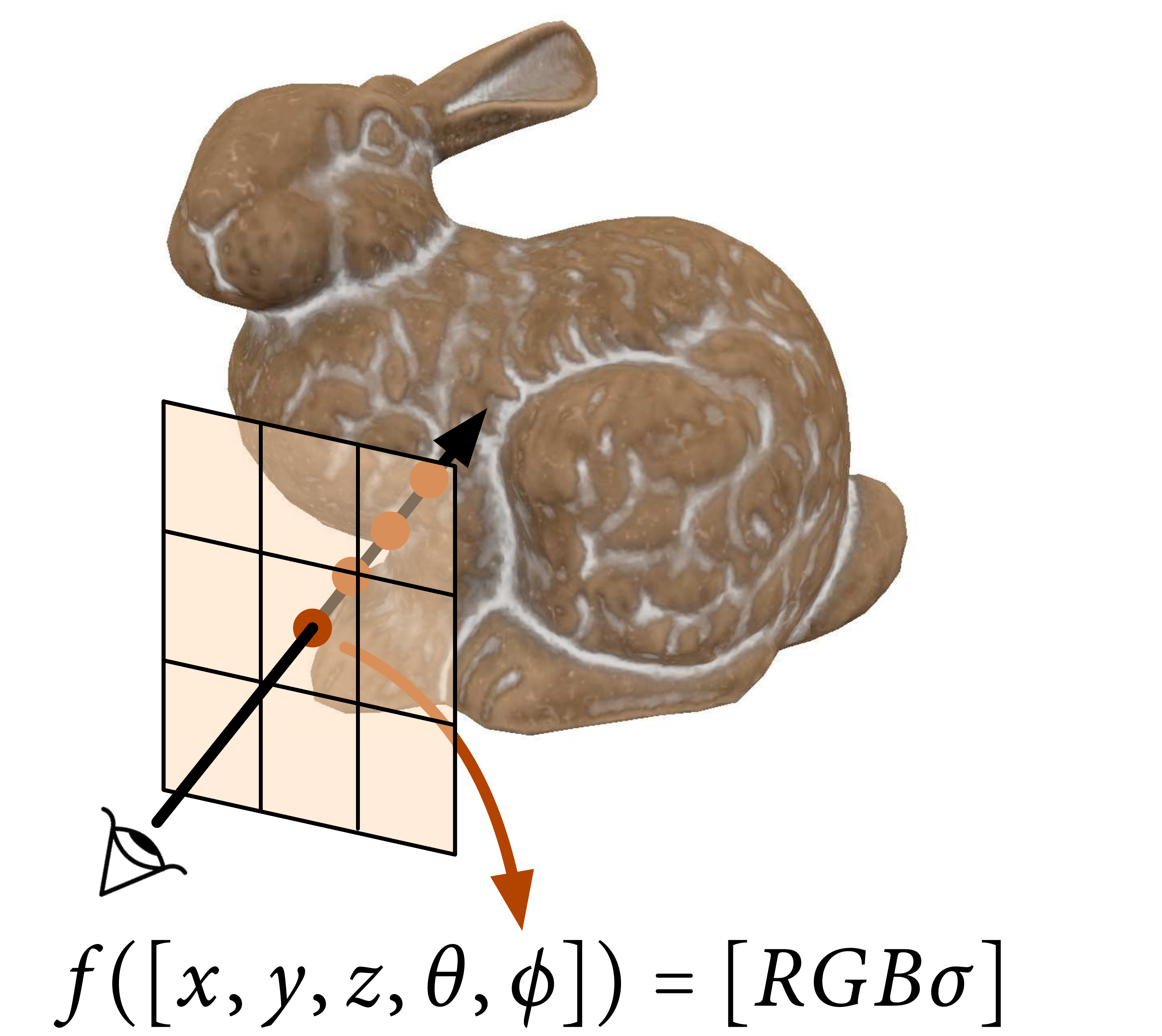}
    \caption*{NeRF}
  \end{subfigure}\hfill
    \begin{subfigure}[b]{0.33\linewidth}
    \centering
    \includegraphics[height=0.95\linewidth]{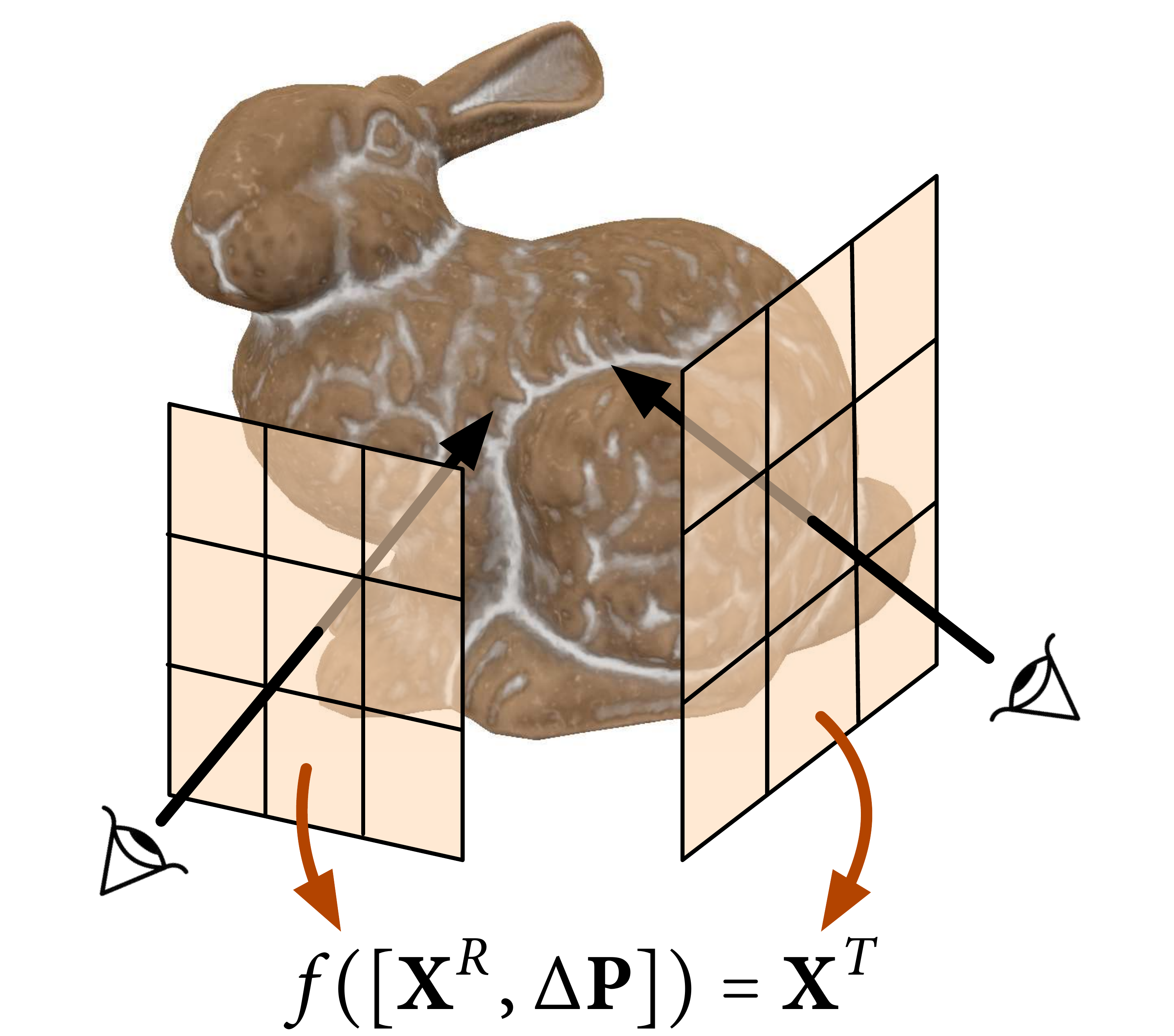}
    \caption*{Zero-1-to-3}
  \end{subfigure}\hfill
  \begin{subfigure}[b]{0.33\linewidth}
    \centering
    \includegraphics[height=0.95\linewidth]{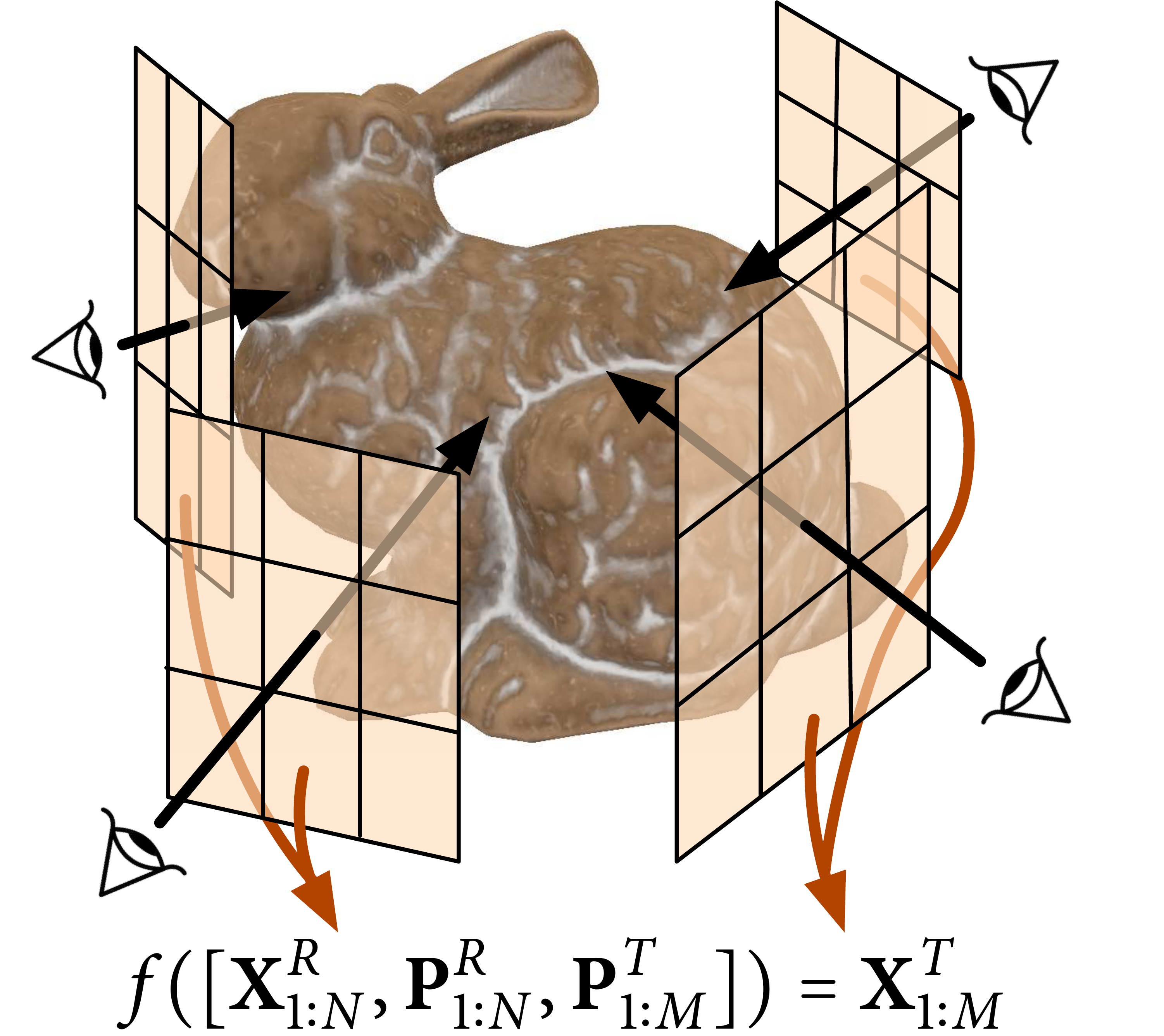}
    \caption*{EscherNet}
  \end{subfigure}
  \caption{{\bf 3D representations overview.} EscherNet generates a set of $M$ target views ${\bf X}_{1:M}^T$ based on their camera poses ${\bf P}_{1:M}^T$, leveraging information gained from a set of $N$ reference views ${\bf X}_{1:N}^R$ and their camera poses ${\bf P}_{1:N}^R$. EscherNet presents a new way of learning implicit 3D representations by only considering the relative camera transformation between the camera poses of ${\bf P}^R$ and ${\bf P}^T$, making it easier to scale with multi-view posed images, independent of any specific coordinate systems. }
  \label{fig:intro}
  \vspace{-0.4cm}
\end{figure}

\paragraph{3D Diffusion Models} The emergence of 2D generative diffusion models has shown impressive capabilities in generating realistic objects and scenes~\cite{ho2020ddpm, rombach2022ldm}. This progress has inspired the early design of text-to-3D diffusion models, such as DreamFusion~\cite{poole2022dreamfusion} and Magic3D~\cite{lin2023magic3d}, by optimising a radiance field guided by score distillation sampling (SDS) from these pre-trained 2D diffusion models. However, SDS necessitates computationally intensive iterative optimisation, often requiring up to an hour for convergence. Additionally, these methods, including recently proposed image-to-3D generation approaches~\cite{deng2023nerdi, xu2023neurallift, melas2023realfusion}, frequently yield unrealistic 3D generation results due to their limited 3D understanding, giving rise to challenges such as the multi-face Janus problem.

To integrate 3D priors more efficiently, an alternative approach involves training 3D generative models directly on 3D datasets, employing representations like point clouds~\cite{nichol2022pointe} or neural fields~\cite{erkocc2023hyperdiffusion, chen2023single, jun2023shape}. However, this design depends on 3D operations, such as 3D convolution and volumetric rendering, which are computationally expensive and challenging to scale.

To address this issue, diffusion models trained on multi-view posed data have emerged as a promising direction, designed with no 3D operations. Zero-1-to-3~\cite{liu2023zero} stands out as a pioneering work, learning view synthesis from paired 2D posed images rendered from large-scale 3D object datasets~\cite{deitke2023objaverse,deitke2023objaversexl}. However, its capability is limited to generating a single target view conditioned on a single reference view. Recent advancements in multi-view diffusion models~\cite{shi2023mvdream, ye2023consistent123, shi2023zero123++, liu2023syncdreamer, long2023wonder3d, liu2023one} focused on 3D generation and can only generate a fixed number of target views with fixed camera poses. In contrast, EscherNet can generate an unrestricted number of target views with arbitrary camera poses, offering superior flexibility in view synthesis.

\section{EscherNet}
\paragraph{Problem Formulation and Notation}
In EscherNet, we recast the view synthesis as a conditional generative modelling problem, formulated as: 
\begin{align}
    \mathcal{X}^T \sim p(\mathcal{X}^T| \mathcal{X}^R, \mathcal{P}^R, \mathcal{P}^T).
\end{align}
Here, $\mathcal{X}^T=\{{\bf X}_{1:M}^T\}$ and $\mathcal{P}^T=\{{\bf P}^T_{1:M}\}$ represent a set of $M$ target views ${\bf X}_{1:M}^T$ with their global camera poses ${\bf P}_{1:M}^T$. Similarly, $\mathcal{X}^R=\{{\bf X}_{1:N}^R\}$ and $\mathcal{P}^R=\{{\bf P}_{1:N}^R\}$ represent a set of $N$ reference views ${\bf X}_{1:N}^R$ with their global camera poses ${\bf P}_{1:N}^R$. Both $N$ and $M$ can take on arbitrary values during both model training and inference. 

We propose a neural architecture design, such that the generation of each target view ${\bf X}^T_i\in \mathcal{X}^T$ solely depends on its relative camera transformation to the reference views $({\bf P}^R_j)^{-1}{\bf P}^T_i,\forall {\bf P}^R_j\in \mathcal{P}^R$, introduced next.

\begin{figure*}[ht!]
  \centering
  \includegraphics*[width=\textwidth]{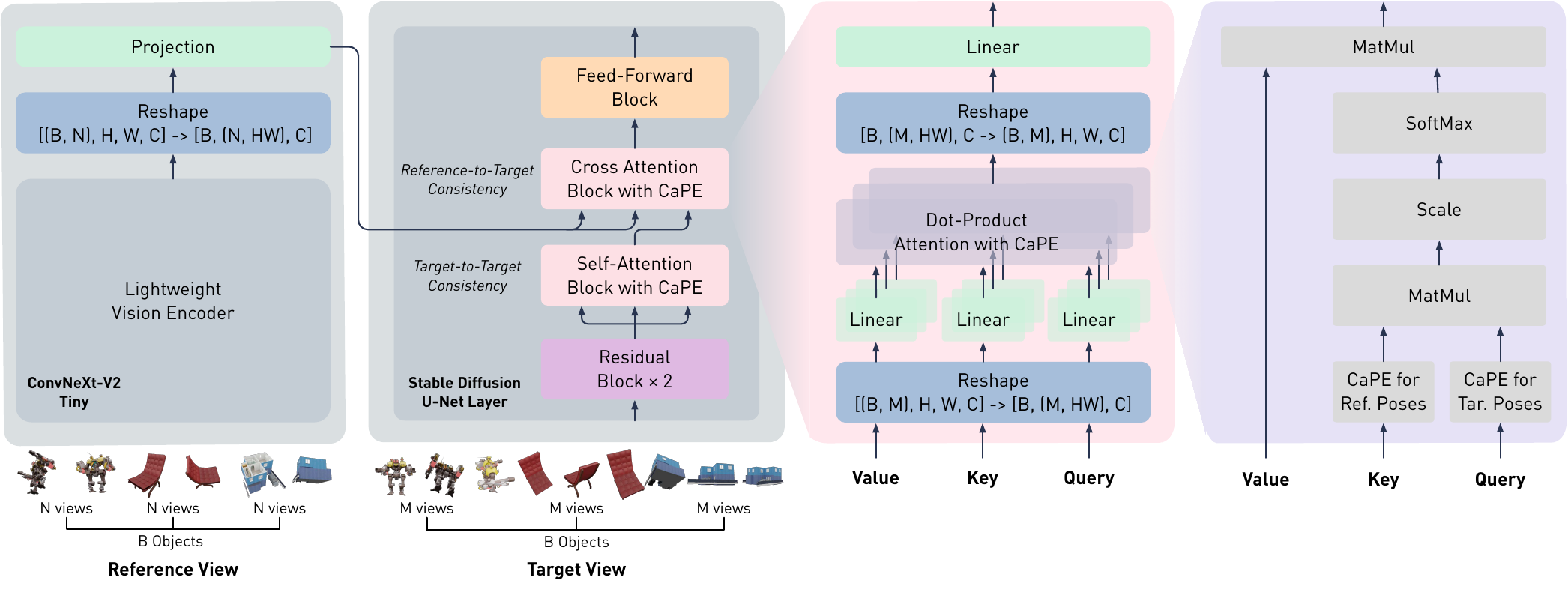}
  \caption{{\bf EscherNet architecture details.} 
  EscherNet adopts the Stable Diffusion architectural design with minimal but important modifications. The lightweight vision encoder captures both high-level and low-level signals from $N$ reference views. In U-Net, we apply self-attention within $M$ target views to encourage target-to-target consistency, and cross-attention within $M$ target and $N$ reference views (encoded by the image encoder) to encourage reference-to-target consistency. In each attention block, CaPE is employed for the key and query, allowing the attention map to learn with relative camera poses, independent of specific coordinate systems. }
  \label{fig:eschernet}
  \vspace{-0.4cm}
\end{figure*}

\subsection{Architecture Design} 
We design EscherNet following two key principles: i) It builds upon an existing 2D diffusion model, inheriting its strong web-scale prior through large-scale training, and ii) It encodes camera poses for each view/image, similar to how language models encode token positions for each token. So our model can naturally handle an arbitrary number of views for {\it any-to-any view synthesis}. 

\paragraph{Multi-View Generation} EscherNet can be seamlessly integrated with any 2D diffusion model with a transformer architecture, with {\it no additional learnable parameters}. In this work, we design EscherNet by adopting a latent diffusion architecture, specifically {\tt StableDiffusion v1.5}~\cite{rombach2022ldm}. This choice enables straightforward comparisons with numerous 3D diffusion models that also leverage the same backbone (more details in the experiment section). 

To tailor the Stable Diffusion model, originally designed for text-to-image generation, to multi-view generation as applied in EscherNet, several key modifications are implemented. In the original Stable Diffusion's denoiser U-Net, the self-attention block was employed to learn interactions within different patches within the same image. In EscherNet, we re-purpose this self-attention block to facilitate learning interactions within distinct patches across $M$ different target views, thereby ensuring target-to-target consistency. Likewise, the cross-attention block, originally used to integrate textual information into image patches, is repurposed in EscherNet to learn interactions within $N$ reference to $M$ target views, ensuring reference-to-target consistency. 

\paragraph{Conditioning Reference Views} In view synthesis, it is crucial that the conditioning signals accurately capture both the high-level semantics and low-level texture details present in the reference views. Previous works in 3D diffusion models~\cite{liu2023zero,liu2023syncdreamer} have employed the strategy of encoding high-level signals through a frozen CLIP pre-trained ViT~\cite{radford2021clip} and encoding low-level signals by concatenating the reference image into the input of the U-Net of Stable Diffusion. However, this design choice inherently constrains the model to handle only one single view.

In EscherNet, we choose to incorporate both high-level and low-level signals in the conditioning image encoder, representing reference views as sets of tokens. This design choice allows our model to maintain flexibility in handling a variable number of reference views. Early experiments have confirmed that using a frozen CLIP-ViT alone may fail to capture low-level textures, preventing the model from accurately reproducing the original reference views given the same reference view poses as target poses. While fine-tuning the CLIP-ViT could address this issue, it poses challenges in terms of training efficiency. Instead, we opt to fine-tune a lightweight vision encoder, specifically {\tt ConvNeXtv2-Tiny}~\cite{woo2023convnextv2}, which is a highly efficient CNN architecture. This architecture is employed to compress our reference views to smaller resolution image features. We treat these image features as conditioning tokens, effectively representing each reference view. This configuration has proven to be sufficient in our experiments, delivering superior results in generation quality while simultaneously maintaining high training efficiency.

\subsection{Camera Positional Encoding (CaPE)}

To encode camera poses efficiently and accurately into reference and target view tokens within a transformer architecture, we introduce Camera Positional Encoding (CaPE), drawing inspiration from recent advancements in the language domain.  We first briefly examine the distinctions between these two domains.

-- In language, token positions (associated with each word) follow a {\it linear and discrete} structure, and their length can be {\it infinite}. Language models are typically trained with fixed maximum token counts (known as context length), and it remains an ongoing research challenge to construct a positional encoding that enables the model to behave reasonably beyond this fixed context length~\cite{peng2023yarn,han2023lm}.

-- In 3D vision, token positions (associated with each camera) follow a {\it cyclic, continuous, and bounded} structure for rotations and a {\it linear, continuous, and unbounded} structure for translations. Importantly, unlike the language domain where the token position always starts from zero, there are no {\it standardised absolute global camera poses} in a 3D space. The relationship between two views depends solely on their relative camera transformation.

We now present two distinct designs for spatial position encoding, representing camera poses using 4 DoF for object-centric rendering and 6 DoF for the generic case, respectively. Our design strategy involves directly applying a transformation on global camera poses embedded in the token feature, which allows the dot-product attention to directly encode the relative camera transformation, independent of any coordinate system.

\paragraph{4 DoF CaPE}
In the case of 4 DoF camera poses, we adopt a spherical coordinate system, similar to~\cite{liu2023zero, liu2023syncdreamer}, denoted as ${\bf P}=\{\alpha, \beta, \gamma, r\}$ including azimuth, elevation, camera orientation along the look-at direction, and camera distance (radius), each position component is {\it disentangled}.
 
Mathematically, the position encoding function $\pi ({\bf v,P})$, characterised by its $d$-dimensional token feature ${\bf v}\in \mathbb{R}^d$ and pose ${\bf P}$, should satisfy the following conditions:
\begin{align}
    \left\langle \pi({\bf v}_1, \theta_1), \pi({\bf v}_2, \theta_2) \right\rangle &= \langle \pi({\bf v}_1, \theta_1-\theta_2), \pi({\bf v}_2, 0) \rangle, \label{eq:condition1}\\  
     \left\langle \pi({\bf v}_1, r_1), \pi({\bf v}_2, r_2) \right\rangle &= \langle \pi({\bf v}_1, r_1/r_2), \pi({\bf v}_2, 1) \rangle. \label{eq:condition2}
\end{align}

Here $\langle \cdot,\cdot \rangle$ represents the dot product operation, $\theta_{1,2} \in \{ \alpha, \beta, \gamma\}$, within $\alpha, \gamma \in[0, 2\pi), \beta \in [0, \pi)$, and $r_{1,2} >0$. Essentially, the relative 4 DoF camera transformation is decomposed to the relative angle difference in rotation and the relative scale difference in view radius.

Notably, Eq.~\ref{eq:condition1} aligns with the formula of rotary position encoding (RoPE)~\cite{su2021roformer} derived in the language domain. Given that $\log(r_1) - \log(r_2) = \log(s\cdot r_1) - \log(s \cdot r_2)$ (for any scalar $s>0$),  we may elegantly combine both Eq.~\ref{eq:condition1} and Eq.~\ref{eq:condition2} in a unified formulation using the design strategy in RoPE by transforming feature vector ${\bf v}$ with a block diagonal rotation matrix ${\bf \phi(P)}$ encoding ${\bf P}$.

{\it -- \textbf{4 DoF CaPE}}: $\pi ({\bf v},{\bf P})= {\bf \phi(P)}{\bf v}$,
{
\footnotesize
\begin{align}
\setlength\arraycolsep{4pt}
{\bf \phi(P)} = 
\begin{bmatrix}
 {\bf \Psi} & 0 & \cdots & 0 \\
 0 & {\bf \Psi} & 0 & \vdots \\
 \vdots & 0 & \ddots & 0  \\
0 & \cdots & 0 &  {\bf \Psi} 
\end{bmatrix},\, {\bf \Psi}=
\begin{bmatrix}
 {\bf \Psi}_\alpha & 0 & \cdots & 0 \\
 0 & {\bf \Psi}_\beta & 0 & \vdots \\
 \vdots & 0 & {\bf \Psi}_\gamma & 0  \\
0 & \cdots & 0 &  {\bf \Psi}_r
\end{bmatrix}.
\end{align}
}
\vspace{-2mm}
{\footnotesize
\begin{align}
\text{Rotation: } {\bf \Psi}_\theta &= 
\begin{bmatrix}
\cos \theta & -\sin \theta \\
\sin \theta & \cos \theta
\end{bmatrix}, \\
\text{View Radius: } {\bf \Psi}_r &= 
\begin{bmatrix}
\cos (f(r)) & -\sin (f(r)) \\
\sin (f(r)) & \cos (f(r))
\end{bmatrix},\\
\text{where } f(r) &= \pi\frac{\log r - \log r_{\min}}{\log r_{\max} - \log r_{\min}} \in [0, \pi]\label{eq:log_norm}.
\end{align}
}

Here, $\dim({\bf v})=d$ should be divisible by $2|{\bf P}|=8$.
Note, it's crucial to apply Eq.~\ref{eq:log_norm} to constrain $\log r$ within the range of rotation $[0, \pi]$, so we ensure the dot product monotonically corresponds to its scale difference.

\paragraph{6 DoF CaPE} In the case of 6 DoF camera poses, denoted as  ${\bf P}=\begin{bsmallmatrix}
    {\bf R} & {\bf t} \\
    {\bf 0} & 1
\end{bsmallmatrix}\in SE(3)$, each position component is {\it entangled}, implying that we are not able to reformulate as a multi-dimensional position as in 4 DoF camera poses.

Mathematically, the position encoding function $\pi ({\bf v},{\bf P})$ should now satisfy the following condition:
{\footnotesize 
\begin{align}
    \left\langle \pi({\bf v}_1, {\bf P}_1), \pi({\bf v}_2, {\bf P}_2) \right\rangle 
    = \left\langle \pi({\bf v}_1,{\bf P}_2^{-1}{\bf P}_1), \pi({\bf v}_2, {\bf I}) \right\rangle. \label{eq:condition3}
\end{align}
}

Let's apply a similar strategy as used in 4 DoF CaPE, which increases the dimensionality of ${\bf P}\in\mathbb{R}^{4\times 4}$ to $\phi({\bf P})\in\mathbb{R}^{d\times d}$ by reconstructing it as a block diagonal matrix, with each diagonal element being ${\bf P}$. Since $\phi({\bf P})$ also forms a real Lie group, we may construct $\pi(\cdot,\cdot)$ for a key and query using the following equivalence:
{\footnotesize
\begin{align}
&(\phi({\bf P}_2^{-1}{\bf P}_1)\,{\bf v}_1)^\intercal \left(\phi({\bf I}){\bf v}_2\right)= ({\bf v}_1^\intercal \phi({\bf P}_1^\intercal{\bf P}_2^{-\intercal})){\bf v}_2 \\
 &= ({\bf v}_1^\intercal\, \phi({\bf P}_1^\intercal))(\phi({\bf P}_2^{-\intercal}){\bf v}_2) =(\phi({\bf P}_1){\bf v}_1)^\intercal(\phi({\bf P}_2^{-\intercal}){\bf v}_2) \\
 &= \langle \pi({\bf v}_1, \phi({\bf P}_1)), \pi({\bf v}_2, \phi({\bf P}_2^{-\intercal})) \rangle.
\end{align}
}
{\it -- \textbf{6 DoF CaPE}}: $\pi ({\bf v},{\bf P})= {\bf \phi(P)}{\bf v},$
{
\footnotesize
\begin{align}
\setlength\arraycolsep{4pt}
{\bf \phi(P)} = 
\begin{bmatrix}
 {\bf \Psi} & 0 & \cdots & 0 \\
 0 & {\bf \Psi} & 0 & \vdots \\
 \vdots & 0 & \ddots & 0  \\
0 & \cdots & 0 &  {\bf \Psi} 
\end{bmatrix},\, 
{\bf \Psi}=
\begin{cases}
{\bf P}&\text{if key}\\
{\bf P}^{-\intercal} &\text{if query}
\end{cases}.\label{eq:6dof}
\end{align}
}
Here, $\dim({\bf v})=d$ should be divisible by $\dim({\bf P})=4$.
Similarly, we need to re-scale the translation ${\bf t}$ for each scene within a unit range for efficient model training. It's worth noting that 6 DoF CaPE is concurrently explored in \cite{Miyato2024GTA}, with a focus on scene-level representations.

In both 4 and 6 DoF CaPE implementation, we can efficiently perform matrix multiplication by simply reshaping the vector ${\bf v}$ to match the dimensions of ${\bf \Psi}$ (8 for 4 DoF, 4 for 6 DoF), ensuring faster computation. The PyTorch implementation is attached in Appendix~\ref{app:code}. 

\vspace{-0.1cm}
\section{Experiments}

\paragraph{Training Datasets} In this work, we focus on object-centric view synthesis, training our model on Objaverse-1.0 which consists of 800K objects~\cite{deitke2023objaverse}. This setting allows us to fairly compare with all other 3D diffusion model baselines trained on the same dataset.  We adopt the same training data used in Zero-1-to-3~\cite{liu2023zero}, which contains 12 randomly rendered views per object with randomised environment lighting. To ensure the data quality, we filter out empty rendered images, which make up roughly 1\% of the training data. 

We trained and reported results using EscherNet with both 4 DoF and 6 DoF CaPE. Our observations revealed that 6 DoF CaPE exhibits a slightly improved performance, which we attribute to its more compressed representation space. However, empirically, we found that 4 DoF CaPE yields visually more consistent results when applied to real-world images. Considering that the training data is confined within a 4 DoF object-centric setting, we present EscherNet with 4 DoF CaPE in the main paper. The results obtained with 6 DoF CaPE are provided in Appendix~\ref{app:6dof}. 

In all experiments, we re-evaluate the baseline models by using their officially open-sourced checkpoints on the same set of reference views for a fair comparison. Our experiment settings are provided in Appendix~\ref{app:training}.

\subsection{Results on Novel View Synthesis}
We evaluate EscherNet in novel view synthesis on the Google Scanned Objects dataset (GSO)~\cite{downs2022google} and the RTMV dataset~\cite{tremblay2022rtmv}, comparing with 3D diffusion models for view synthesis,  such as Zero-1-to-3~\cite{liu2023zero} and RealFusion~\cite{melas2023realfusion} (primarily for generation quality with minimal reference views). Additionally, we also evaluate on NeRF Synthetic Dataset~\cite{mildenhall2020nerf}, comparing with state-of-the-art scene-specific neural rendering methods, such as InstantNGP~\cite{muller2022instant} and 3D Gaussian Splatting~\cite{kerbl20233gaussian} (primarily for rendering accuracy with multiple reference views).

Notably, many other 3D diffusion models~\cite{liu2023syncdreamer,long2023wonder3d,shi2023zero123++,ye2023consistent123,liu2023one} prioritise 3D generation rather than view synthesis. This limitation confines them to predicting target views with {\it fixed target poses}, making them not directly comparable.

\paragraph{Compared to 3D Diffusion Models} 
In Tab.~\ref{tab:NVS} and Fig.~\ref{tab:fig_NVS}, we show that EscherNet significantly outperforms 3D diffusion baselines, by a large margin, both quantitatively and qualitatively. Particularly, we outperform Zero-1-to-3-XL despite it being trained on $\times 10$ more training data, and RealFusion despite it requiring expensive score distillation for iterative scene optimisation~\cite{poole2022dreamfusion}. It's worth highlighting that Zero-1-to-3 by design is inherently limited to generating a single target view and cannot ensure self-consistency across multiple target views, while EscherNet can generate multiple consistent target views jointly and provides more precise camera control.

\begin{table}[ht!]
   \centering
   \scriptsize
   \setlength{\tabcolsep}{0.25em}
       \begin{tabular}{lcccccccccc}
       \toprule
       &  \multirow{2}[2]{*}{\makecell[c]{Training \\ Data}} & \multirow{2}[2]{*}{\makecell{\# Ref.\\ Views}} & \multicolumn{3}{c}{{\bf GSO-30}}  & \multicolumn{3}{c}{{\bf RTMV}} \\
       \cmidrule(lr){4-6}      \cmidrule(lr){7-9}
          & &  &PSNR$\uparrow$ & SSIM$\uparrow$ & LPIPS$\downarrow$  &PSNR$\uparrow$ & SSIM$\uparrow$ & LPIPS$\downarrow$  \\
        \midrule 
        RealFusion  & -    & 1 & 12.76 &0.758 &0.382 &-&-&- \\
        Zero123  &  800K  &1  &18.51 &0.856 &0.127  &10.16 &0.505 &0.418\\
        Zero123-XL & 10M  &1 &18.93 &0.856 &0.124 &10.59 &0.520 &0.401\\
        \midrule
        EscherNet    & 800k  &1  &20.24 &0.884 &0.095 &10.56 &0.518 &0.410\\
        EscherNet   & 800k  &2  &22.91 &0.908 &0.064 &12.66 &0.585 &0.301\\
        EscherNet    & 800k  &3  &24.09 &0.918 &0.052 &13.59 &0.611 &0.258\\
        EscherNet   & 800k &5  &25.09 &0.927 &0.043 &14.52 &0.633 &0.222\\
        EscherNet  & 800k &10   &25.90 &0.935 &0.036 &15.55 &0.657 &0.185\\
         \bottomrule
       \end{tabular}
   \caption{{\bf Novel view synthesis performance on GSO and RTMV datasets.} EscherNet outperforms Zero-1-to-3-XL with significantly less training data and RealFusion without extra SDS optimisation. Additionally, EscherNet's performance exhibits further improvement with the inclusion of more reference views.}
   \label{tab:NVS}
   \vspace{-0.4cm}
\end{table}

\paragraph{Compared to Neural Rendering Methods} 
In Tab.~\ref{tab:nerf2} and Fig.~\ref{tab:fig_nerf2}, we show that EscherNet again offers plausible view synthesis in a zero-shot manner, without scene-specific optimisation required by both InstantNGP and 3D Gaussian Splatting. Notably, EscherNet leverages a generalised understanding of objects acquired through large-scale training, allowing it to interpret given views both semantically and spatially, even when conditioned on a limited number of reference views. However, with an increase in the number of reference views, both InstantNGP and 3D Gaussian Splatting exhibit a significant improvement in the rendering quality. To achieve a photo-realistic neural rendering while retaining the advantages of a generative formulation remains an important research challenge.

\begin{table}[t!]
   \centering
   \footnotesize
   \setlength{\tabcolsep}{0.45em}
       \begin{tabular}{lcccccccc}
       \toprule
        \multicolumn{9}{c}{\# Reference Views (Less $\to$ More)} \\
         & 1 & 2 & 3 & 5 & 10 & 20 & 50 & 100 \\
        \midrule 
        \multicolumn{9}{l}{{\bf InstantNGP (Scene Specific Training)}} \\
         PSNR$\uparrow$ & 10.92 & 12.42& 14.27& 18.17& 22.96& 24.99& 26.86& 27.30 \\
        SSIM$\uparrow$  & 0.449 & 0.521& 0.618& 0.761& 0.881& 0.917& 0.946& 0.953\\
        LPIPS$\downarrow$  & 0.627 & 0.499& 0.391& 0.228& 0.091& 0.058& 0.034& 0.031 \\
        \midrule 
        \multicolumn{9}{l}{{\bf GaussianSplatting (Scene Specific Training)}} \\
         PSNR$\uparrow$ & 9.44 & 10.78& 12.87& 17.09& 23.04& 25.34& 26.98& 27.11 \\
        SSIM$\uparrow$  & 0.391 & 0.432& 0.546& 0.732& 0.876& 0.919& 0.942& 0.944\\
        LPIPS$\downarrow$  & 0.610 & 0.541& 0.441& 0.243& 0.085& 0.054& 0.041& 0.041 \\
        \midrule 
        \multicolumn{9}{l}{{\bf EscherNet (Zero Shot Inference)}} \\
        PSNR$\uparrow$  & 13.36 & 14.95& 16.19& 17.16& 17.74& 17.91& 18.05& 18.15 \\
        SSIM$\uparrow$  & 0.659 & 0.700& 0.729& 0.748& 0.761& 0.765& 0.769& 0.771\\
        LPIPS$\downarrow$ & 0.291 & 0.208& 0.161& 0.127& 0.114& 0.106& 0.099& 0.097\\
        \bottomrule
       \end{tabular}
   \caption{{\bf Novel view synthesis performance on NeRF Synthetic dataset.} EscherNet outperforms both InstantNGP and Gaussian Splatting when provided with fewer than five reference views while requiring no scene-specific optimisation. However, as the number of reference views increases, both methods show a more significant improvement in rendering quality.}
   \label{tab:nerf2}
   \vspace{-0.4cm}
\end{table}

\begin{figure}[ht!]
   \centering
   \scriptsize
  \renewcommand{\arraystretch}{0.6}
   \setlength{\tabcolsep}{0em}
       \begin{tabular}{*{6}{C{0.165\linewidth}}}
       \toprule
        \multicolumn{6}{c}{\# Reference Views (Less $\to$ More)} \\
         1 & 2 & 3 & 5 & 10 & 20 \\
        \midrule 
        \multicolumn{6}{l}{{\bf InstantNGP (Scene Specific Training)}} \\
        \makecell{\includegraphics[width=\linewidth]{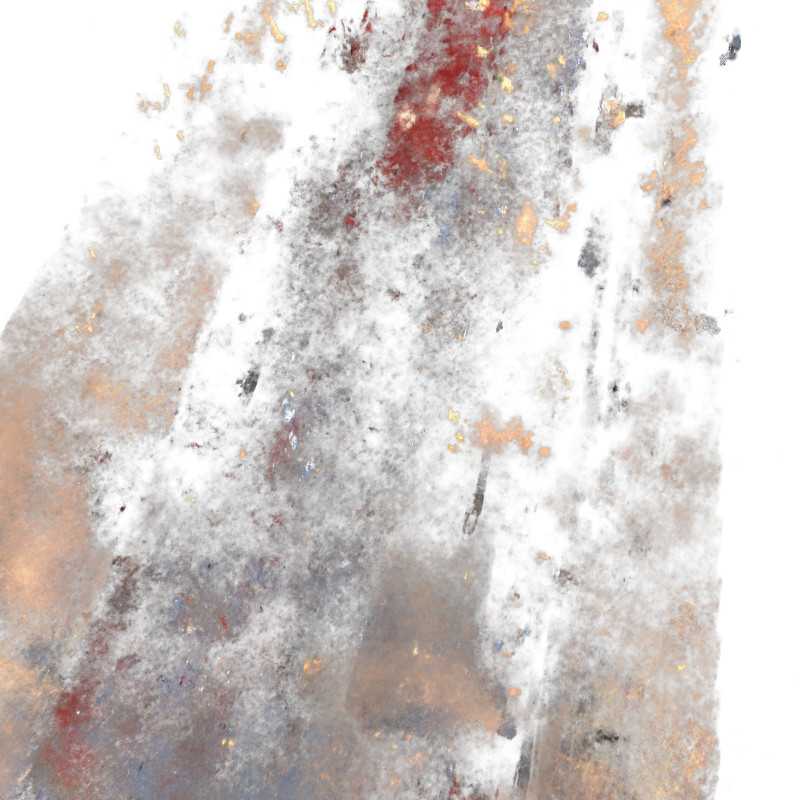} \\ PSNR 10.37} & \makecell{\includegraphics[width=\linewidth]{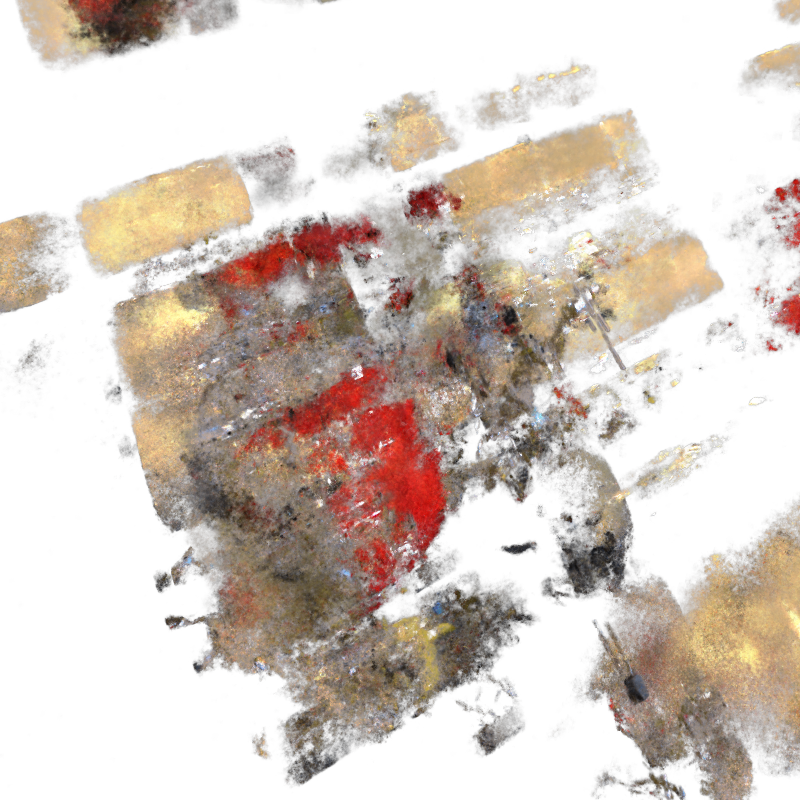} \\ PSNR 11.72} & \makecell{\includegraphics[width=\linewidth]{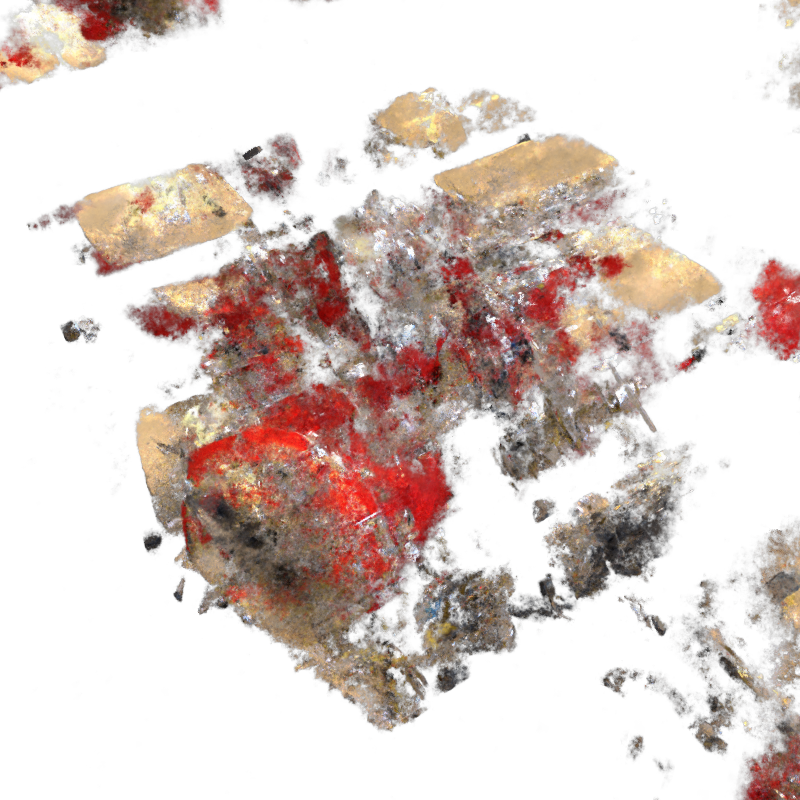} \\ PSNR 12.82} & \makecell{\includegraphics[width=\linewidth]{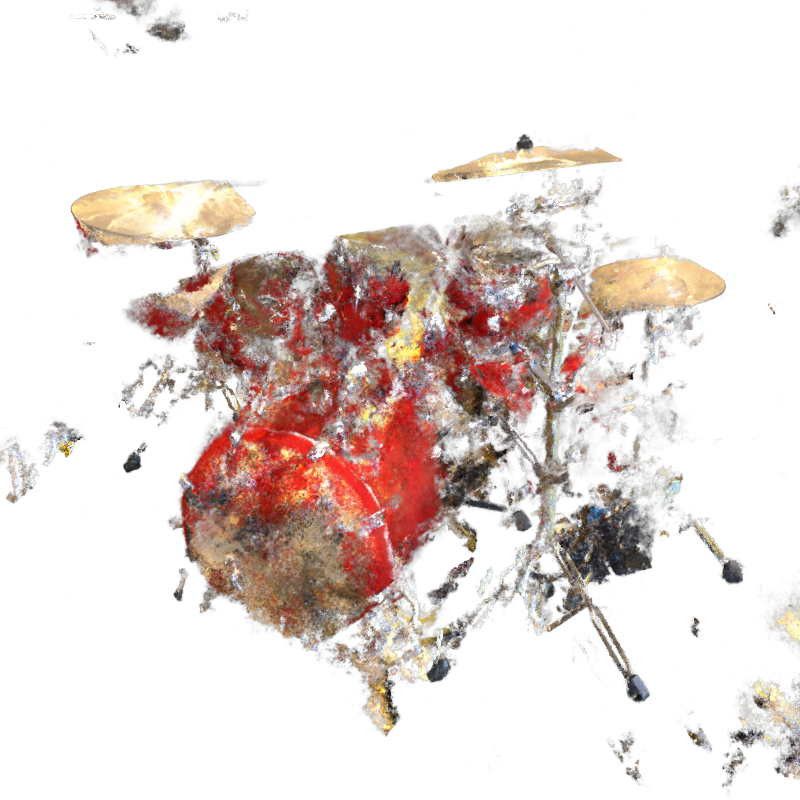} \\ PSNR 15.58} & \makecell{\includegraphics[width=\linewidth]{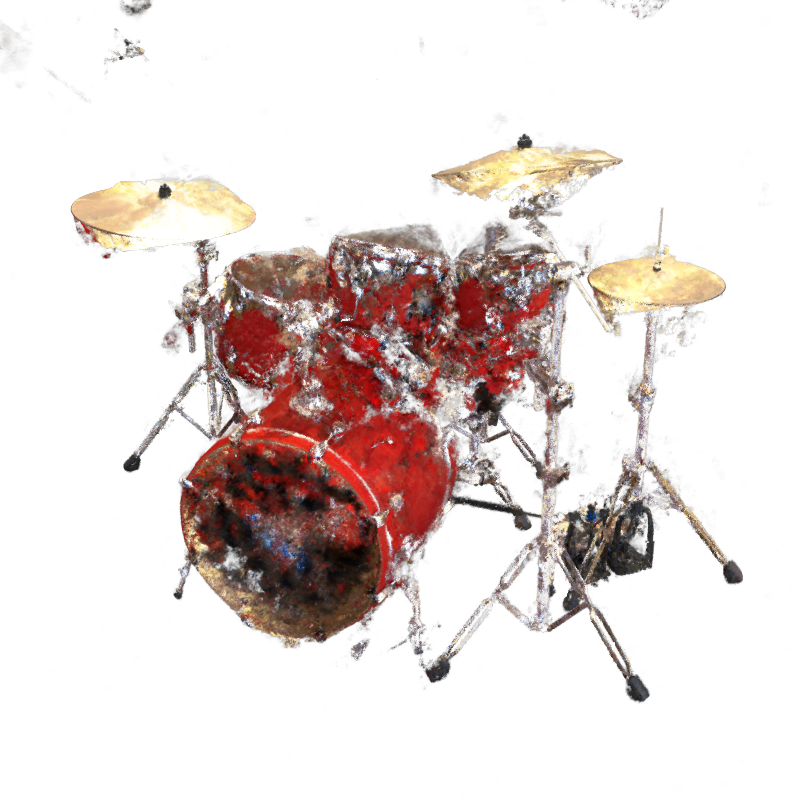} \\ PSNR 19.71} & \makecell{\includegraphics[width=\linewidth]{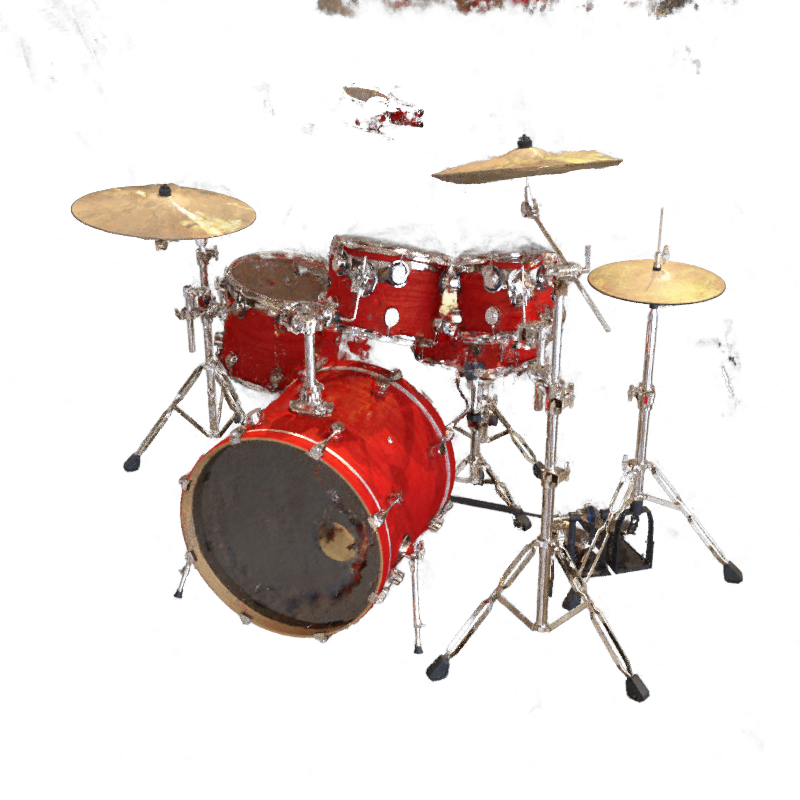} \\ PSNR 21.28} \\
        \midrule 
        \multicolumn{6}{l}{{\bf 3D Gaussian Splatting (Scene Specific Training)}} \\
        \makecell{\includegraphics[width=\linewidth]{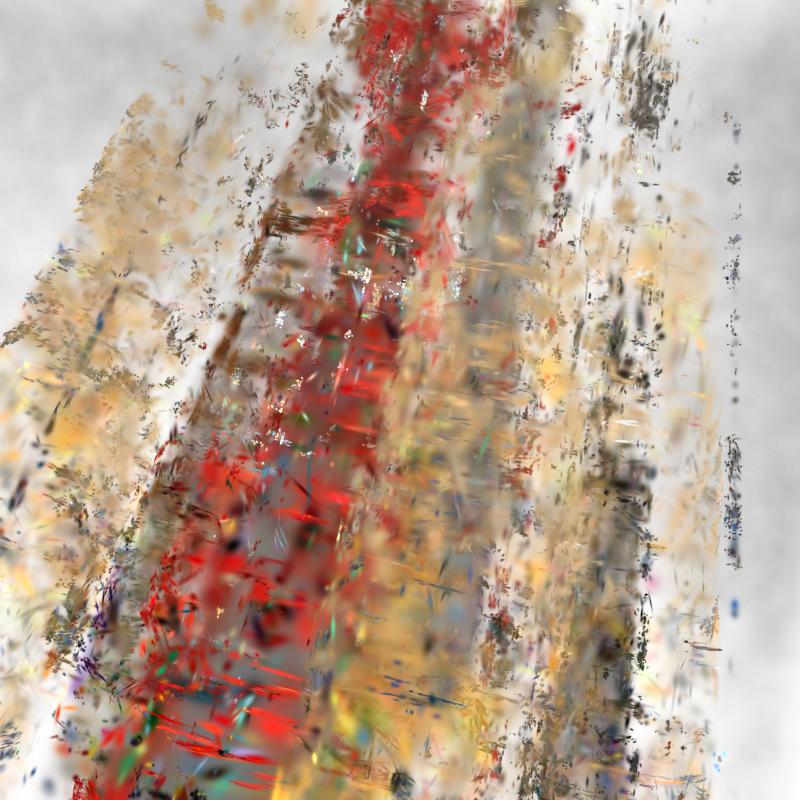} \\ PSNR 9.14} & \makecell{\includegraphics[width=\linewidth]{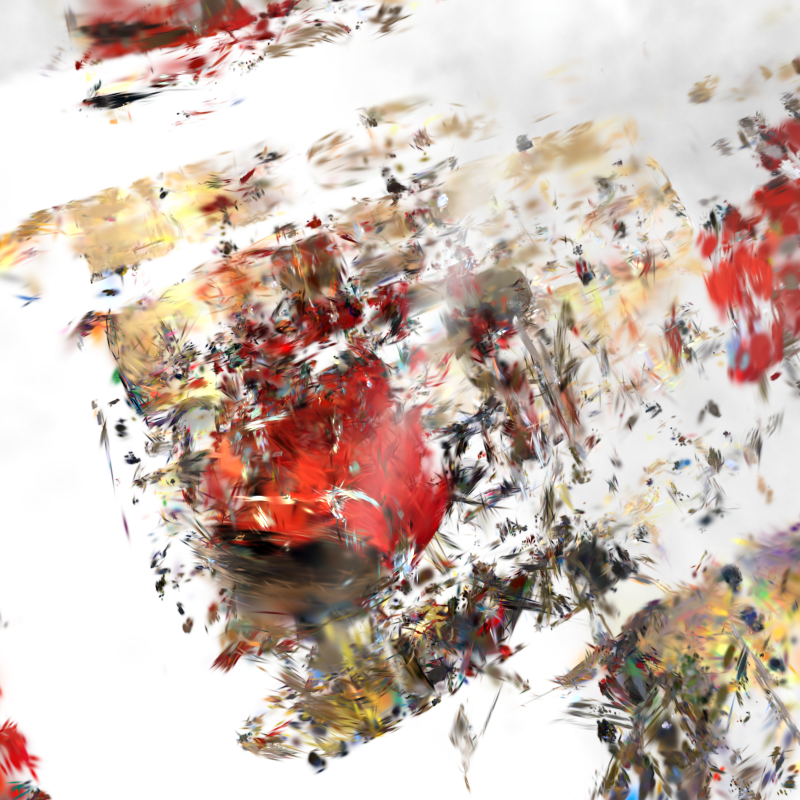} \\ PSNR 10.63} & \makecell{\includegraphics[width=\linewidth]{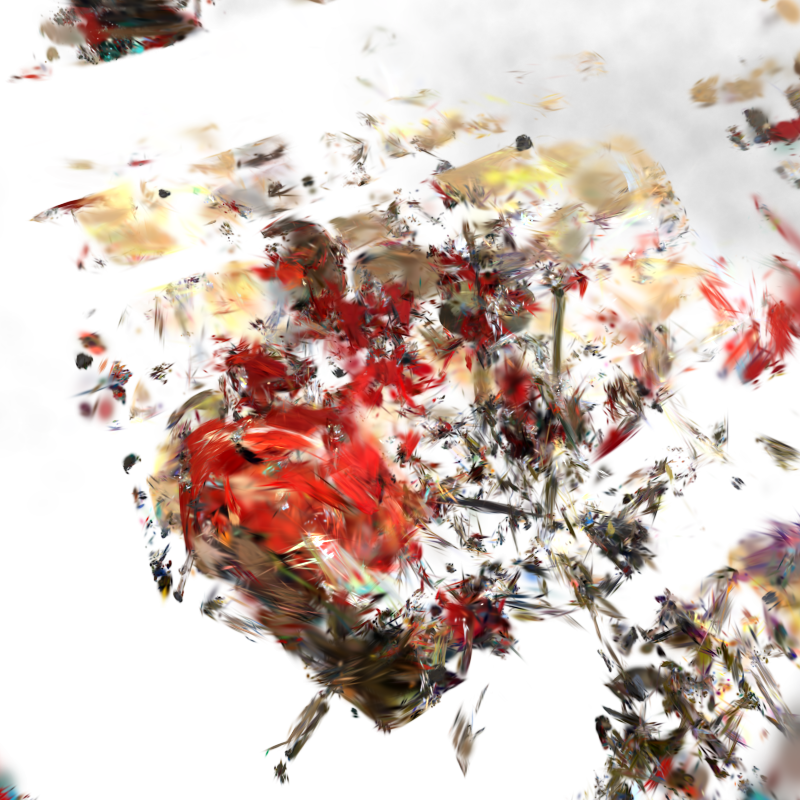} \\ PSNR 11.43} & \makecell{\includegraphics[width=\linewidth]{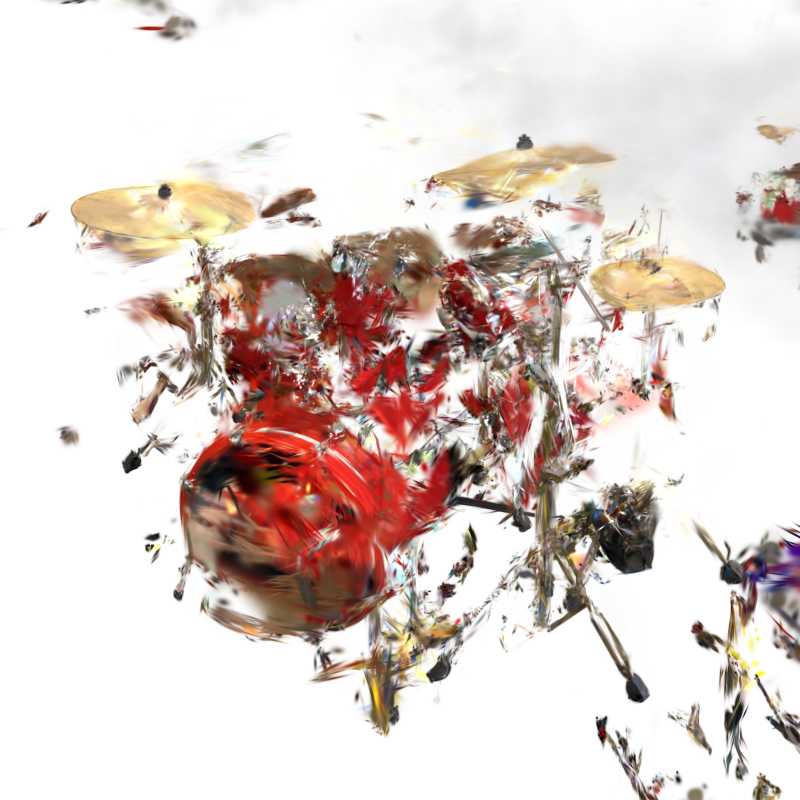} \\ PSNR 14.81} & \makecell{\includegraphics[width=\linewidth]{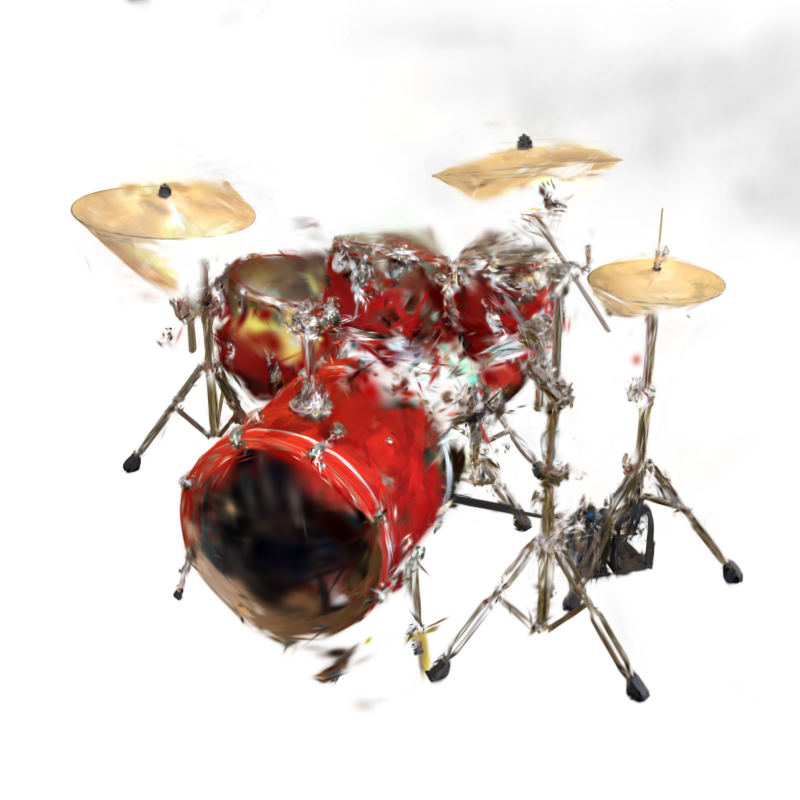} \\ PSNR 20.15} & \makecell{\includegraphics[width=\linewidth]{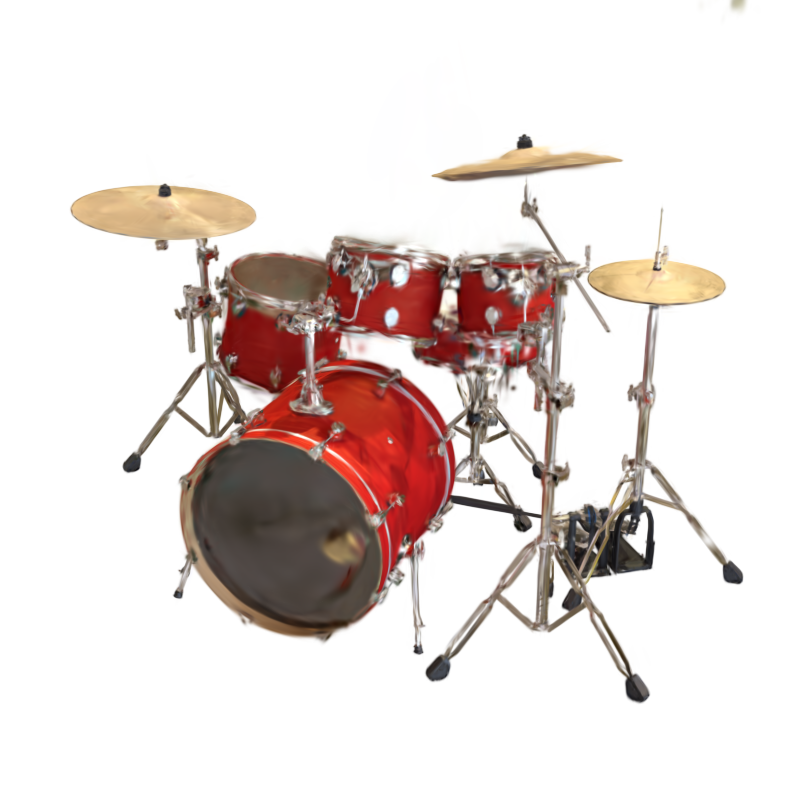} \\ PSNR 22.88} \\
        \midrule 
        \multicolumn{6}{l}{{\bf EscherNet (Zero Shot Inference)}} \\
        \makecell{\includegraphics[width=\linewidth]{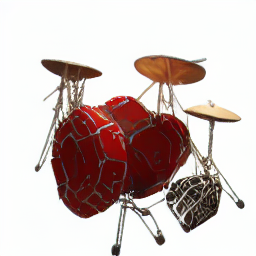} \\ PSNR 10.10} & \makecell{\includegraphics[width=\linewidth]{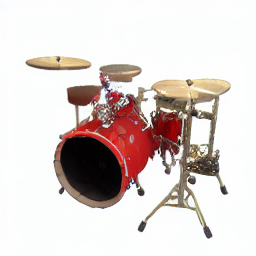} \\ PSNR 13.25} & \makecell{\includegraphics[width=\linewidth]{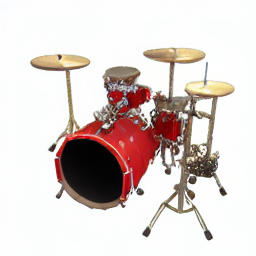} \\ PSNR 13.43} & \makecell{\includegraphics[width=\linewidth]{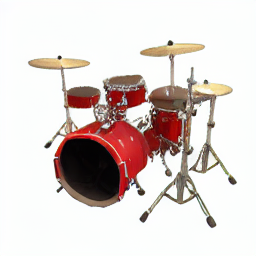} \\ PSNR 14.33} & \makecell{\includegraphics[width=\linewidth]{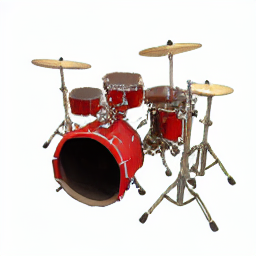} \\ PSNR 14.97} & \makecell{\includegraphics[width=\linewidth]{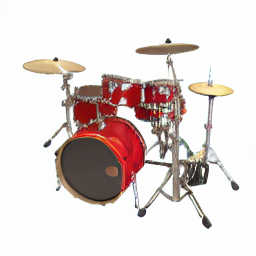} \\ PSNR 15.65} \\
        \bottomrule
       \end{tabular}
   \caption{{\bf Generated views visualisation on the NeRF Synthetic drum scene.} 
   EscherNet generates plausible view synthesis even when provided with very limited reference views, while neural rendering methods fail to generate any meaningful content. However, when we have more than 10 reference views, scene-specific methods exhibit a substantial improvement in rendering quality. 
   We report the mean PSNR averaged across all test views from the drum scene. Results for other scenes and/or with more reference views are shown in Appendix~\ref{app:nerf}.}
   \label{tab:fig_nerf2}
   \vspace{-0.4cm}
\end{figure}

\begin{figure*}[ht!]
   \centering
   \footnotesize
   \renewcommand{\arraystretch}{0.6}
   \setlength{\tabcolsep}{0em}
       \begin{tabular}{L{0.102\linewidth} *{12}{C{0.074\linewidth}}}
       \toprule
        \makecell[l]{Reference\\ Views}
        & \multicolumn{3}{C{0.22\linewidth}}{
        \includegraphics[width=\linewidth]{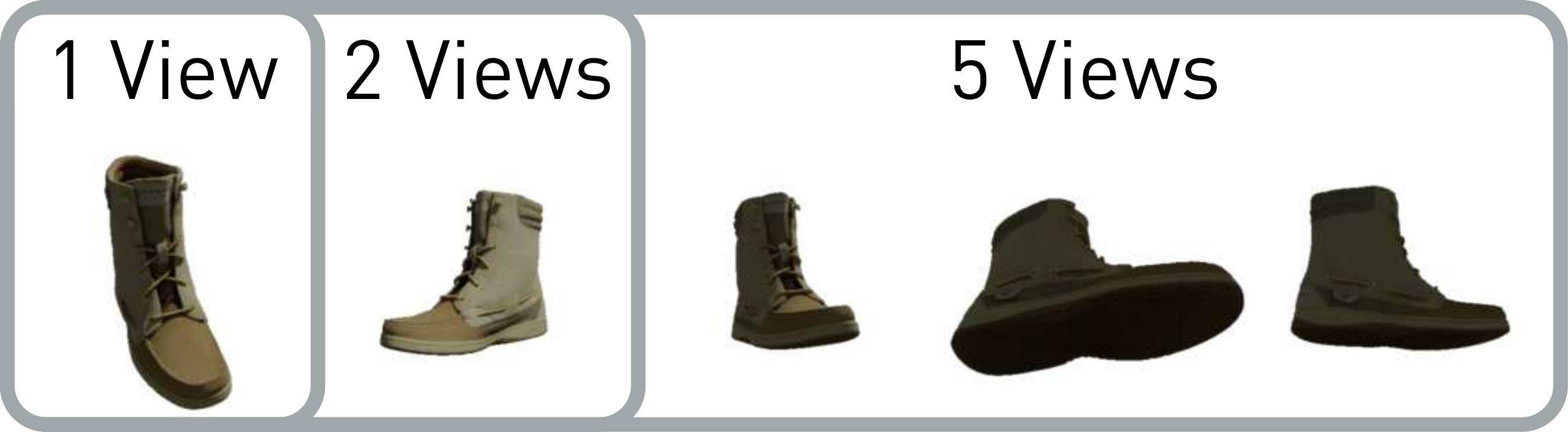}
        }
        & \multicolumn{3}{C{0.22\linewidth}}{
        \includegraphics[width=\linewidth]{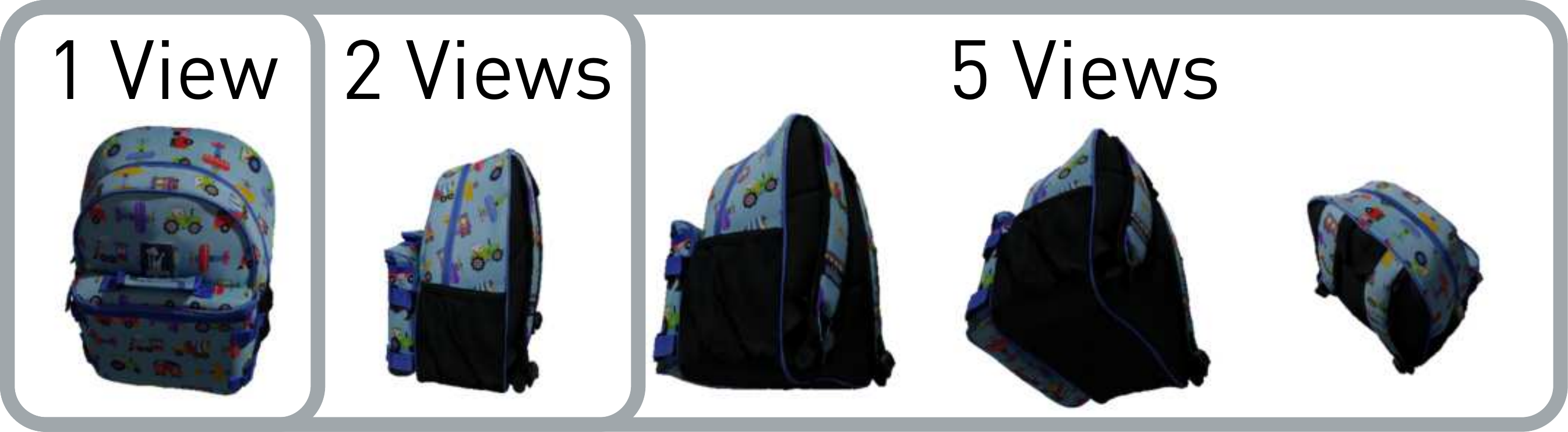}
        }
        & \multicolumn{3}{C{0.22\linewidth}}{       
        \includegraphics[width=\linewidth]{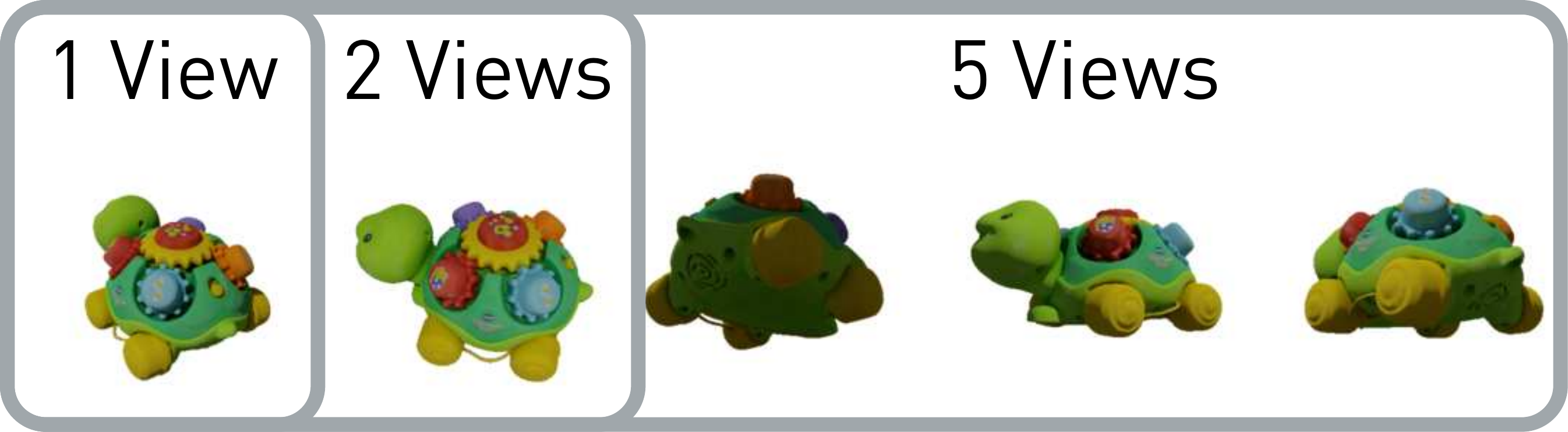}
        } 
        & \multicolumn{3}{C{0.22\linewidth}}{
        \includegraphics[width=\linewidth]{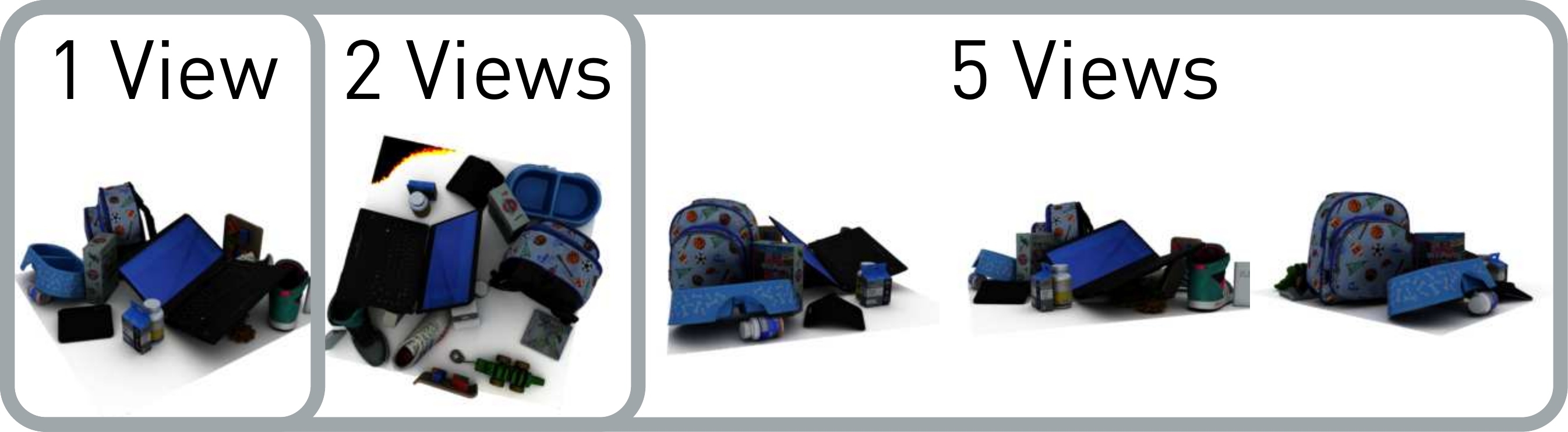}
        } 
        \\
        \hline
        \makecell[l]{Zero-1-to-3-XL\\ {[1 View]}} 
        &\includegraphics[trim={1cm 1cm 1cm 1cm}, clip, width=\linewidth]{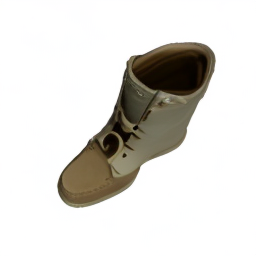}
        &\includegraphics[trim={1cm 0.5cm 1cm 1cm}, clip, width=\linewidth]{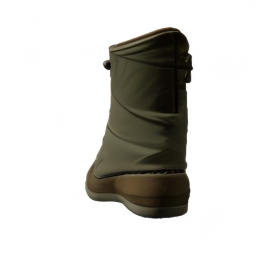} 
        &\includegraphics[trim={1cm 1cm 0.5cm 1cm}, clip, width=\linewidth]{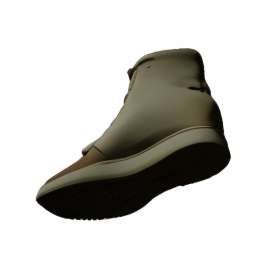}
        &\includegraphics[width=\linewidth]{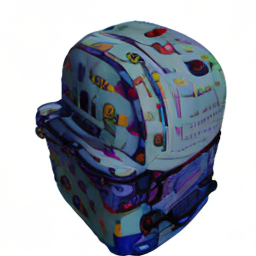}
        &\includegraphics[width=\linewidth]{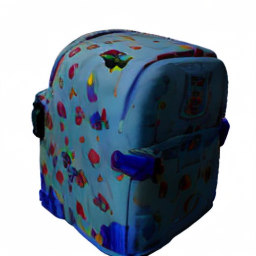} 
        &\includegraphics[width=\linewidth]{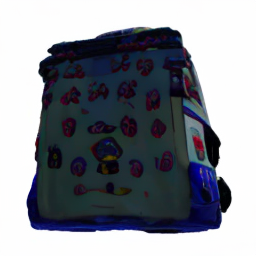}
        &\includegraphics[trim={1cm 2cm 0.5cm 1cm}, clip, width=\linewidth]{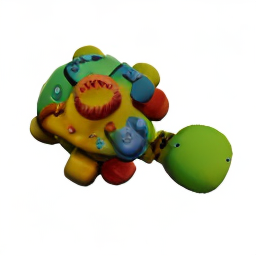}   
        &\includegraphics[trim={1cm 1cm 2cm 1cm}, clip, width=\linewidth]{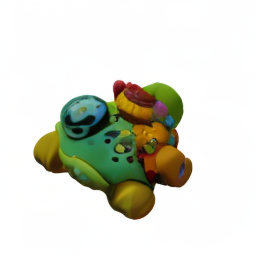}
        &\includegraphics[trim={1cm 1.8cm 1cm 1cm}, clip, width=\linewidth]{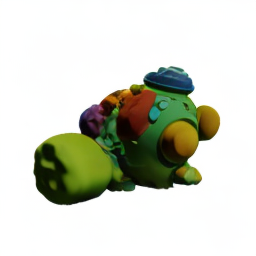}
        &\includegraphics[width=\linewidth]{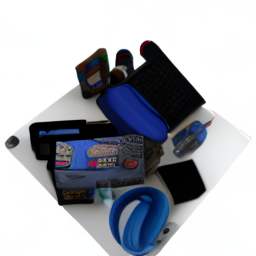} &\includegraphics[width=\linewidth]{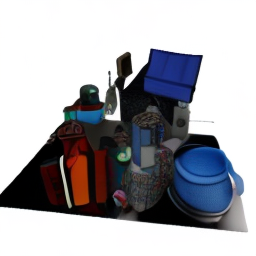}   
        &\includegraphics[width=\linewidth]{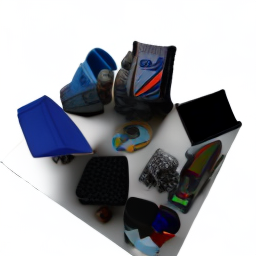} \\
        \makecell[l]{EscherNet\\ {[1 View]}} 
        &\includegraphics[trim={1cm 1cm 1cm 1cm}, clip, width=\linewidth]{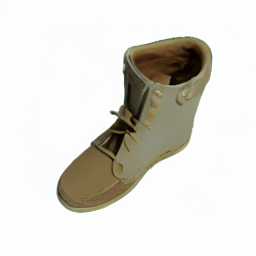}
        &\includegraphics[trim={1cm 0.5cm 1cm 1cm}, clip, width=\linewidth]{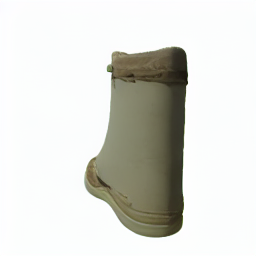} 
        &\includegraphics[trim={1cm 1cm 0.5cm 1cm}, clip, width=\linewidth]{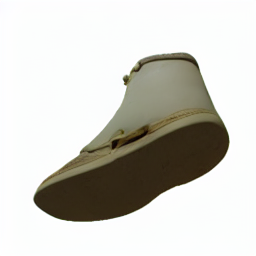}
        &\includegraphics[width=\linewidth]{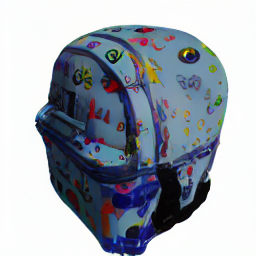} 
        &\includegraphics[width=\linewidth]{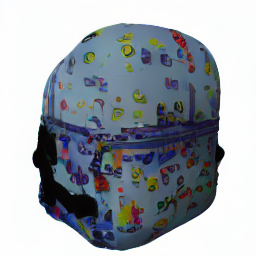} 
        &\includegraphics[width=\linewidth]{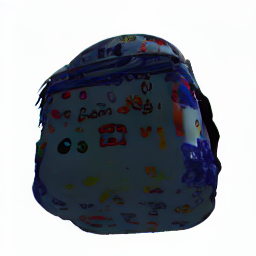}
        &\includegraphics[trim={1cm 2cm 0.5cm 1cm}, clip, width=\linewidth]{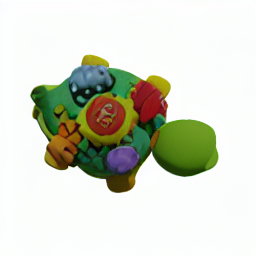}   
        &\includegraphics[trim={1cm 1cm 2cm 1cm}, clip, width=\linewidth]{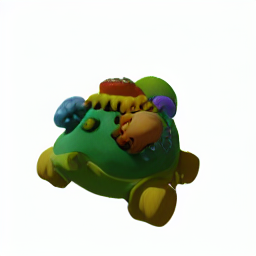}
        &\includegraphics[trim={1cm 1.8cm 1cm 1cm}, clip, width=\linewidth]{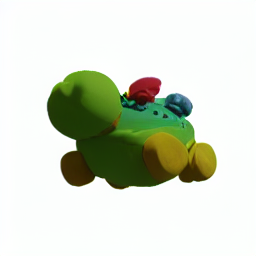}
        &\includegraphics[width=\linewidth]{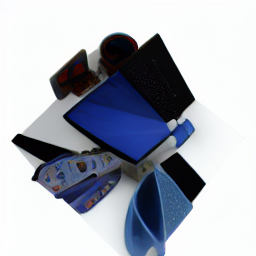} &\includegraphics[width=\linewidth]{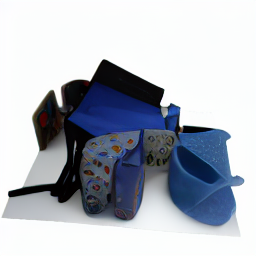}   
        &\includegraphics[width=\linewidth]{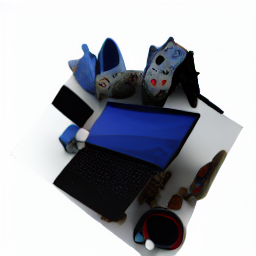} \\    
        \makecell[l]{EscherNet\\ {[2 Views]}} 
        &\includegraphics[trim={1cm 1cm 1cm 1cm}, clip, width=\linewidth]{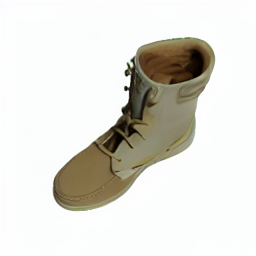}
        &\includegraphics[trim={1cm 0.5cm 1cm 1cm}, clip, width=\linewidth]{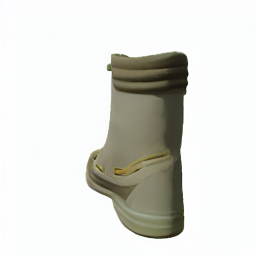} 
        &\includegraphics[trim={1cm 1cm 0.5cm 1cm}, clip, width=\linewidth]{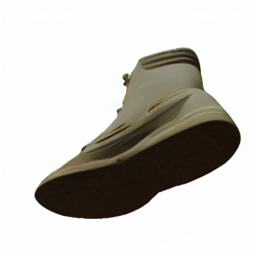}
        &\includegraphics[width=\linewidth]{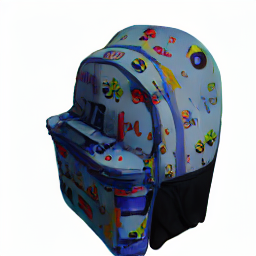} 
        &\includegraphics[width=\linewidth]{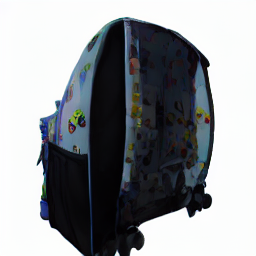} 
        &\includegraphics[width=\linewidth]{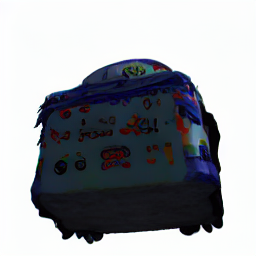}
        &\includegraphics[trim={1cm 2cm 0.5cm 1cm}, clip, width=\linewidth]{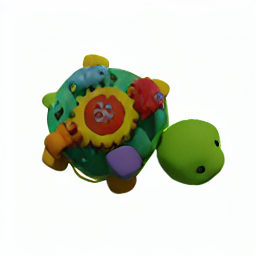}   
        &\includegraphics[trim={1cm 1cm 2cm 1cm}, clip, width=\linewidth]{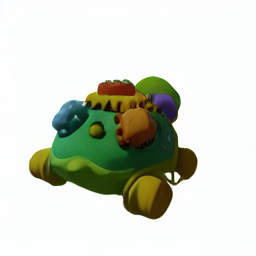} 
        &\includegraphics[trim={1cm 1.8cm 1cm 1cm}, clip, width=\linewidth]{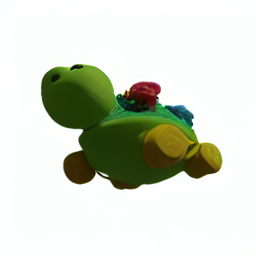}
        &\includegraphics[width=\linewidth]{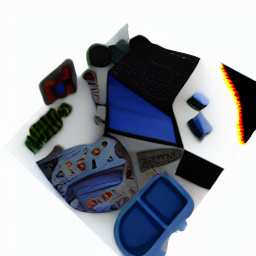} &\includegraphics[width=\linewidth]{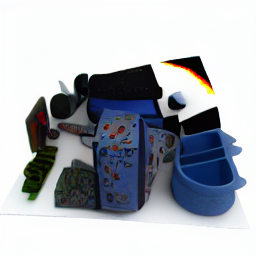}   
        &\includegraphics[width=\linewidth]{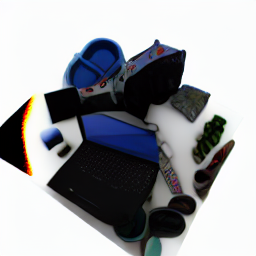} \\
        \makecell[l]{EscherNet\\ {[5 Views]}} 
        &\includegraphics[trim={1cm 1cm 1cm 1cm}, clip, width=\linewidth]{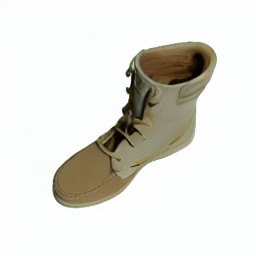}
        &\includegraphics[trim={1cm 0.5cm 1cm 1cm}, clip, width=\linewidth]{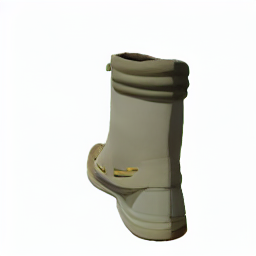} 
        &\includegraphics[trim={1cm 1cm 0.5cm 1cm}, clip, width=\linewidth]{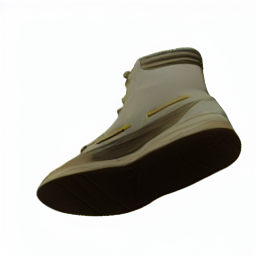}
        &\includegraphics[width=\linewidth]{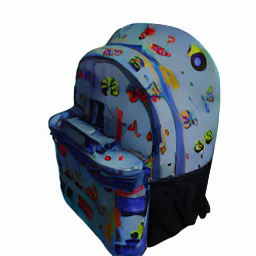} 
        &\includegraphics[width=\linewidth]{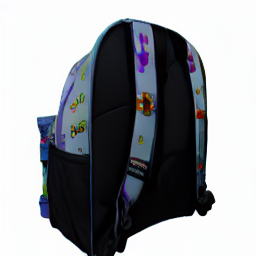} 
        &\includegraphics[width=\linewidth]{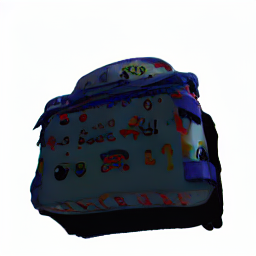}
        &\includegraphics[trim={1cm 2cm 0.5cm 1cm}, clip, width=\linewidth]{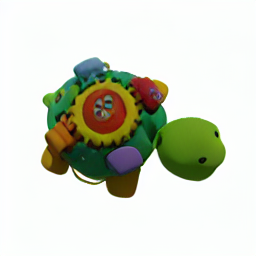}   
        &\includegraphics[trim={1cm 1cm 2cm 1cm}, clip, width=\linewidth]{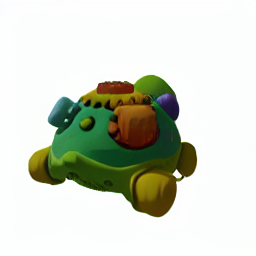}
        &\includegraphics[trim={1cm 1.8cm 1cm 1cm}, clip, width=\linewidth]{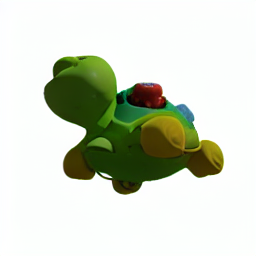}
        &\includegraphics[width=\linewidth]{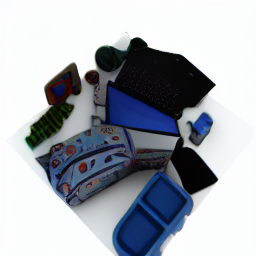} &\includegraphics[width=\linewidth]{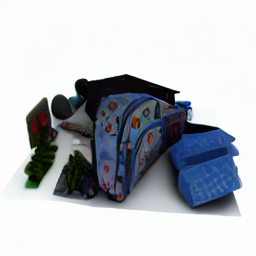}   
        &\includegraphics[width=\linewidth]{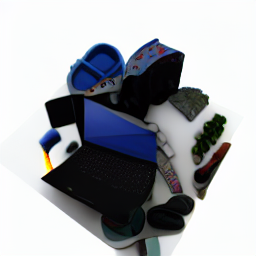} \\

        \hline
        \makecell[l]{Ground\\Truth} 
        &\includegraphics[trim={1cm 1cm 1cm 1cm}, clip, width=\linewidth]{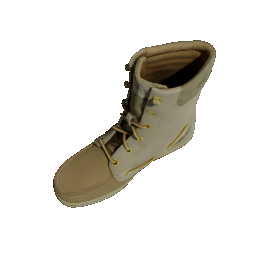}
        &\includegraphics[trim={1cm 0.5cm 1cm 1cm}, clip, width=\linewidth]{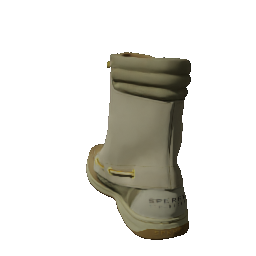} 
        &\includegraphics[trim={1cm 1cm 0.5cm 1cm}, clip, width=\linewidth]{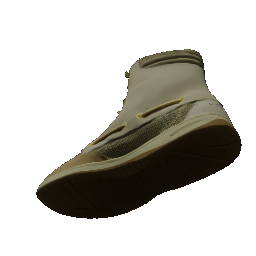}
        &\includegraphics[width=\linewidth]{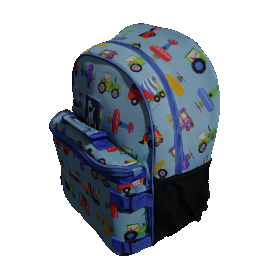}
        &\includegraphics[width=\linewidth]{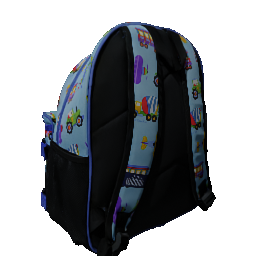} 
        &\includegraphics[width=\linewidth]{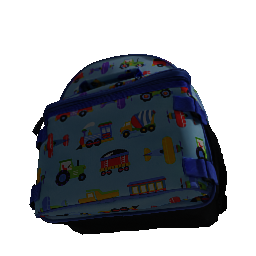}
        &\includegraphics[trim={1cm 2cm 0.5cm 1cm}, clip, width=\linewidth]{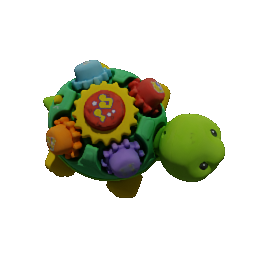}   
        &\includegraphics[trim={1cm 1cm 2cm 1cm}, clip, width=\linewidth]{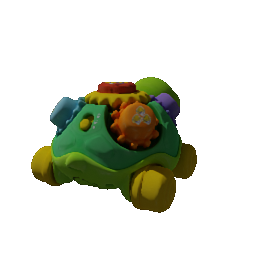}
        &\includegraphics[trim={1cm 1.8cm 1cm 1cm}, clip, width=\linewidth]{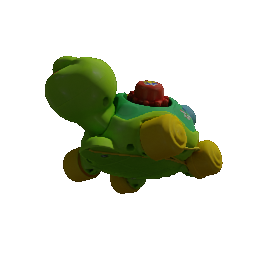}
        &\includegraphics[width=\linewidth]{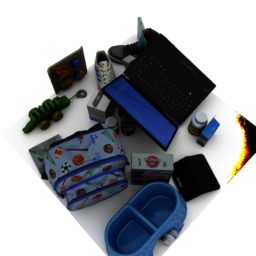} &\includegraphics[width=\linewidth]{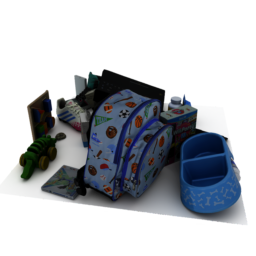}   
        &\includegraphics[width=\linewidth]{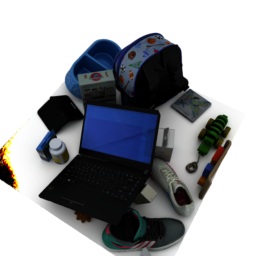} \\
        \bottomrule
       \end{tabular}
   \caption{{\bf Novel view synthesis visualisation on GSO and RTMV datasets.} EscherNet outperforms Zero-1-to-3-XL, delivering superior generation quality and finer camera control. Notably, when conditioned with additional views, EscherNet exhibits an enhanced resemblance of the generated views to ground-truth textures, revealing more refined texture details such as in the backpack straps and turtle shell.}
   \label{tab:fig_NVS} 
   \vspace{-0.4cm}
\end{figure*}

\subsection{Results on 3D Generation}

In this section, we perform single/few-image 3D generation on the GSO dataset. We compare with SoTA 3D generation baselines: Point-E~\cite{nichol2022pointe} for direct point cloud generation, Shape-E~\cite{jun2023shape} for direct NeRF generation, DreamGaussian~\cite{jun2023shape} for optimising 3D Gaussian~\cite{kerbl20233gaussian} with SDS guidance, One-2-3-45~\cite{liu2023one} for decoding an SDF using multiple views predicted from Zero-1-to-3, and SyncDreamer~\cite{liu2023syncdreamer} for fitting an SDF using NeuS~\cite{wang2021neus} from 16 consistent fixed generated views. We additionally include NeuS trained on reference views for few-image 3D reconstruction baselines.  

Given any reference views, EscherNet can generate multiple 3D consistent views, allowing for the straightforward adoption with NeuS~\cite{wang2021neus} for 3D reconstruction. We generate 36 fixed views, varying the azimuth from 0$^{\circ}$ to 360$^{\circ}$ with a rendering every 30$^{\circ}$ at a set of elevations (-30$^{\circ}$, 0$^{\circ}$, 30$^{\circ}$), which serve as inputs for our NeuS reconstruction.

\begin{figure*}[ht!]
   \centering
   \footnotesize
  \renewcommand{\arraystretch}{0.6}
   \setlength{\tabcolsep}{0.34em}
       \begin{tabular}{C{0.08\linewidth}|*{4}{C{0.17\linewidth}}|C{0.17\linewidth}}
       \toprule
       \includegraphics[trim={1.5cm 1.5cm 1.5cm 1.5cm}, clip,width=0.08\textwidth]{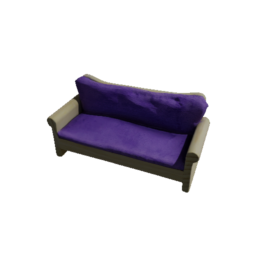} &\includegraphics[trim={1.5cm 1.5cm 1.5cm 1.5cm}, clip, width=0.48\linewidth]{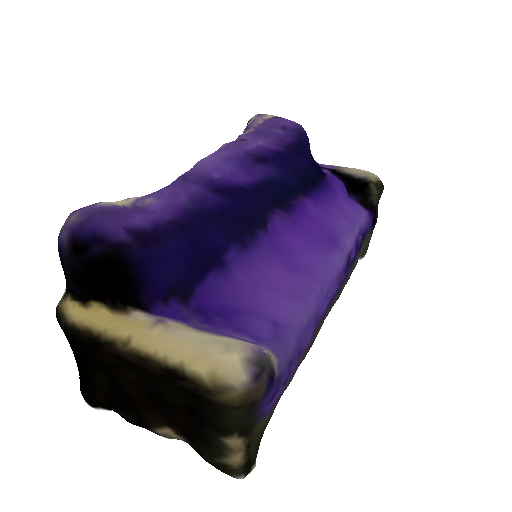}%
       \includegraphics[trim={1.5cm 1.5cm 1.5cm 1.5cm}, clip, width=0.48\linewidth]{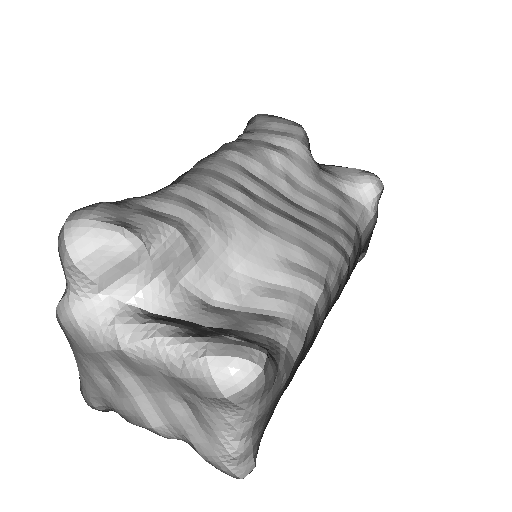} & 
       \includegraphics[trim={1.5cm 1.5cm 1.5cm 1.5cm}, clip, width=0.48\linewidth]{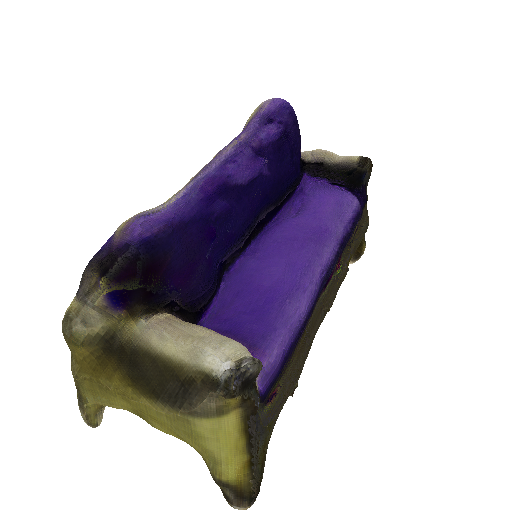}%
       \includegraphics[trim={1.5cm 1.5cm 1.5cm 1.5cm}, clip, width=0.48\linewidth]{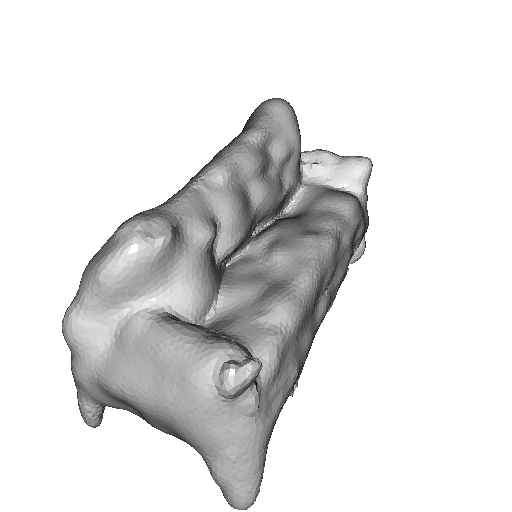} &
       \includegraphics[trim={1.5cm 1.5cm 1.5cm 1.5cm}, clip, width=0.48\linewidth]{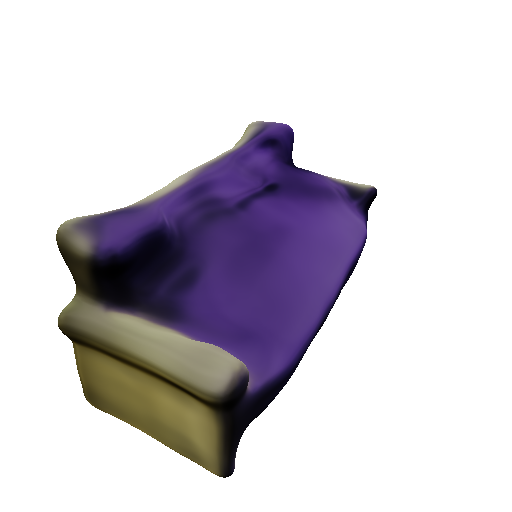}%
       \includegraphics[trim={1.5cm 1.5cm 1.5cm 1.5cm}, clip, width=0.48\linewidth]{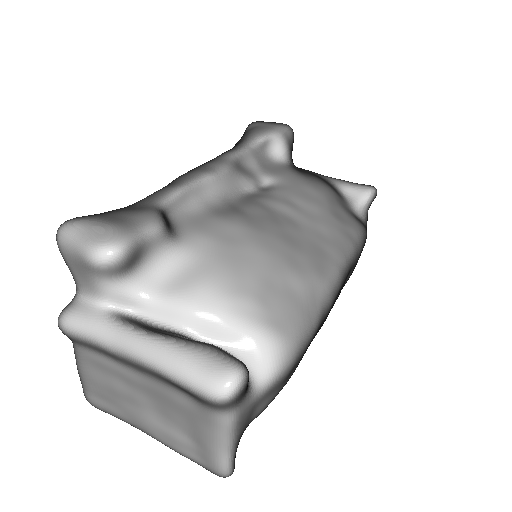} &
       \includegraphics[trim={1.5cm 1.5cm 1.5cm 1.5cm}, clip, width=0.48\linewidth]{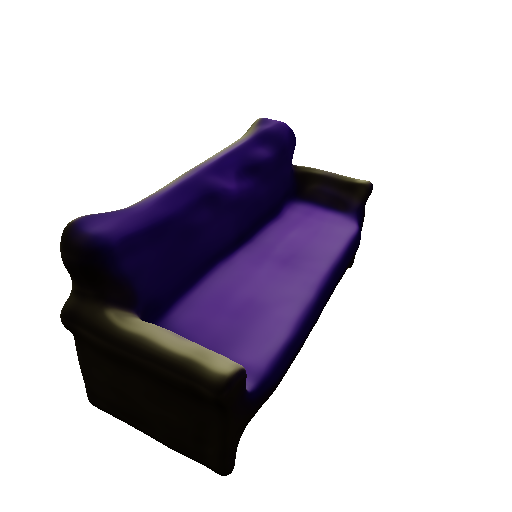}%
       \includegraphics[trim={1.5cm 1.5cm 1.5cm 1.5cm}, clip, width=0.48\linewidth]{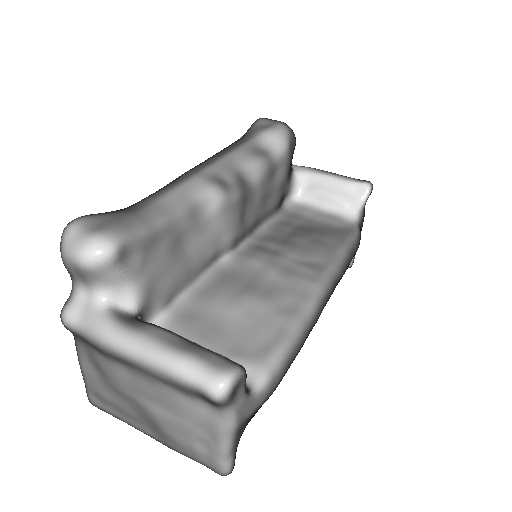} &
       \includegraphics[trim={1.5cm 1.5cm 1.5cm 1.5cm}, clip, width=0.48\linewidth]{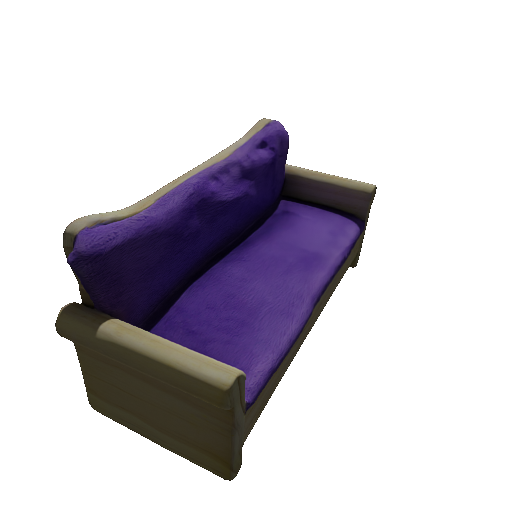}%
       \includegraphics[trim={1.5cm 1.5cm 1.5cm 1.5cm}, clip, width=0.48\linewidth]{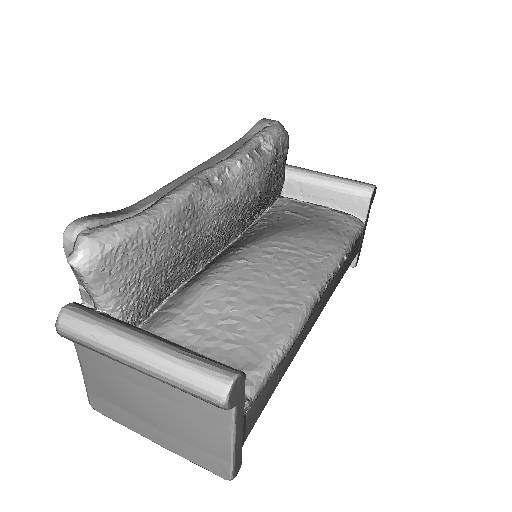} \\
       \includegraphics[trim={1.5cm 0cm 1cm 1.5cm}, clip,width=\linewidth]{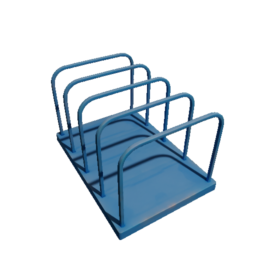}&\includegraphics[trim={0cm 0cm 1.5cm 1.5cm}, clip, width=0.48\linewidth]{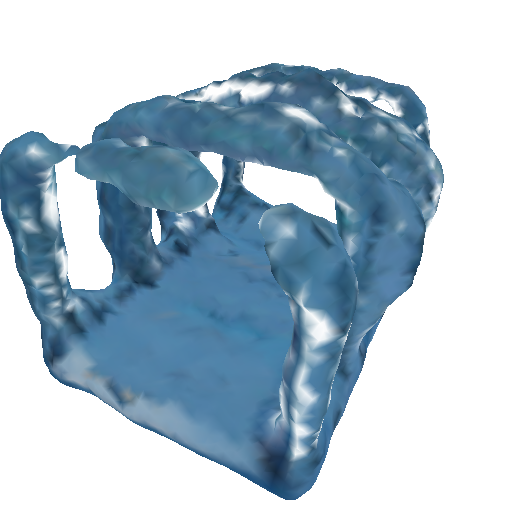}%
       \includegraphics[trim={0cm 0cm 1.5cm 1.5cm}, clip, width=0.48\linewidth]{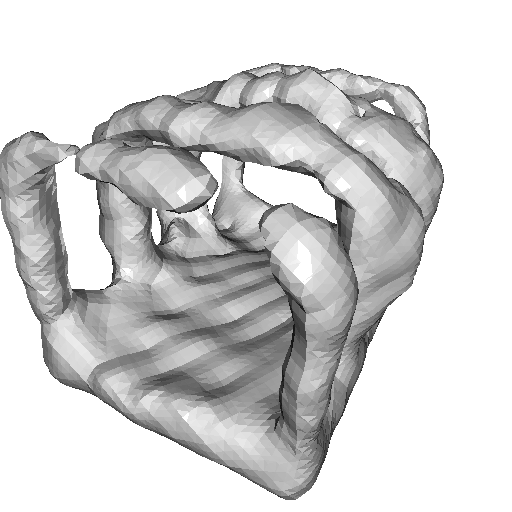} & 
       \includegraphics[trim={0cm 0cm 1.5cm 1.5cm}, clip, width=0.48\linewidth]{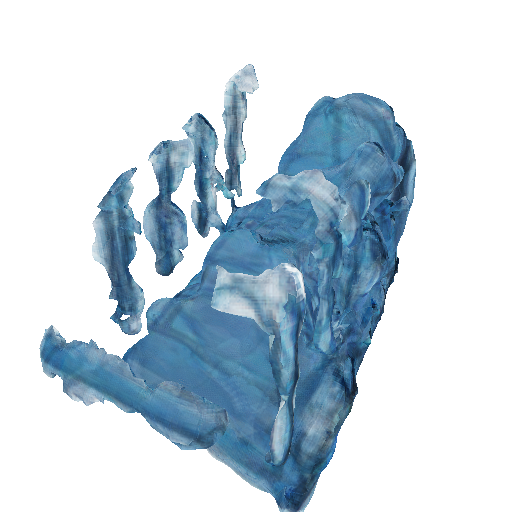}%
       \includegraphics[trim={0cm 0cm 1.5cm 1.5cm}, clip, width=0.48\linewidth]{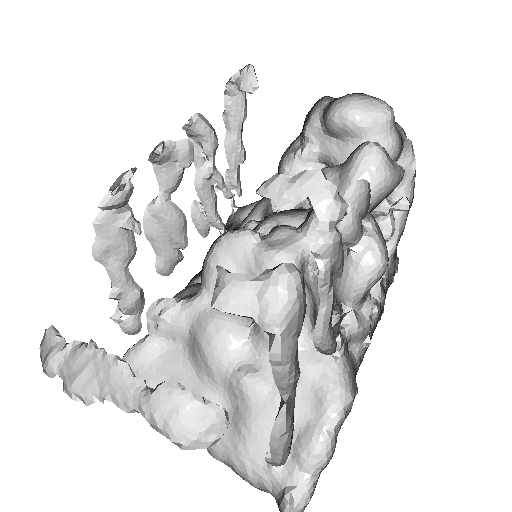} &
       \includegraphics[trim={0cm 0cm 1.5cm 1.5cm}, clip, width=0.48\linewidth]{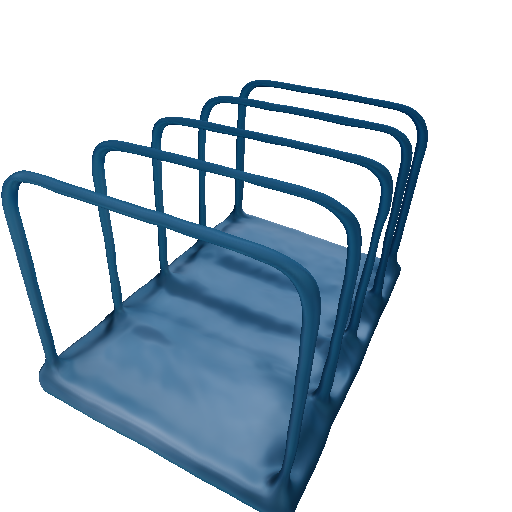}%
       \includegraphics[trim={0cm 0cm 1.5cm 1.5cm}, clip, width=0.48\linewidth]{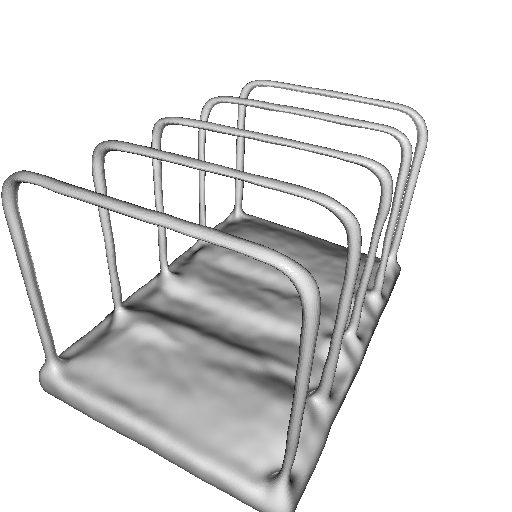} &
       \includegraphics[trim={0cm 0cm 1.5cm 1.5cm}, clip, width=0.48\linewidth]{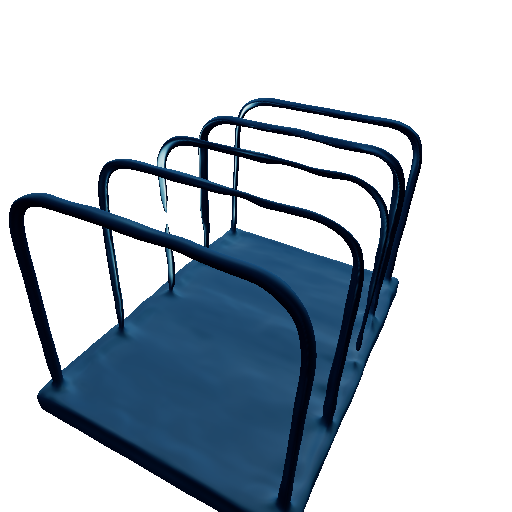}%
       \includegraphics[trim={0cm 0cm 1.5cm 1.5cm}, clip, width=0.48\linewidth]{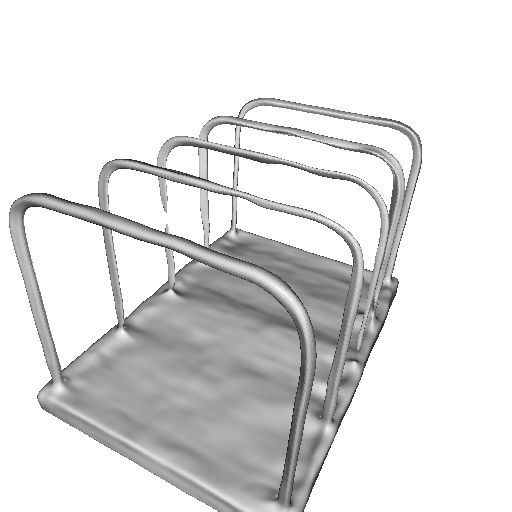} &
       \includegraphics[trim={0cm 0cm 1.5cm 1.5cm}, clip, width=0.48\linewidth]{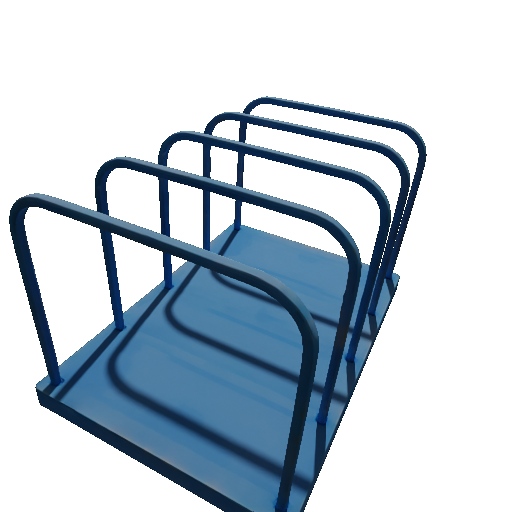}%
       \includegraphics[trim={0.cm 0cm 0cm 0cm}, clip, width=0.48\linewidth]{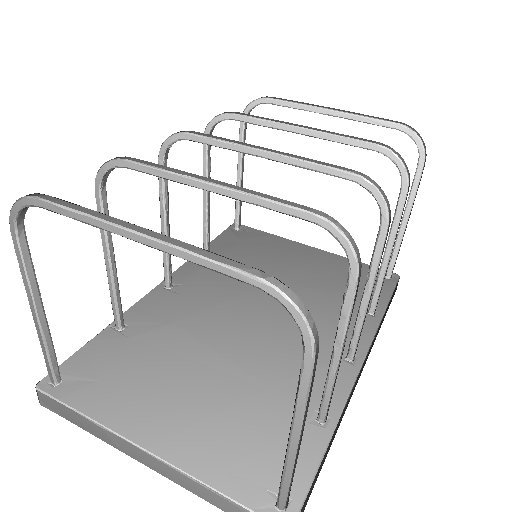} \\
       \includegraphics[trim={1.5cm 0cm 1.5cm 1.5cm}, clip,width=\linewidth]{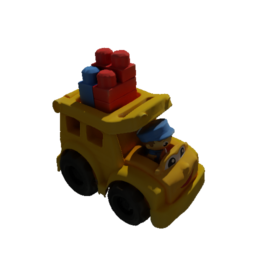}&\includegraphics[trim={3cm 0cm 1.5cm 2.5cm}, clip, width=0.48\linewidth]{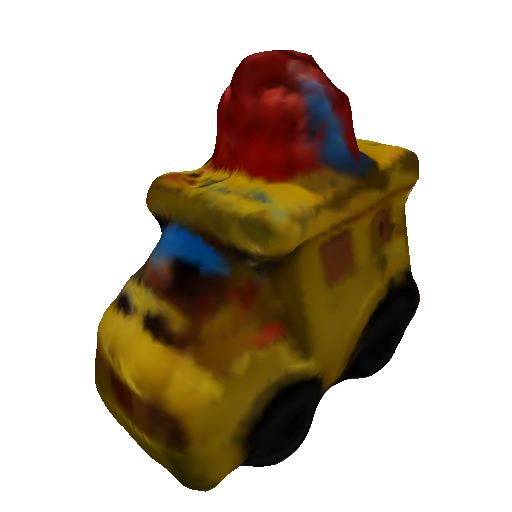}%
       \includegraphics[trim={3cm 0cm 1.5cm 2.5cm}, clip, width=0.48\linewidth]{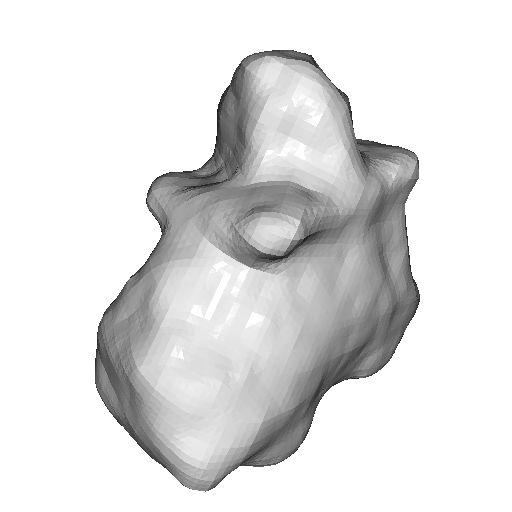} & 
       \includegraphics[trim={3cm 0cm 1.5cm 2.5cm}, clip, width=0.48\linewidth]{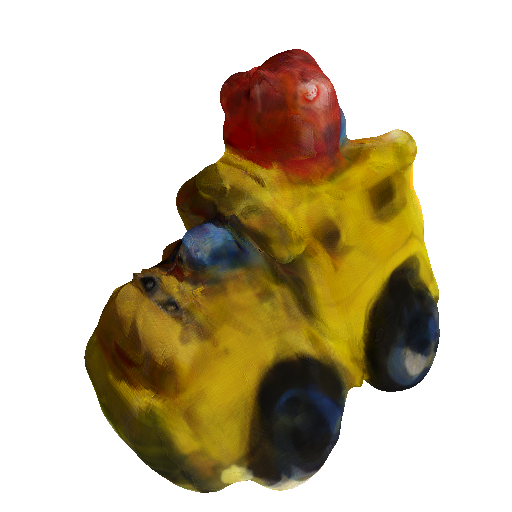}%
       \includegraphics[trim={3cm 0cm 1.5cm 2.5cm}, clip, width=0.48\linewidth]{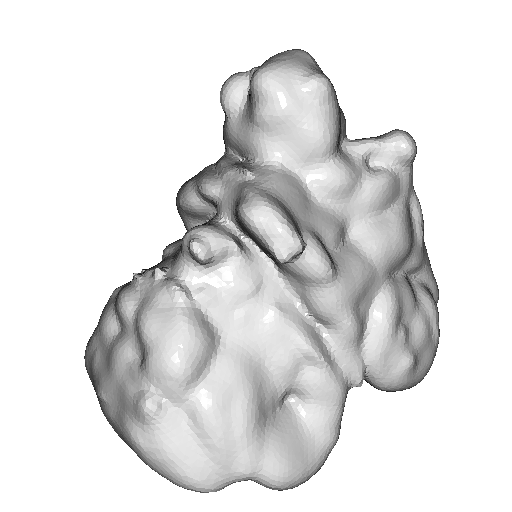} &
       \includegraphics[trim={3cm 0cm 1.5cm 2.5cm}, clip, width=0.48\linewidth]{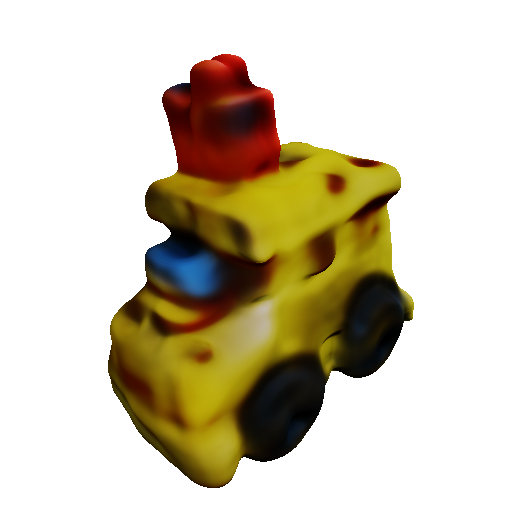}%
       \includegraphics[trim={3cm 0cm 1.5cm 2.5cm}, clip, width=0.48\linewidth]{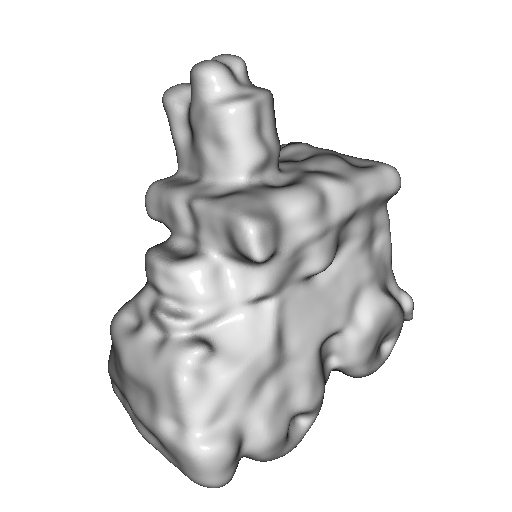} &
       \includegraphics[trim={3cm 0cm 1.5cm 2.5cm}, clip, width=0.48\linewidth]{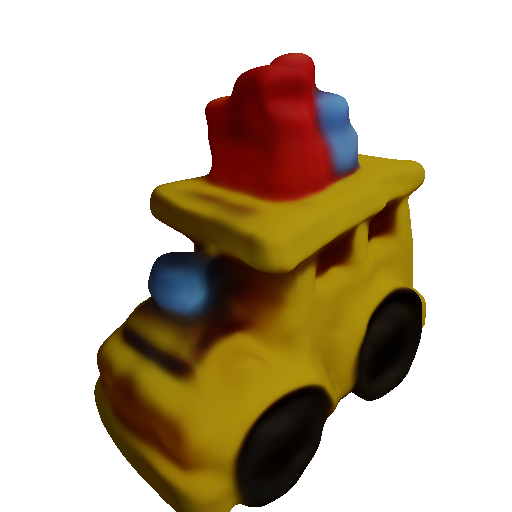}%
       \includegraphics[trim={3cm 0cm 1.5cm 2.5cm}, clip, width=0.48\linewidth]{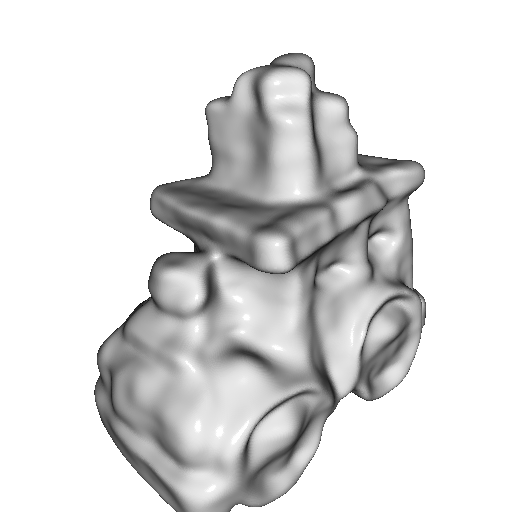} &
       \includegraphics[trim={3cm 0cm 1.5cm 2.5cm}, clip, width=0.48\linewidth]{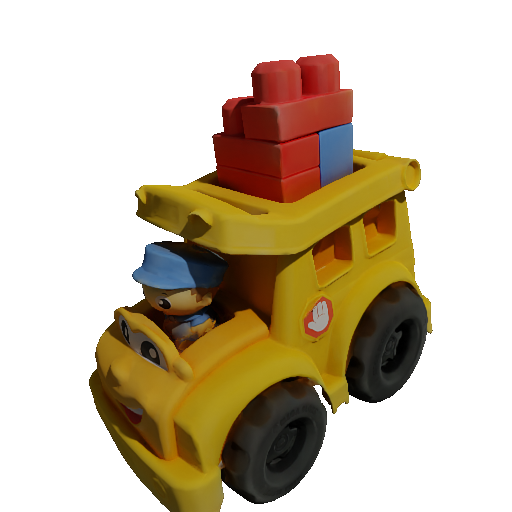}%
       \includegraphics[trim={3cm 0cm 1.5cm 2.5cm}, clip, width=0.48\linewidth]{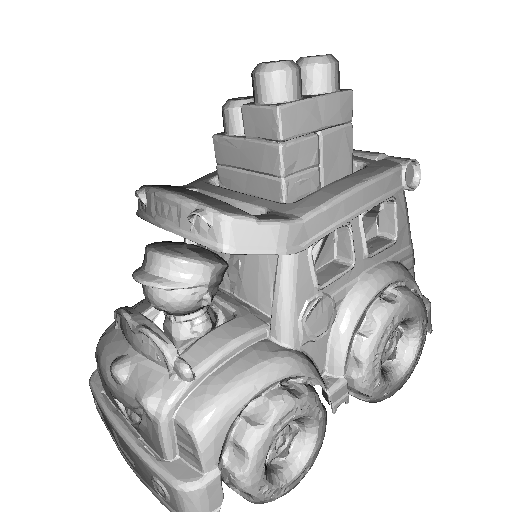} \\
       \includegraphics[trim={2cm 1cm 2cm 2cm}, clip,width=\linewidth]{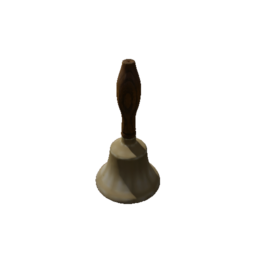}&\includegraphics[trim={4cm 0cm 2cm 4cm}, clip, width=0.48\linewidth]{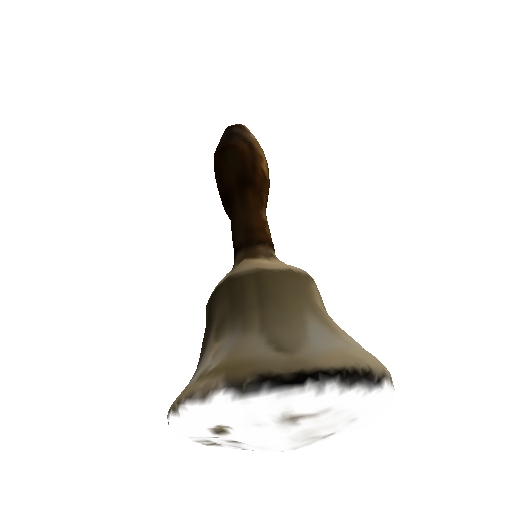}%
       \includegraphics[trim={4cm 0cm 2cm 4cm}, clip, width=0.48\linewidth]{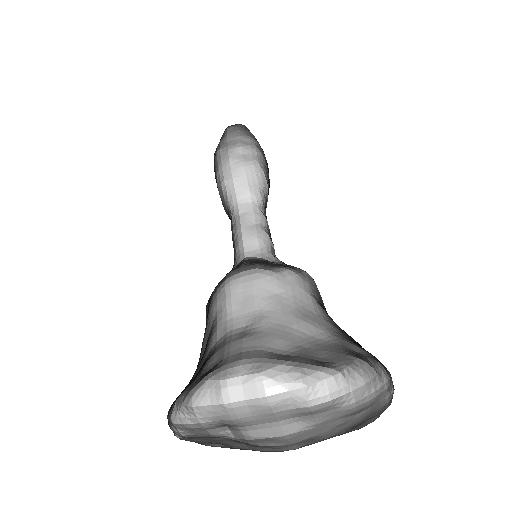} & 
       \includegraphics[trim={4cm 0cm 2cm 4cm}, clip, width=0.48\linewidth]{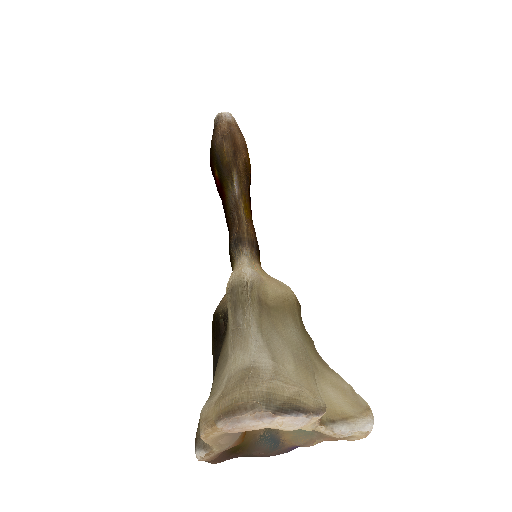}%
       \includegraphics[trim={4cm 0cm 2cm 4cm}, clip, width=0.48\linewidth]{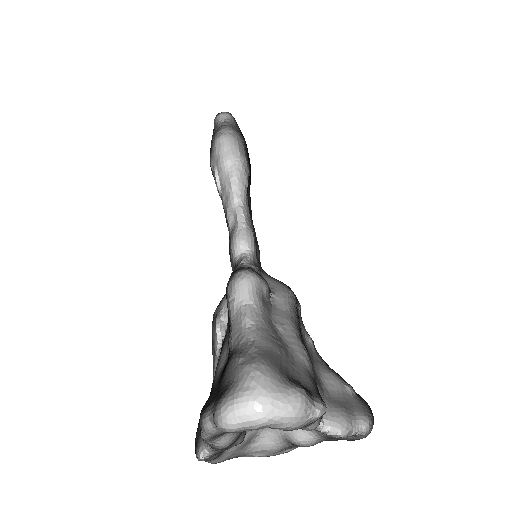} &
       \includegraphics[trim={4cm 0cm 2cm 4cm}, clip, width=0.48\linewidth]{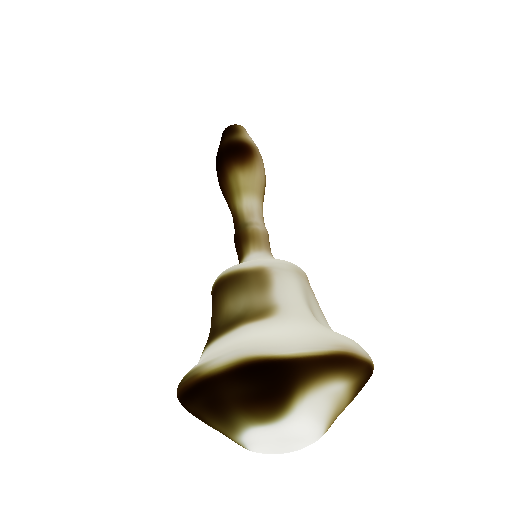}%
       \includegraphics[trim={4cm 0cm 2cm 4cm}, clip, width=0.48\linewidth]{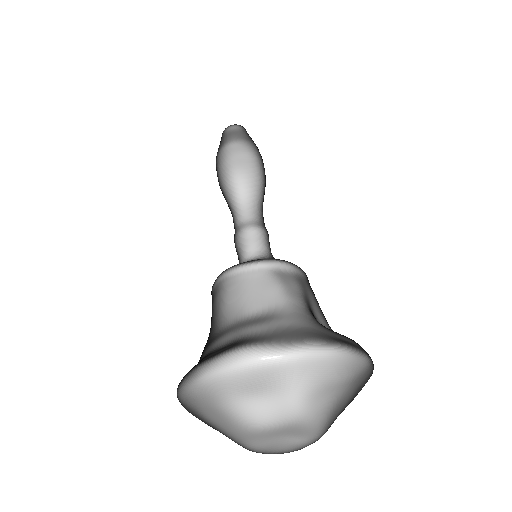} &
       \includegraphics[trim={4cm 0cm 2cm 4cm}, clip, width=0.48\linewidth]{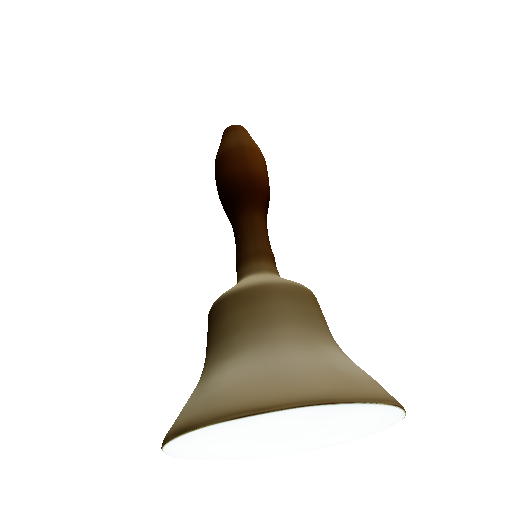}%
       \includegraphics[trim={4cm 0cm 2cm 4cm}, clip, width=0.48\linewidth]{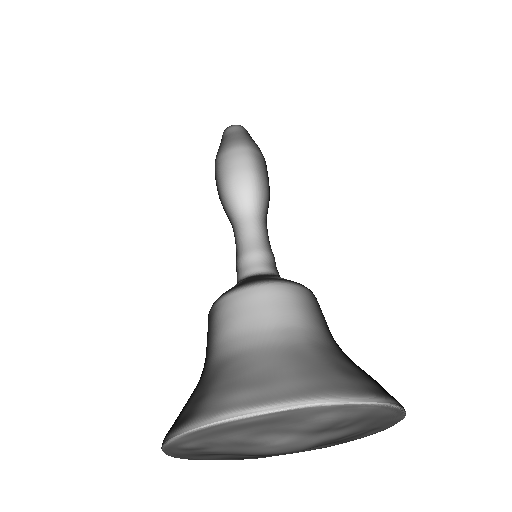} &
       \includegraphics[trim={4cm 0cm 2cm 4cm}, clip, width=0.48\linewidth]{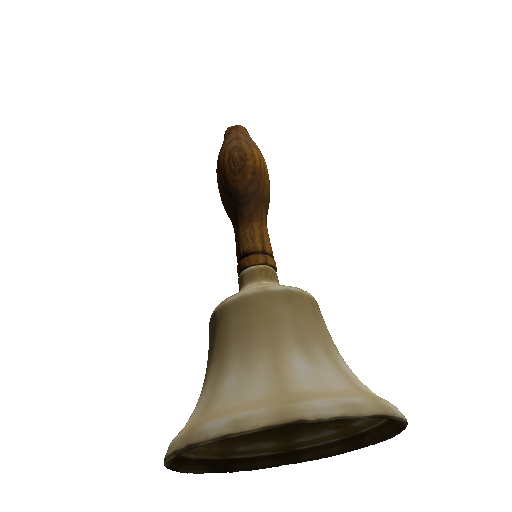}%
       \includegraphics[trim={4cm 0cm 2cm 4cm}, clip, width=0.48\linewidth]{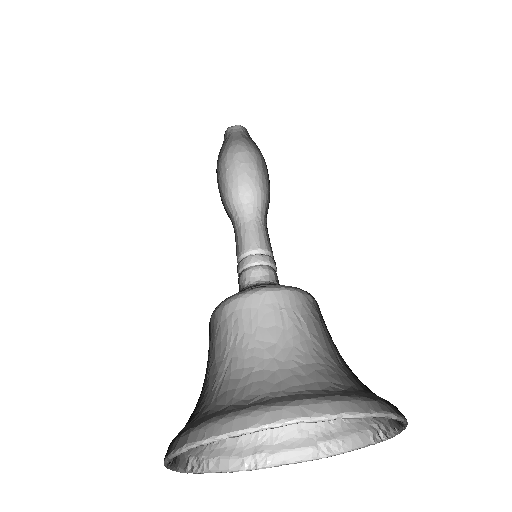} \\
       Reference & One-2-3-45-XL & DreamGaussian-XL & SyncDreamer & EscherNet  & Ground-Truth\\
         \bottomrule
       \end{tabular}
   \caption{{\bf Single view 3D reconstruction visualisation on GSO.} 
    EscherNet's ability to generate dense and consistent novel views significantly improves the reconstruction of complete and well-constrained 3D geometry. In contrast, One-2-3-45-XL and DreamGaussian-XL, despite leveraging a significantly larger pre-trained model, tend to produce over-smoothed and noisy reconstructions; SyncDreamer, constrained by sparse fixed-view synthesis, struggles to tightly constrain geometry, particularly in areas in sofa and the bottom part of the bell. }
   \label{tab:fig_3D}
   \vspace{-0.4cm}
\end{figure*}

\paragraph{Results} In Tab.~\ref{tab:3D} and Fig.~\ref{tab:fig_3D}, we show that EscherNet stands out by achieving significantly superior 3D reconstruction quality compared to other image-to-3D generative models. Specifically, EscherNet demonstrates an approximate 25\% improvement in Chamfer distance over SyncDreamer, considered as the current best model, when conditioned on a single reference view, and a 60\% improvement when conditioned on 10 reference views. This impressive performance is attributed to EscherNet's ability to flexibly handle any number of reference and target views, providing comprehensive and accurate constraints for 3D geometry. In contrast, SyncDreamer faces challenges due to sensitivity to elevation angles and constraints imposed by a fixed 30$^{\circ}$ elevation angle by design, thus hindering learning a holistic representation of complex objects. This limitation results in degraded reconstruction, particularly evident in the lower regions of the generated geometry.

\begin{table}[t!]
   \centering
   \scriptsize
 \renewcommand{\arraystretch}{0.6}
   \setlength{\tabcolsep}{0.7em}
   \begin{tabular}{lccc}
   \toprule
     &  \makecell{\# Ref. Views}  & \makecell{Chamfer Dist. $\downarrow$} & \makecell{Volume  IoU $\uparrow$} \\
    \midrule 
    Point-E &1 &0.0447 &0.2503\\
    Shape-E &1 &0.0448 &0.3762 \\
    One2345 &1 &0.0632 &0.4209 \\
    One2345-XL &1 & 0.0667 &0.4016 \\
    DreamGaussian & 1 &0.0605 &0.3757 \\
    DreamGaussian-XL & 1 &0.0459&0.4531 \\
    SyncDreamer & 1&0.0400 &0.5220 \\
    \midrule
    NeuS    &3 & 0.0366 &0.5352 \\
    NeuS    &5 & 0.0245 &0.6742 \\
    NeuS    &10 & 0.0195 &0.7264 \\
    \midrule
    EscherNet & 1 &0.0314 &0.5974 \\
    EscherNet & 2 &0.0215 &0.6868 \\
    EscherNet & 3 &0.0190 &0.7189 \\
    EscherNet & 5 &0.0175 &0.7423 \\
    EscherNet & 10 &0.0167&0.7478 \\
     \bottomrule
   \end{tabular}
   \caption{{\bf 3D reconstruction performance on GSO.} EscherNet outperforms all other image-to-3D baselines in generating more visually appealing with accurate 3D geometry, particularly when conditioned on multiple reference views. }
   \label{tab:3D}
   \vspace{-0.6cm}
\end{table}

\subsection{Results on Text-to-3D Generation}

\begin{figure}[ht!]
\centering
\footnotesize
 \renewcommand{\arraystretch}{0.4}
\setlength{\tabcolsep}{0.0em}
\begin{tabular}{*{6}{C{0.165\linewidth}}}
\multicolumn{3}{C{0.5\linewidth}}{\makecell{\small \it A bald eagle carved\\ \small\it out of wood. $\Rightarrow$} } & \multicolumn{3}{R{0.5\linewidth}}{\includegraphics[trim={0.5cm 0.5cm 0.5cm 0.5cm}, clip, width=\linewidth]{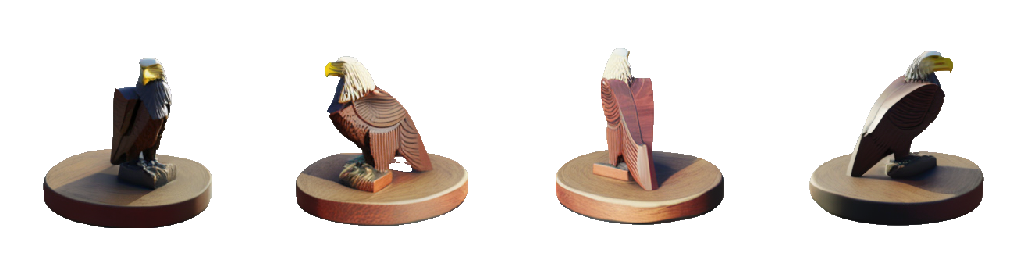}} \\
\includegraphics[trim={0.5cm 0.5cm 0.5cm 0.5cm}, clip, width=\linewidth]{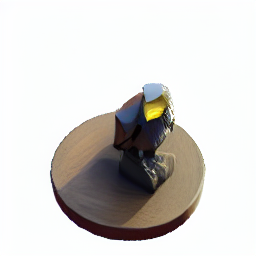}   & 
\includegraphics[trim={0.5cm 0.5cm 0.5cm 0.5cm}, clip, width=\linewidth]{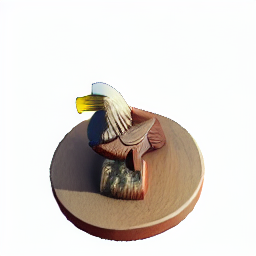}    & \includegraphics[trim={0.5cm 0.5cm 0.5cm 0.5cm}, clip, width=\linewidth]{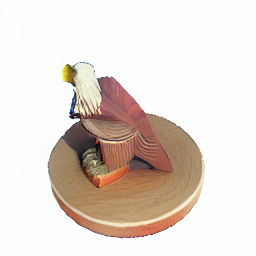}    & \includegraphics[trim={0.5cm 0.5cm 0.5cm 0.5cm}, clip, width=\linewidth]{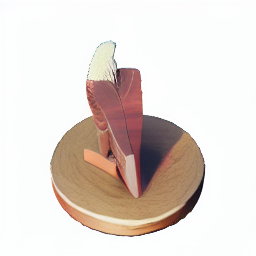}    & \includegraphics[trim={0.5cm 0.5cm 0.5cm 0.5cm}, clip, width=\linewidth]{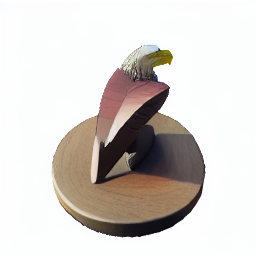}    & 
\includegraphics[trim={0.5cm 0.5cm 0.5cm 0.5cm}, clip, width=\linewidth]{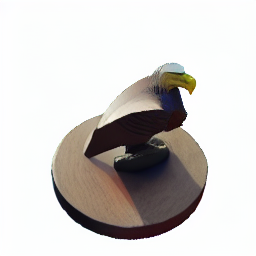}   \\
\includegraphics[trim={0.5cm 0.5cm 0.5cm 0.5cm}, clip, width=\linewidth]{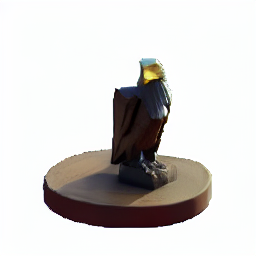}    & \includegraphics[trim={0.5cm 0.5cm 0.5cm 0.5cm}, clip, width=\linewidth]{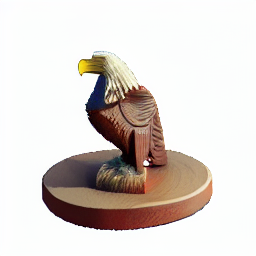}    & \includegraphics[trim={0.5cm 0.5cm 0.5cm 0.5cm}, clip, width=\linewidth]{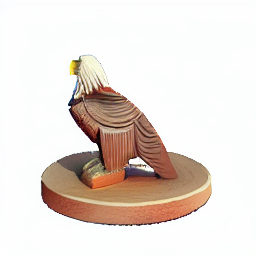}    & \includegraphics[trim={0.5cm 0.5cm 0.5cm 0.5cm}, clip, width=\linewidth]{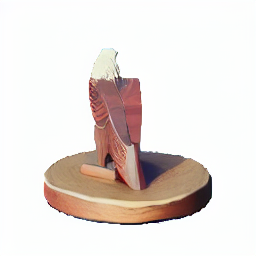}    & \includegraphics[trim={0.5cm 0.5cm 0.5cm 0.5cm}, clip, width=\linewidth]{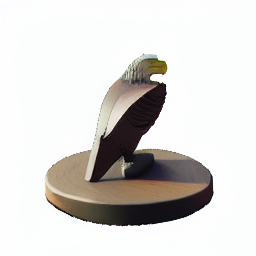}    & 
\includegraphics[trim={0.5cm 0.5cm 0.5cm 0.5cm}, clip, width=\linewidth]{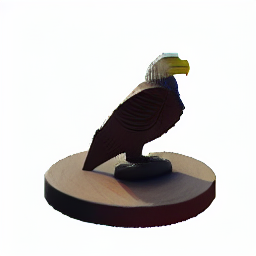}   \\
\includegraphics[trim={0.5cm 0.5cm 0.5cm 0.5cm}, clip, width=\linewidth]{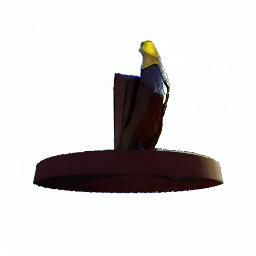}    & \includegraphics[trim={0.5cm 0.5cm 0.5cm 0.5cm}, clip, width=\linewidth]{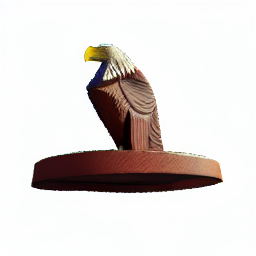}    & \includegraphics[width=\linewidth]{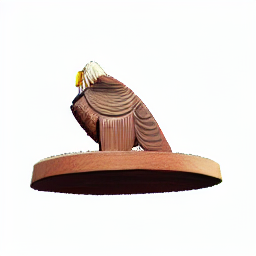}    & \includegraphics[trim={0.5cm 0.5cm 0.5cm 0.5cm}, clip, width=\linewidth]{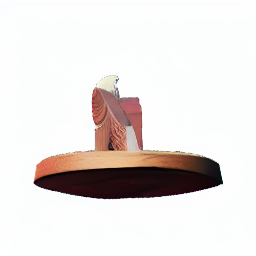}    & \includegraphics[trim={0.5cm 0.5cm 0.5cm 0.5cm}, clip, width=\linewidth]{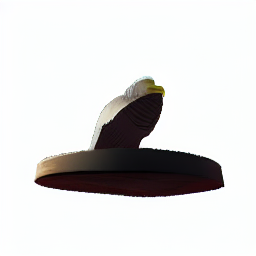}    & 
\includegraphics[trim={0.5cm 0.5cm 0.5cm 0.5cm}, clip, width=\linewidth]{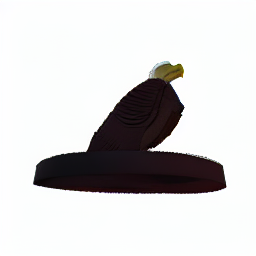}   \\
\midrule
\multicolumn{4}{C{0.6\linewidth}}{\makecell{\small \it A robot made of vegetables, 4K. $\Rightarrow$} } & \multicolumn{2}{C{0.15\linewidth}}{\includegraphics[width=\linewidth]{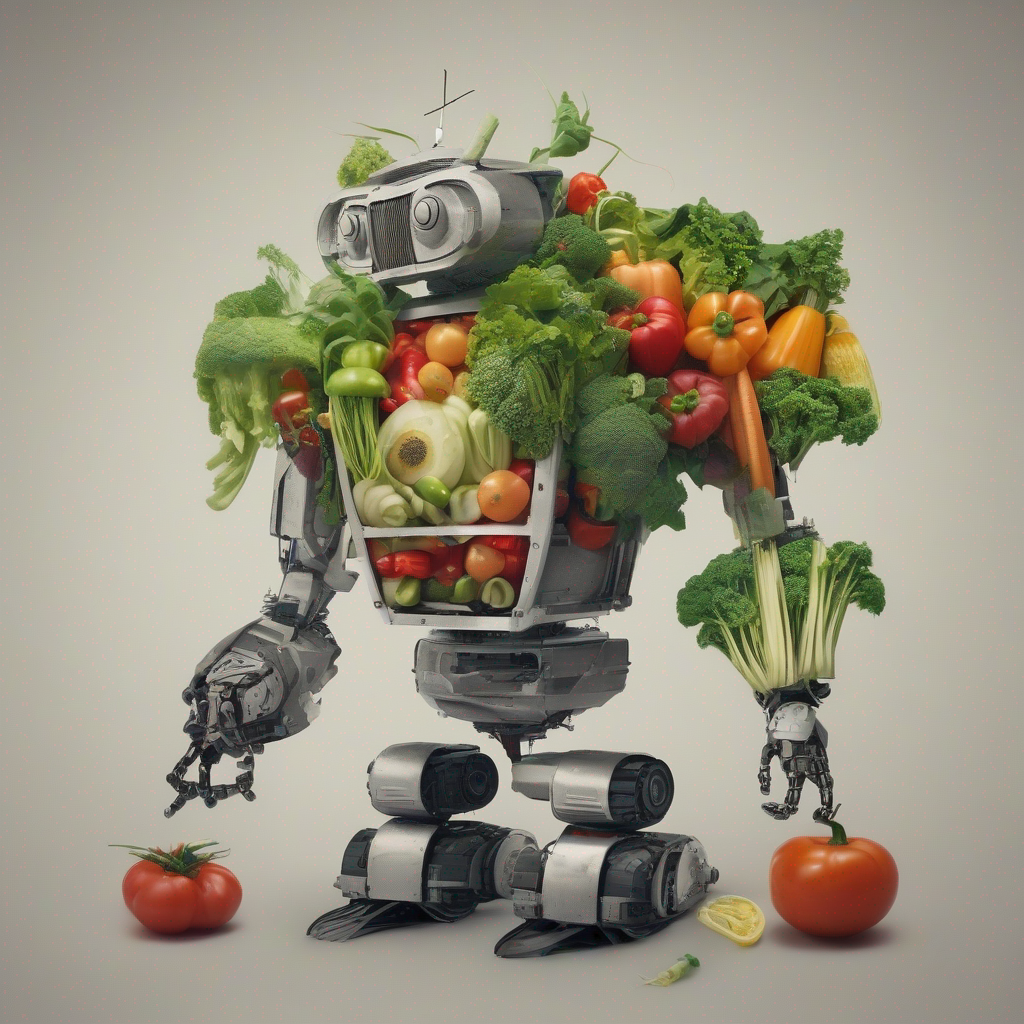}} \\
\includegraphics[trim={1cm 1cm 1cm 1cm}, clip, width=\linewidth]{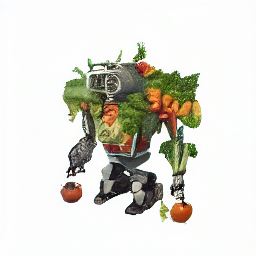}    & \includegraphics[trim={1cm 1cm 1cm 1cm}, clip, width=\linewidth]{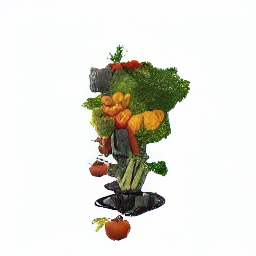}    & \includegraphics[trim={1cm 1cm 1cm 1cm}, clip, width=\linewidth]{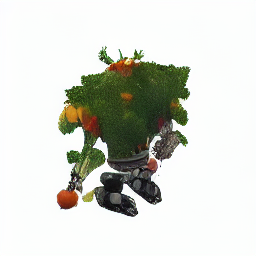}    & \includegraphics[trim={1cm 1cm 1cm 1cm}, clip, width=\linewidth]{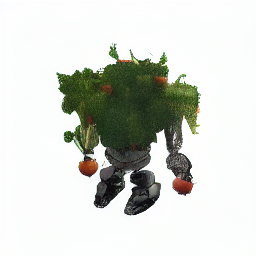}    & \includegraphics[trim={1cm 1cm 1cm 1cm}, clip, width=\linewidth]{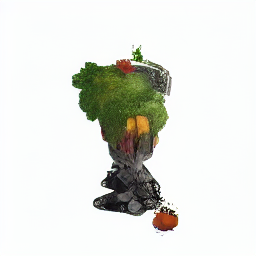}    & 
\includegraphics[trim={1cm 1cm 1cm 1cm}, clip, width=\linewidth]{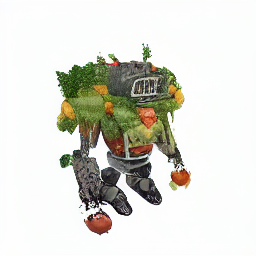}   \\
\includegraphics[trim={1cm 1cm 1cm 1cm}, clip, width=\linewidth]{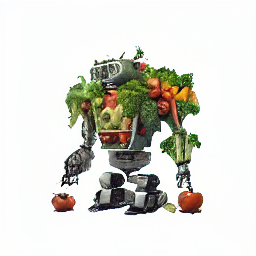}    & \includegraphics[trim={1cm 1cm 1cm 1cm}, clip, width=\linewidth]{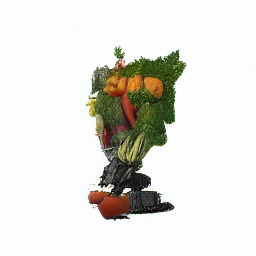}    & \includegraphics[trim={1cm 1cm 1cm 1cm}, clip, width=\linewidth]{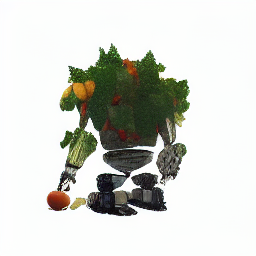}    & \includegraphics[trim={1cm 1cm 1cm 1cm}, clip, width=\linewidth]{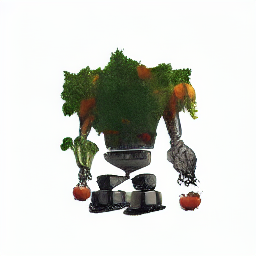}    & \includegraphics[trim={1cm 1cm 1cm 1cm}, clip, width=\linewidth]{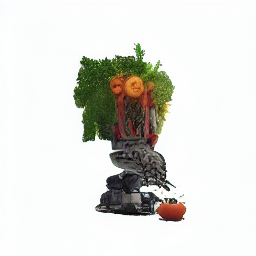}    & 
\includegraphics[trim={1cm 1cm 1cm 1cm}, clip, width=\linewidth]{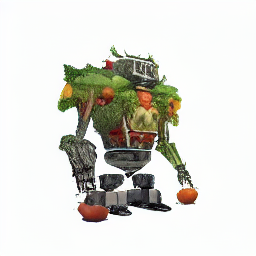}   \\
\includegraphics[trim={1cm 1cm 1cm 1cm}, clip, width=\linewidth]{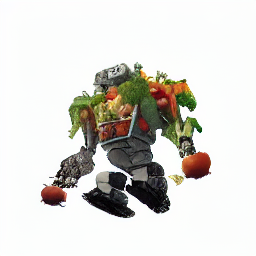}    & \includegraphics[trim={1cm 1cm 1cm 1cm}, clip, width=\linewidth]{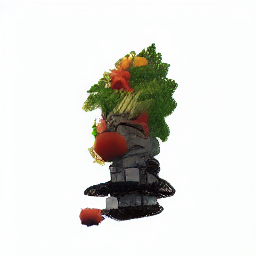}    & \includegraphics[trim={1cm 1cm 1cm 1cm}, clip, width=\linewidth]{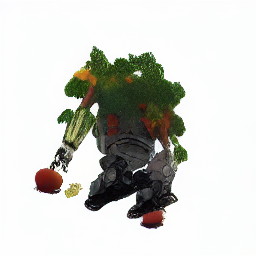}    & \includegraphics[trim={1cm 1cm 1cm 1cm}, clip, width=\linewidth]{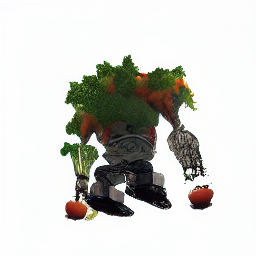}    & \includegraphics[trim={1cm 1cm 1cm 1cm}, clip, width=\linewidth]{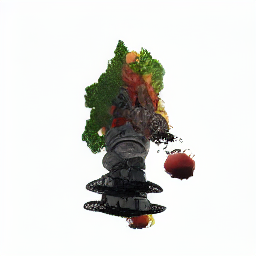}    & 
\includegraphics[trim={1cm 1cm 1cm 1cm}, clip, trim={1cm 1cm 1cm 1cm}, clip, width=\linewidth]{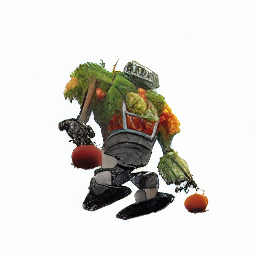}   \\
\end{tabular}
\caption{{\bf Text-to-3D visualisation with MVDream (up) and SDXL (bottom).} EscherNet offers compelling and realistic view synthesis for synthetic images generated with user-provided text prompts. Additional results are shown in Appendix~\ref{app:text23d}. }
\label{tab:fig_mvdream}
\vspace{-0.4cm}
\end{figure}

EscherNet's flexibility in accommodating any number of reference views enables a straightforward approach to the text-to-3D generation problem by breaking it down into two stages: text-to-image, relying on any off-the-shelf text-to-image generative model, and then image-to-3D, relying on EscherNet. In Fig.~\ref{tab:fig_mvdream}, we present visual results of dense novel view generation using a text-to-4view model with MVDream~\cite{shi2023mvdream} and a text-to-image model with SDXL~\cite{podell2023sdxl}. Remarkably, even when dealing with out-of-distribution and counterfactual content, EscherNet generates consistent 3D novel views with appealing textures.

\section{Conclusions}
In this paper, we have introduced EscherNet, a multi-view conditioned diffusion model designed for scalable view synthesis. Leveraging Stable Diffusion's 2D architecture empowered by the innovative Camera Positional Embedding (CaPE), EscherNet adeptly learns implicit 3D representations from varying number of reference views, achieving consistent 3D novel view synthesis. We provide detailed discussions and additional ablative analysis in Appendix~\ref{app:discussions}.

\paragraph{Limitations and Discussions} 
EscherNet's flexibility in handling any number of reference views allows for autoregressive generation, similar to autoregressive language models~\cite{brown2020gpt3, chowdhery2022palm}. While this approach significantly reduces inference time, it leads to a degraded generation quality. Additionally, EscherNet's current capability operates within a 3 DoF setting constrained by its training dataset, which may not align with real-world scenarios, where views typically span in $SE(3)$ space. Future work will explore scaling EscherNet with 6 DoF training data with real-world scenes, striving for a more general 3D representation.

\section*{Acknowledgement}
This research is funded by EPSRC Prosperity Partnerships (EP/S036636/1) and Dyson Technology Ltd.
Xin Kong holds a China Scholarship Council-Imperial Scholarship. We would like to thank Sayak Paul and HuggingFace for contributing the training compute that facilitated early project exploration. We would also like to acknowledge Yifei Ren for his valuable discussions on formulating the 6DoF CaPE.

{\small
    \bibliographystyle{ieeenat_fullname}
    \bibliography{reference}
}
\newpage

\onecolumn
\appendix
\section*{\begin{center}\Large \bf EscherNet: A Generative Model for Scalable View Synthesis \\ (Appendix)\end{center}}

% \section*{Acknowledgement}
% This research is funded by EPSRC Prosperity Partnerships (EP/S036636/1) and Dyson Technology Ltd.
% Xin Kong holds a China Scholarship Council-Imperial Scholarship. We would like to thank Sayak Paul and HuggingFace for contributing the training compute that facilitated early project exploration. We would also like to acknowledge Yifei Ren for his valuable discussions on formulating the 6DoF CaPE.
% \vspace{10mm}

\section{Python Implementation of CaPE}
\label{app:code}

\begin{lstlisting}[language=Python, caption=Python implementation for 4 DoF CaPE.]
def compute_4dof_cape(v, P, s):
    """
    :param v: input feature vector with its dimension must be divisible by 8
    :param P: list = [alpha, beta, gamma, r]
    :param s: a small scalar for radius
    :return: rotated v with its corresponding camera pose P
    """
    v = v.reshape([-1, 8])
    psi = np.zeros([8, 8])
    for i in range(4):
        if i < 3:
            psi[2 * i:2 * (i + 1), 2 * i:2 * (i + 1)] = \
                np.array([[np.cos(P[i]), -np.sin(P[i])], [np.sin(P[i]), np.cos(P[i])]])
        else:
            psi[2 * i:2 * (i + 1), 2 * i:2 * (i + 1)] = \
                np.array([[np.cos(s * np.log(P[i])), -np.sin(s * np.log(P[i]))],
                          [np.sin(s * np.log(P[i])), np.cos(s * np.log(P[i]))]])

    return v.dot(psi).reshape(-1)
\end{lstlisting}

\begin{lstlisting}[language=Python, caption=Python implementation for 6 DoF CaPE.]
def compute_6dof_cape(v, P, s=0.001, key=True):
    """
    :param v: input feature vector with its dimension must be divisible by 4
    :param P: 4 x 4 SE3 matrix
    :param s: a small scalar for translation
    :return: rotated v with its corresponding camera pose P
    """
    v = v.reshape([-1, 4])
    P[:3, 3] *= s
    psi = P if key else np.linalg.inv(P).T
    return v.dot(psi).reshape(-1)
\end{lstlisting}
\newpage

\section{Additional Training Details and Experimental Settings}
\label{app:training}

\paragraph{Optimisation and Implementation}  EscherNet is trained using the AdamW optimiser~\cite{loshchilov2018adamw} with a learning rate of $1\cdot 10^{-4}$ and weight decay of $0.01$ for $[256\times 256]$ resolution images. We incorporate cosine annealing, reducing the learning rate to $1\cdot 10^{-5}$ over a total of 100,000 training steps, while linearly warming up for the initial 1000 steps. To speed up training, we implement automatic mixed precision with a precision of {\tt bf16} and employ gradient checkpointing. Our training batches consist of 3 reference views and 3 target views randomly sampled with replacement from 12 views for each object, with a total batch size of 672 (112 batches per GPU). The entire model training process takes 1 week on 6 NVIDIA A100 GPUs.

\paragraph{Metrics} 
For 2D metrics used in view synthesis, we employ PSNR, SSIM~\cite{wang2004image}, LPIPS~\cite{zhang2018unreasonable}. For 3D metrics used in 3D generation, we employ Chamfer Distance and Volume IoU. To ensure a fair and efficient evaluation process, each baseline method and our approach are executed only once per scene per viewpoint. This practice has proven to provide stable averaged results across multiple scenes and viewpoints.

\subsection{Evaluation Details}
\paragraph{In NeRF Synthetic Dataset~\cite{mildenhall2020nerf},} we consider and evaluate all 8 scenes provided in the original dataset. To assess performance with varying numbers of reference views, we train all baseline methods and our approach using the same set of views randomly sampled from the training set. The evaluation is conducted on all target views defined in the test sets across all 8 scenes (with 200 views per scene). For InstantNGP~\cite{muller2022instant}, we run 10k steps ($\approx$ 1min) for each scene. For 3D Gaussian Splatting~\cite{kerbl20233gaussian}, we run 5k steps ($\approx$ 2min) for each scene.

\paragraph{In Google Scanned Dataset (GSO)~\cite{downs2022google},} we evaluate the same 30 objects chosen by SyncDreamer~\cite{liu2023syncdreamer}. For each object, we render 25 views with randomly generated camera poses and a randomly generated environment lighting condition to construct our test set. For each object, we choose the first 10 images as our reference views and the subsequent 15 images as our target views for evaluation. It's crucial to note that all reference and target views are rendered with random camera poses, establishing a more realistic and challenging evaluation setting compared to the evaluation setups employed in other baselines: {\it e.g.} SyncDreamer uses an evenly distributed environment lighting to render all GSO data, and the reference view for each object is manually selected based on human preference.\footnote{\url{https://github.com/liuyuan-pal/SyncDreamer/issues/21}} Additionally, the evaluated target view is also manually selected based on human preference chosen among four independent generations.\footnote{\url{https://github.com/liuyuan-pal/SyncDreamer/issues/21\#issuecomment-1770345260}}

In evaluating 3D generation, we randomly sample 4096 points evenly distributed from the generated 3D mesh or point cloud across all methods. Each method's generated mesh is aligned to the ground-truth mesh using the camera pose of the reference views. Specifically in Point-E~\cite{nichol2022pointe} and Shape-E~\cite{jun2023shape}, we rotate 90/180 degrees along each x/y/z axis to determine the optimal alignment for the final mesh pose. Our evaluation approach again differs from SyncDreamer, which initially projects the 3D mesh into their fixed 16 generated views to obtain depth maps. Then, points are sampled from these depth maps for the final evaluation.\footnote{\url{https://github.com/liuyuan-pal/SyncDreamer/issues/44}}

\paragraph{In RTMV Dataset~\cite{tremblay2022rtmv},} we follow the evaluation setting used in Zero-1-to-3~\cite{liu2023zero}, which consists of 10 complex scenes featuring a pile of multiple objects from the GSO dataset. Similar to the construction of our GSO test set, we then randomly select a fixed subset of the first 10 images as our reference views and the subsequent 10 views as our target views for evaluation.

\newpage
\section{Additional Results on 6 DoF CaPE}
\label{app:6dof}

To validate the effectiveness of the 6 DoF CaPE design, we demonstrate its performance in novel view synthesis on GSO and RTMV datasets in Tab.~\ref{tab:NVS_6DoF} and on the NeRF Synthetic dataset in Tab.~\ref{tab:nerf_6DoF}. We also provide 3D reconstruction results on GSO dataset in Tab.~\ref{tab:3D_6DoF}. It is evident that EscherNet with 6 DoF CaPE achieves comparable, and often, slightly improved results when compared to our 4 DoF CaPE design.

\begin{table}[ht!]
   \centering
   \scriptsize
   \setlength{\tabcolsep}{0.25em}
  \begin{subtable}{0.49\linewidth}
       \begin{tabular}{lcccccccccc}
       \toprule
       &  \multirow{2}[2]{*}{\makecell[c]{Training \\ Data}} & \multirow{2}[2]{*}{\makecell{\# Ref.\\ Views}} & \multicolumn{3}{c}{{\bf GSO-30}}  & \multicolumn{3}{c}{{\bf RTMV}} \\
       \cmidrule(lr){4-6}      \cmidrule(lr){7-9}
          & &  &PSNR$\uparrow$ & SSIM$\uparrow$ & LPIPS$\downarrow$  &PSNR$\uparrow$ & SSIM$\uparrow$ & LPIPS$\downarrow$  \\
        \midrule 
        RealFusion  & -    & 1 & 12.76 &0.758 &0.382 &-&-&- \\
        Zero123  &  800K  &1  &18.51 &0.856 &0.127  &10.16 &0.505 &0.418\\
        Zero123-XL & 10M  &1 &18.93 &0.856 &0.124 &10.59 &0.520 &0.401\\
        \midrule
         EscherNet - 4 DoF     & 800k  &1  &20.24 &0.884 &0.095 &10.56 &0.518 &0.410\\
         EscherNet - 4 DoF    & 800k  &2  &22.91 &0.908 &0.064 &12.66 &0.585 &0.301\\
         EscherNet - 4 DoF     & 800k  &3  &24.09 &0.918 &0.052 &13.59 &0.611 &0.258\\
         EscherNet - 4 DoF    & 800k &5  &25.09 &0.927 &0.043 &14.52 &0.633 &0.222\\
         EscherNet - 4 DoF   & 800k &10   &25.90 &0.935 &0.036 &15.55 &0.657 &0.185\\
        \midrule
         EscherNet - 6 DoF     & 800k  &1  &20.89 &0.886 &0.093 &12.30 &0.569 &0.332\\
        EscherNet - 6 DoF   & 800k  &2  &23.92 &0.917 &0.057 &14.18 &0.618 &0.252 \\
        EscherNet - 6 DoF    & 800k  &3  &25.21 &0.927 &0.045 &15.06 &0.643 &0.217 \\
        EscherNet - 6 DoF   & 800k &5  &26.59 &0.937 &0.036 &15.71 &0.663 &0.190\\
        EscherNet - 6 DoF  & 800k &10   &27.75 &0.947 &0.030 &16.58 &0.688 &0.160\\
         \bottomrule
       \end{tabular}
   \caption{{\bf Novel view synthesis performance on GSO and RTMV datasets.}}
   \label{tab:NVS_6DoF}
   \end{subtable}\hfill
     \begin{subtable}{0.49\linewidth}
     \centering
   \setlength{\tabcolsep}{0.5em}
  \begin{tabular}{lccc}
   \toprule
 &  \makecell{\# Ref. Views}  & \makecell{Chamfer Dist. $\downarrow$} & \makecell{Volume  IoU $\uparrow$} \\
    \midrule 
    Point-E &1 &0.0447 &0.2503\\
    Shape-E &1 &0.0448 &0.3762 \\
    One2345 &1 &0.0632 &0.4209 \\
    One2345-XL &1 & 0.0667 &0.4016 \\
    DreamGaussian & 1 &0.0605 &0.3757 \\
    DreamGaussian-XL & 1 &0.0459&0.4531 \\
    SyncDreamer & 1&0.0400 &0.5220 \\
    \midrule
    NeuS    &3 & 0.0366 &0.5352 \\
    NeuS    &5 & 0.0245 &0.6742 \\
    NeuS    &10 & 0.0195 &0.7264 \\
    \midrule
    EscherNet - 4 DoF & 1&0.0314 &0.5974 \\
    EscherNet - 4 DoF& 2 &0.0215 &0.6868 \\
    EscherNet - 4 DoF& 3 &0.0190 &0.7189 \\
    EscherNet - 4 DoF& 5 &0.0175 &0.7423 \\
    EscherNet - 4 DoF& 10 &0.0167&0.7478 \\
    \midrule
    EscherNet - 6 DoF & 1 &0.0274 &0.6382 \\
    EscherNet - 6 DoF & 2 &0.0196 &0.7100 \\
    EscherNet - 6 DoF & 3 &0.0180 &0.7348 \\
    EscherNet - 6 DoF & 5 &0.0176 &0.7392 \\
    EscherNet - 6 DoF & 10 &0.0160&0.7628 \\
     \bottomrule
   \end{tabular}
   \caption{{\bf 3D reconstruction performance on GSO.} }
   \label{tab:3D_6DoF}
    \end{subtable}\\
    \begin{subtable}{0.49\linewidth}
    \centering
    \setlength{\tabcolsep}{0.45em}
       \begin{tabular}{lcccccccc}
       \toprule
        \multicolumn{9}{c}{\# Reference Views (Less $\to$ More)} \\
         & 1 & 2 & 3 & 5 & 10 & 20 & 50 & 100 \\
        \midrule 
        \multicolumn{9}{l}{{\bf InstantNGP (Scene Specific Training)}} \\
         PSNR$\uparrow$ & 10.92 & 12.42& 14.27& 18.17& 22.96& 24.99& 26.86& 27.30 \\
        SSIM$\uparrow$  & 0.449 & 0.521& 0.618& 0.761& 0.881& 0.917& 0.946& 0.953\\
        LPIPS$\downarrow$  & 0.627 & 0.499& 0.391& 0.228& 0.091& 0.058& 0.034& 0.031 \\
        \midrule 
        \multicolumn{9}{l}{{\bf GaussianSplatting (Scene Specific Training)}} \\
         PSNR$\uparrow$ & 9.44 & 10.78& 12.87& 17.09& 23.04& 25.34& 26.98& 27.11 \\
        SSIM$\uparrow$  & 0.391 & 0.432& 0.546& 0.732& 0.876& 0.919& 0.942& 0.944\\
        LPIPS$\downarrow$  & 0.610 & 0.541& 0.441& 0.243& 0.085& 0.054& 0.041& 0.041 \\
        \midrule 
        \multicolumn{9}{l}{{\bf EscherNet - 4 DoF (Zero Shot Inference)}} \\
        PSNR$\uparrow$  & 13.36 & 14.95& 16.19& 17.16& 17.74& 17.91& 18.05& 18.15 \\
        SSIM$\uparrow$   & 0.659 & 0.700& 0.729& 0.748& 0.761& 0.765& 0.769& 0.771\\
        LPIPS$\downarrow$  & 0.291 & 0.208& 0.161& 0.127& 0.114& 0.106& 0.099& 0.097\\
        \midrule 
        \multicolumn{9}{l}{{\bf EscherNet - 6 DoF (Zero Shot Inference)}} \\
        PSNR$\uparrow$  & 13.73 & 15.66& 16.91& 17.72& 18.47& 18.77& 19.24& 19.28 \\
        SSIM$\uparrow$  & 0.664 & 0.712& 0.745& 0.762& 0.779& 0.786& 0.795& 0.796\\
        LPIPS$\downarrow$ & 0.294 & 0.197& 0.149&0.120& 0.103& 0.095& 0.085& 0.084\\
        \bottomrule
   \end{tabular}
   \caption{{\bf Novel view synthesis performance on NeRF Synthetic dataset.}}
   \label{tab:nerf_6DoF}
   \end{subtable}
   \caption{EscherNet 6 DoF presents a similar and sometimes improved performance than EscherNet 4 DoF.}
\end{table}

\newpage
\section{Additional Results on NeRF Synthetic Dataset}
\label{app:nerf}

We present additional visualisation on the NeRF Synthetic Dataset using EscherNet trained with 4 DoF CaPE. 

\begin{table}[ht!]
   \centering
   \footnotesize
   \setlength{\tabcolsep}{0em}
       \begin{tabular}{cccccccc}
       \toprule
        \multicolumn{8}{c}{\# Reference Views (Less $\to$ More)} \\
         1 & 2 & 3 & 5 & 10 & 20 & 50 & 100 \\
        \midrule 
        \multicolumn{8}{l}{{\bf InstantNGP (Scene Specific Training)}} \\
        \makecell{\includegraphics[width=0.125\textwidth]{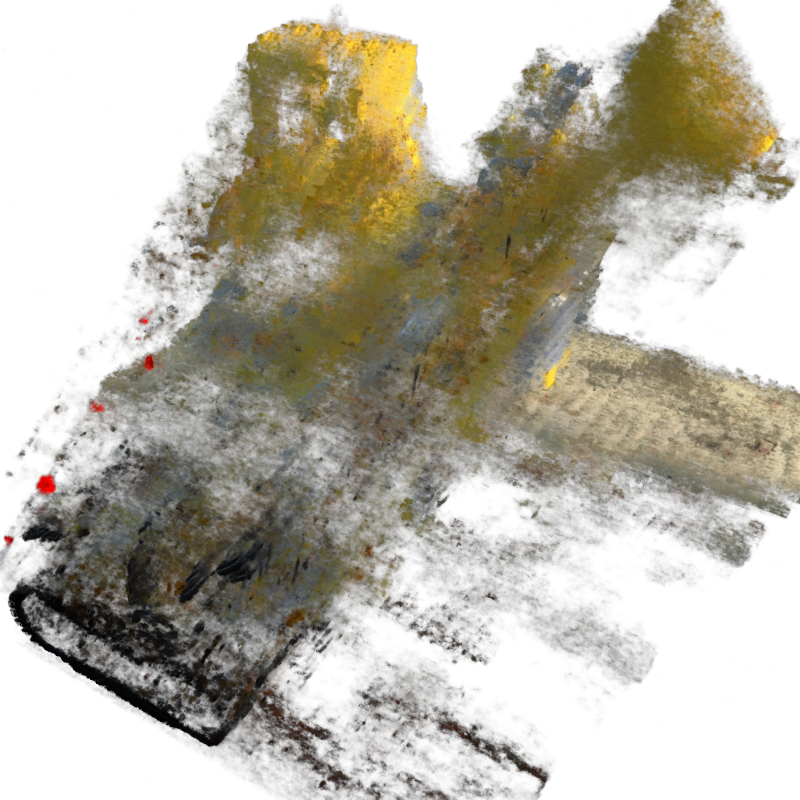} \\ PSNR 9.45} & \makecell{\includegraphics[width=0.125\textwidth]{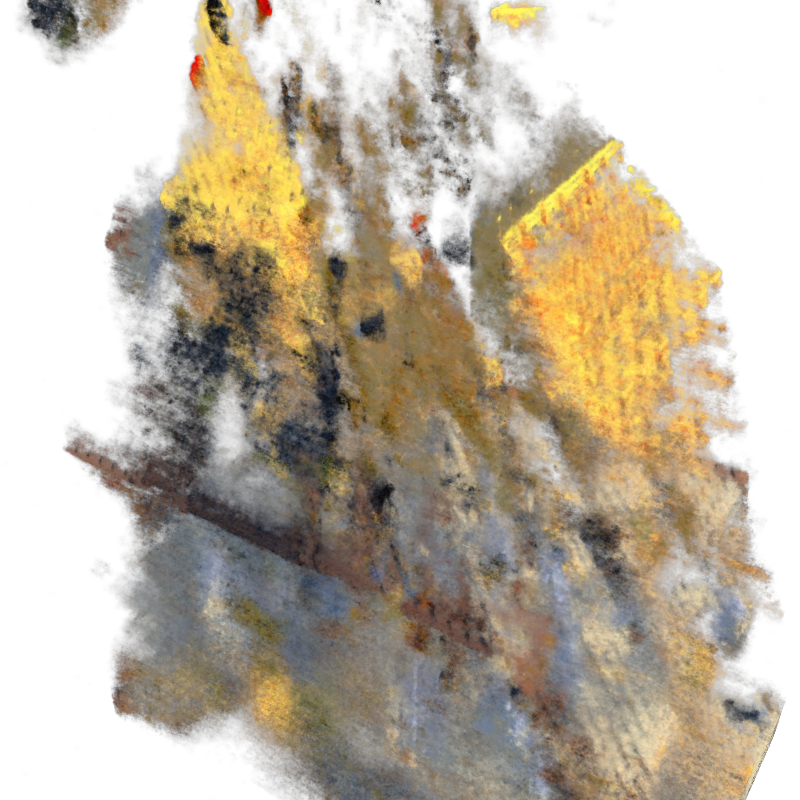} \\ PSNR 11.41} & \makecell{\includegraphics[width=0.125\textwidth]{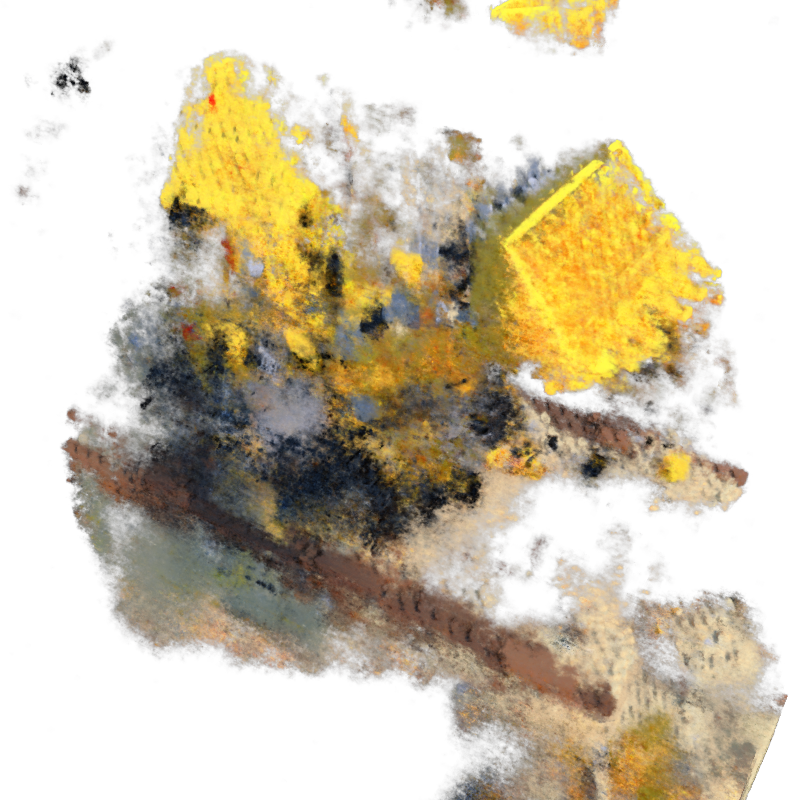} \\ PSNR 13.64} & \makecell{\includegraphics[width=0.125\textwidth]{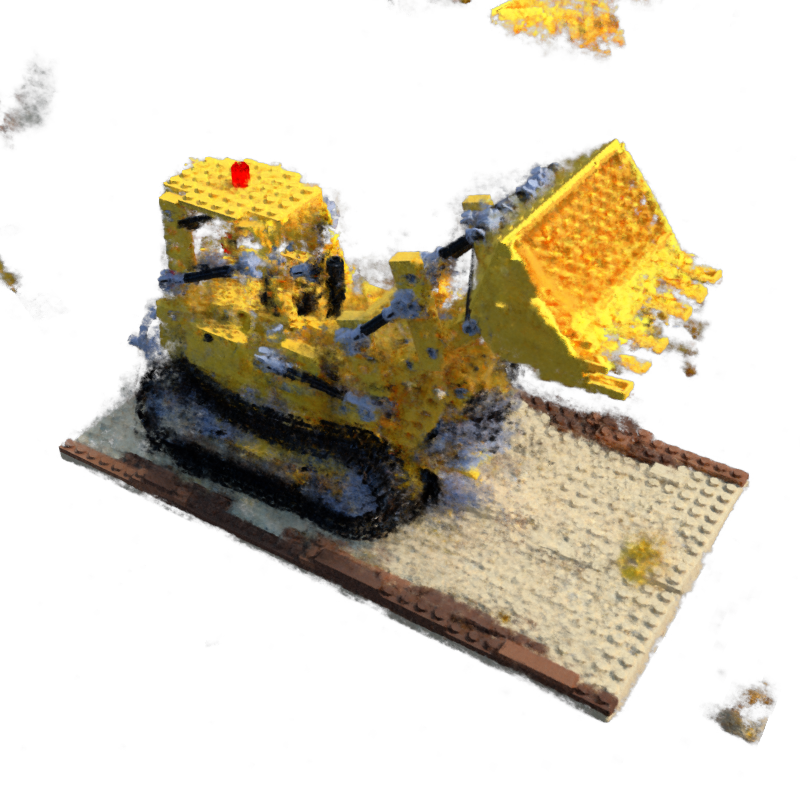} \\ PSNR 19.30} & \makecell{\includegraphics[width=0.125\textwidth]{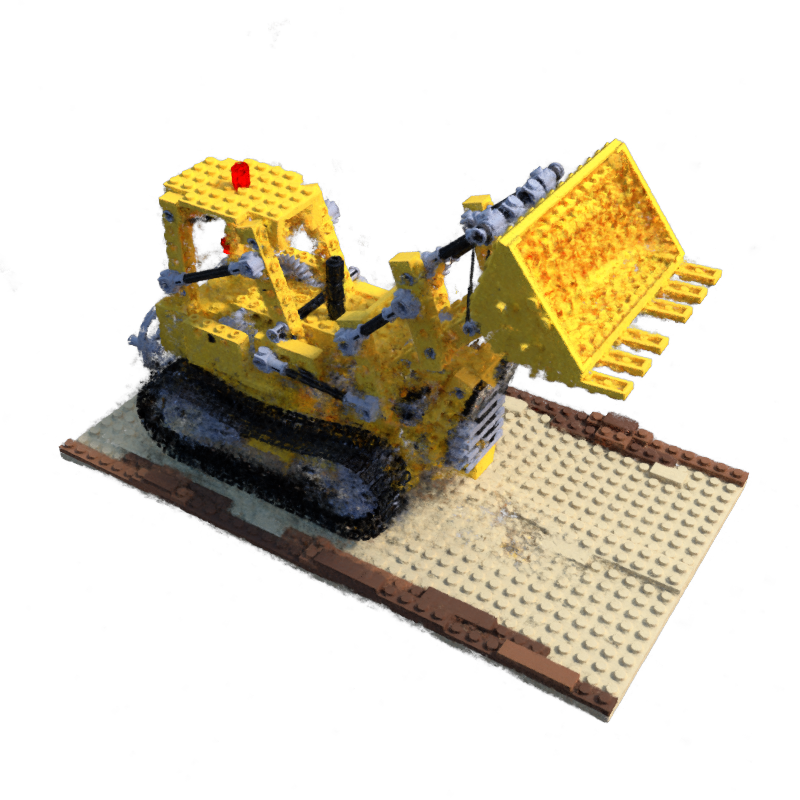} \\ PSNR 23.14} & \makecell{\includegraphics[width=0.125\textwidth]{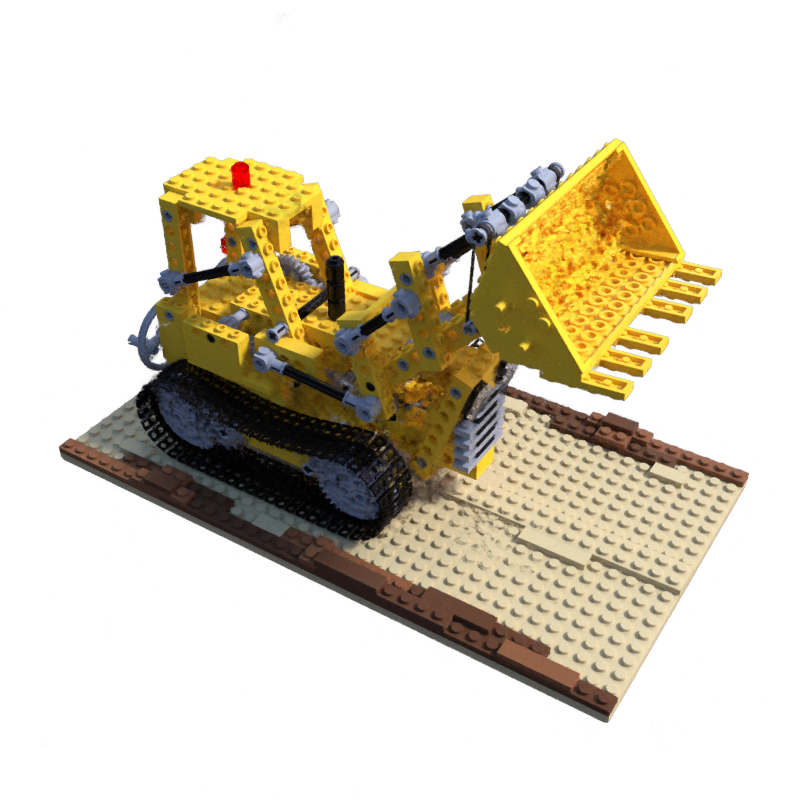} \\ PSNR 26.18} & \makecell{\includegraphics[width=0.125\textwidth]{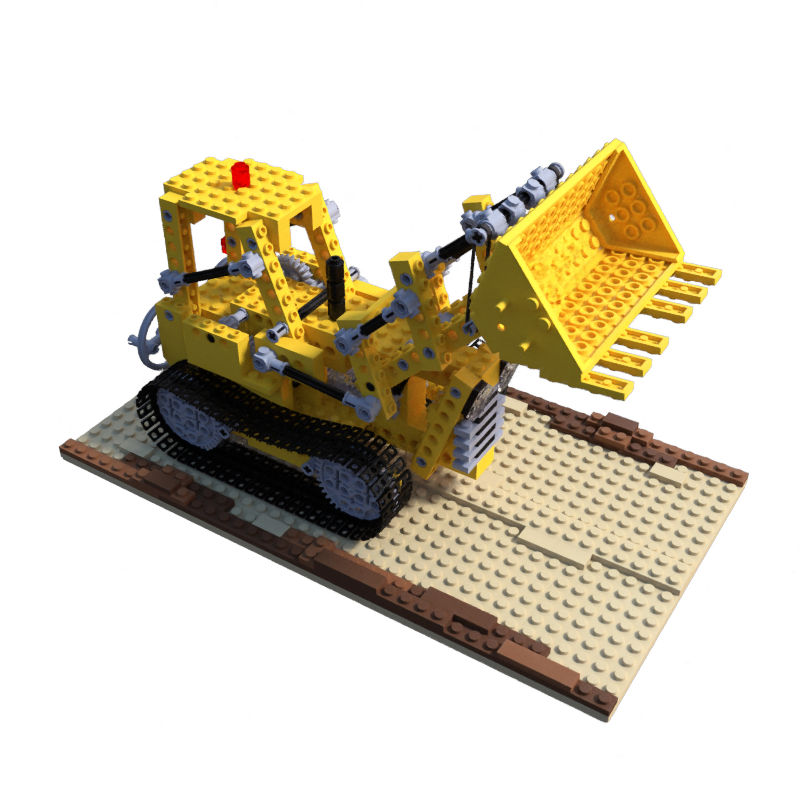} \\ PSNR 28.54} & \makecell{\includegraphics[width=0.125\textwidth]{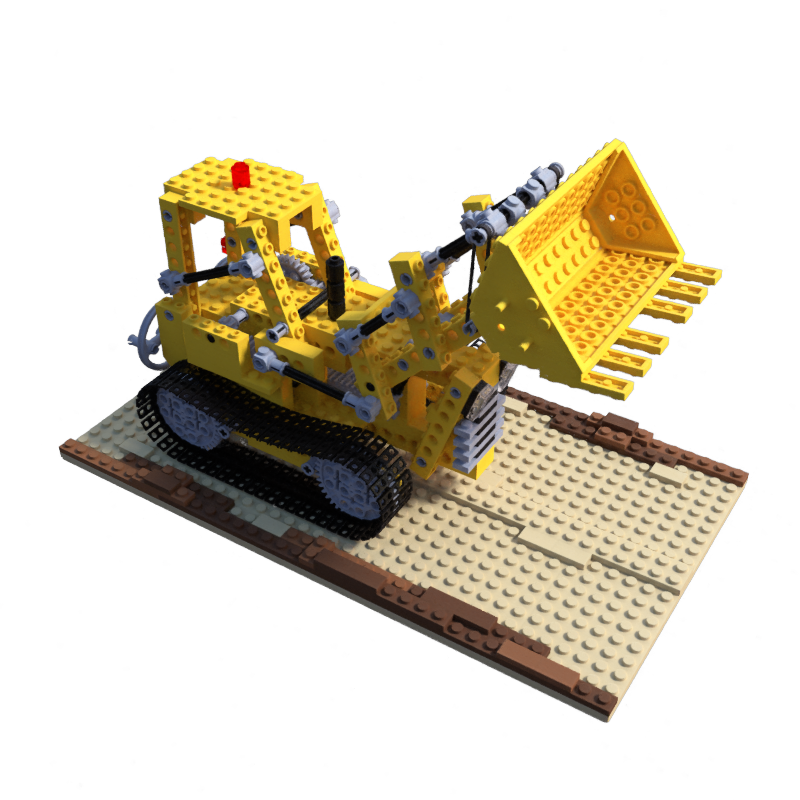} \\ PSNR 28.87} \\
        \midrule
        \makecell{\includegraphics[width=0.125\textwidth]{images/NeRF/instantNGP/drums_T1_56.png} \\ PSNR 10.37} & \makecell{\includegraphics[width=0.125\textwidth]{images/NeRF/instantNGP/drums_T2_56.png} \\ PSNR 11.72} & \makecell{\includegraphics[width=0.125\textwidth]{images/NeRF/instantNGP/drums_T3_56.png} \\ PSNR 12.82} & \makecell{\includegraphics[width=0.125\textwidth]{images/NeRF/instantNGP/drums_T5_56.png} \\ PSNR 15.58} & \makecell{\includegraphics[width=0.125\textwidth]{images/NeRF/instantNGP/drums_T10_56.png} \\ PSNR 19.71} & \makecell{\includegraphics[width=0.125\textwidth]{images/NeRF/instantNGP/drums_T20_56.png} \\ PSNR 21.28} & \makecell{\includegraphics[width=0.125\textwidth]{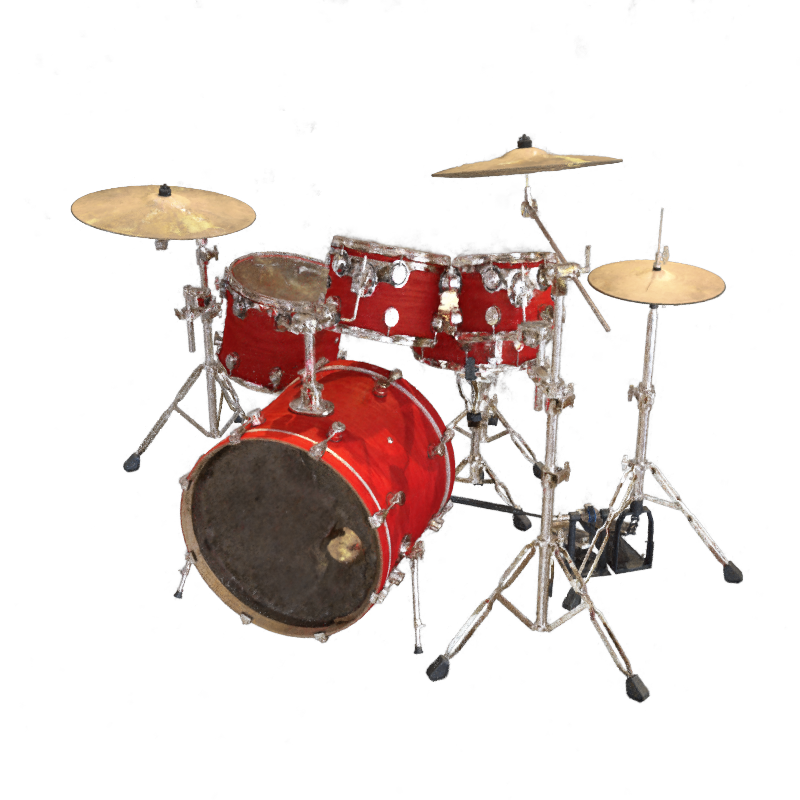} \\ PSNR 23.09} & \makecell{\includegraphics[width=0.125\textwidth]{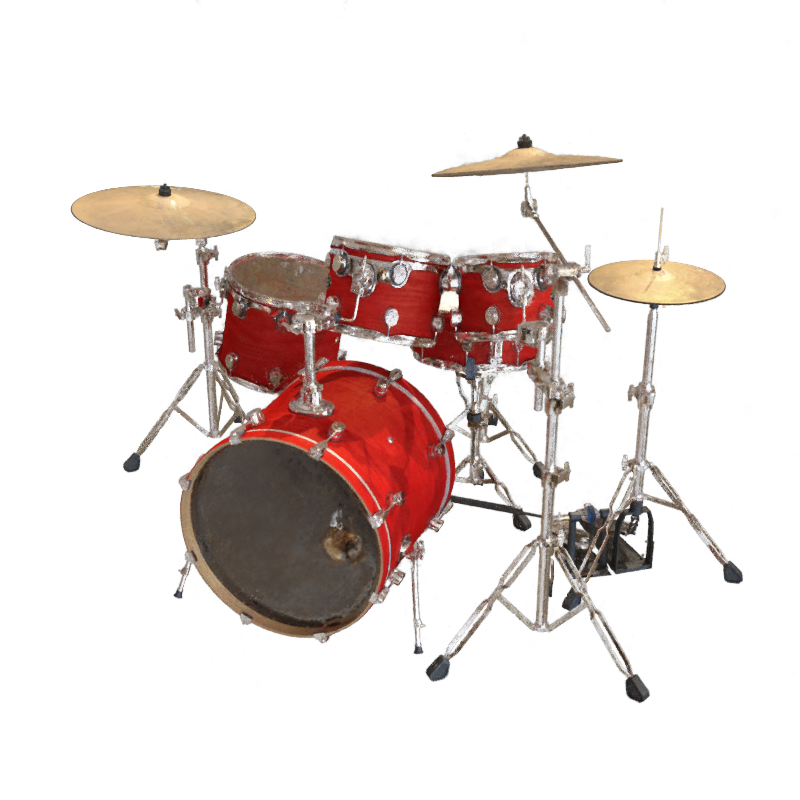} \\ PSNR 23.78} \\
        \midrule 
        \multicolumn{8}{l}{{\bf 3D Gaussian Splatting (Scene Specific Training)}} \\
        \makecell{\includegraphics[width=0.125\textwidth]{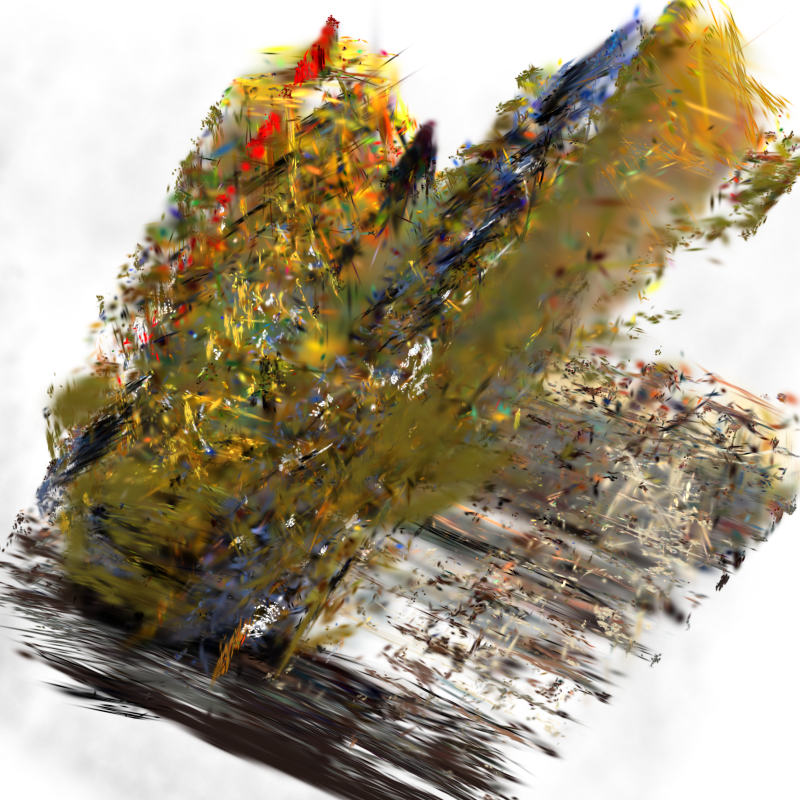} \\ PSNR 8.07} & \makecell{\includegraphics[width=0.125\textwidth]{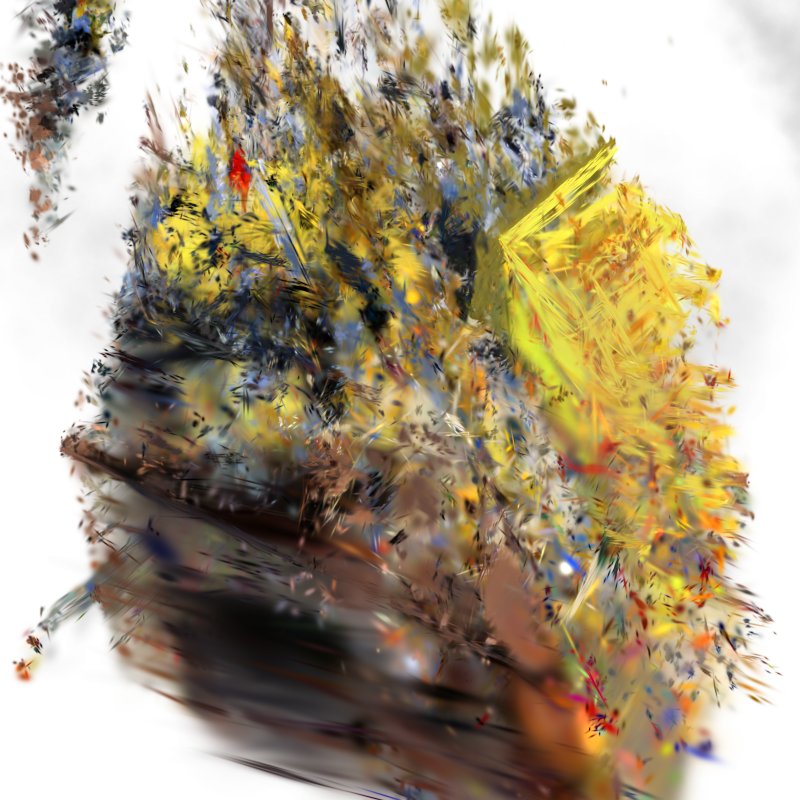} \\ PSNR 9.16} & \makecell{\includegraphics[width=0.125\textwidth]{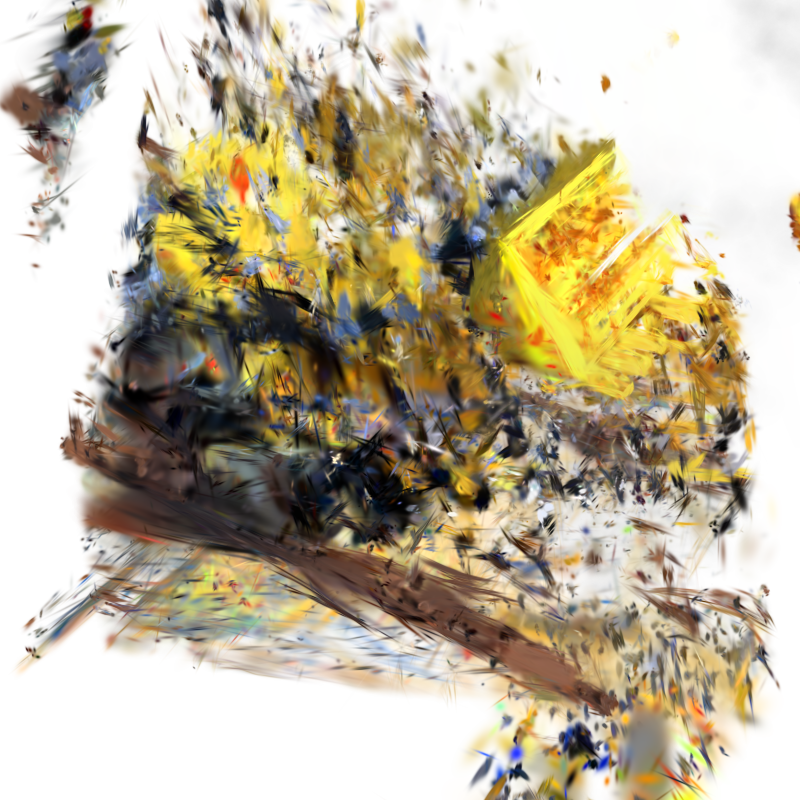} \\ PSNR 11.72} & \makecell{\includegraphics[width=0.125\textwidth]{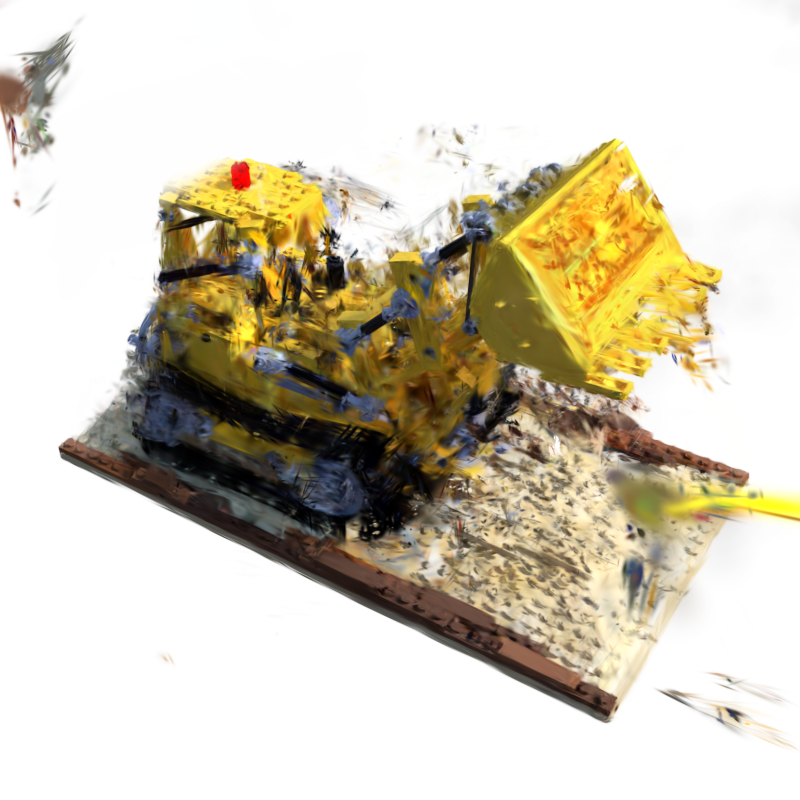} \\ PSNR 17.32} & \makecell{\includegraphics[width=0.125\textwidth]{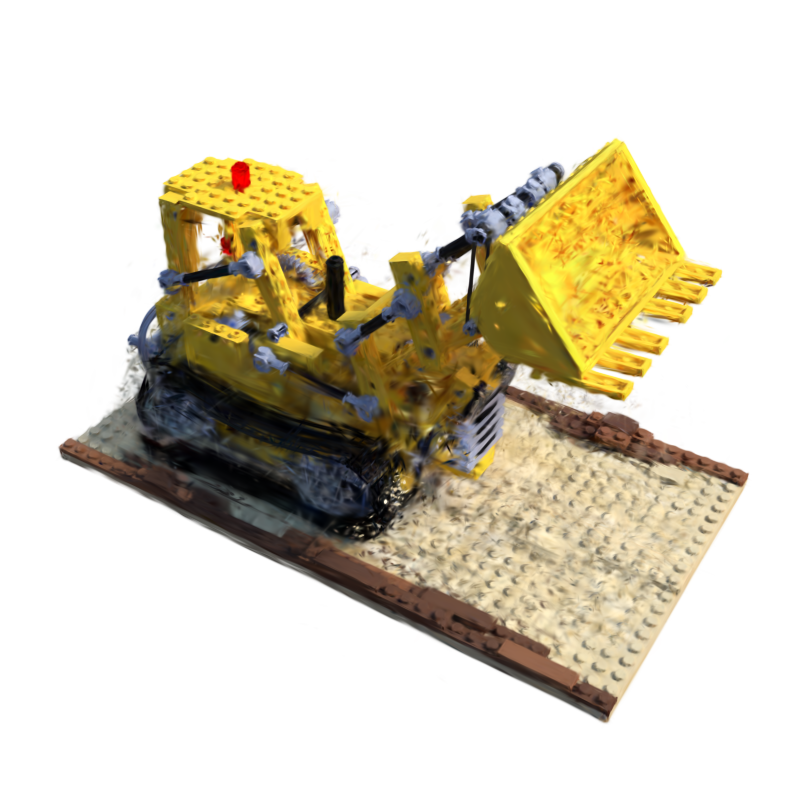} \\ PSNR 24.19} & \makecell{\includegraphics[width=0.125\textwidth]{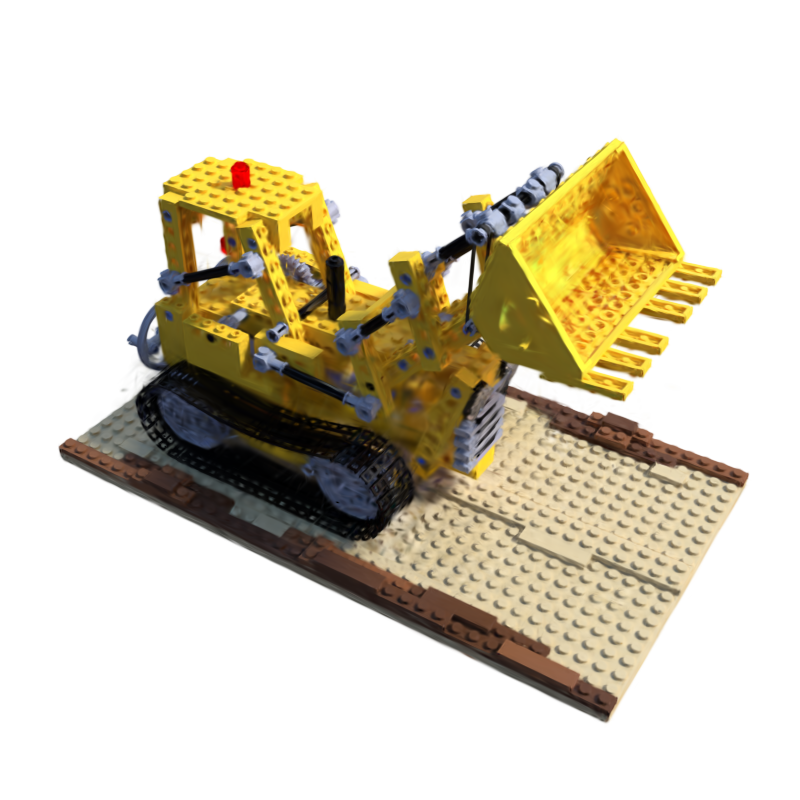} \\ PSNR 25.34} & \makecell{\includegraphics[width=0.125\textwidth]{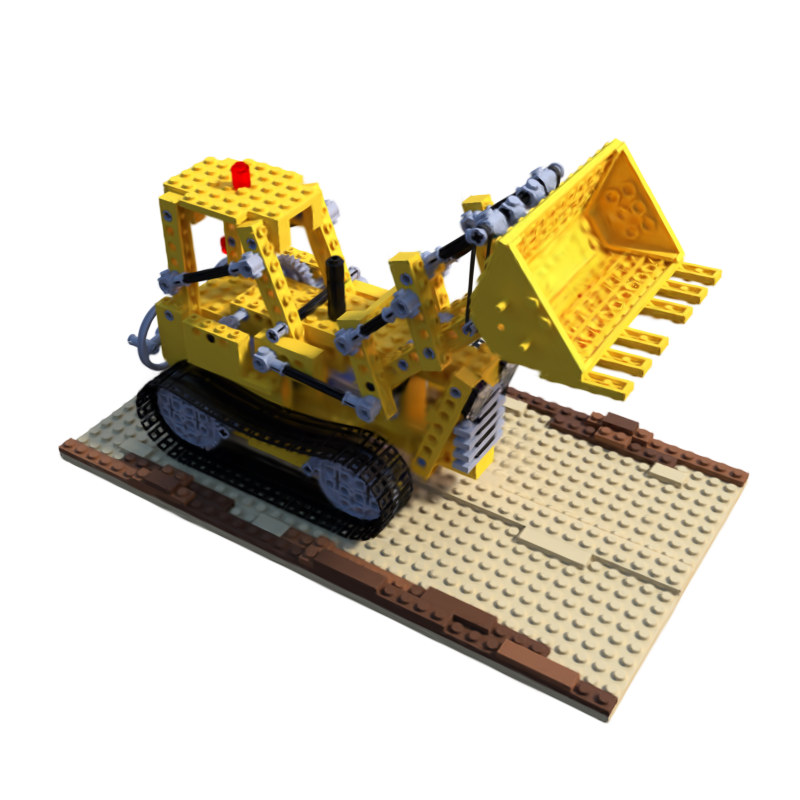} \\ PSNR 26.98} & \makecell{\includegraphics[width=0.125\textwidth]{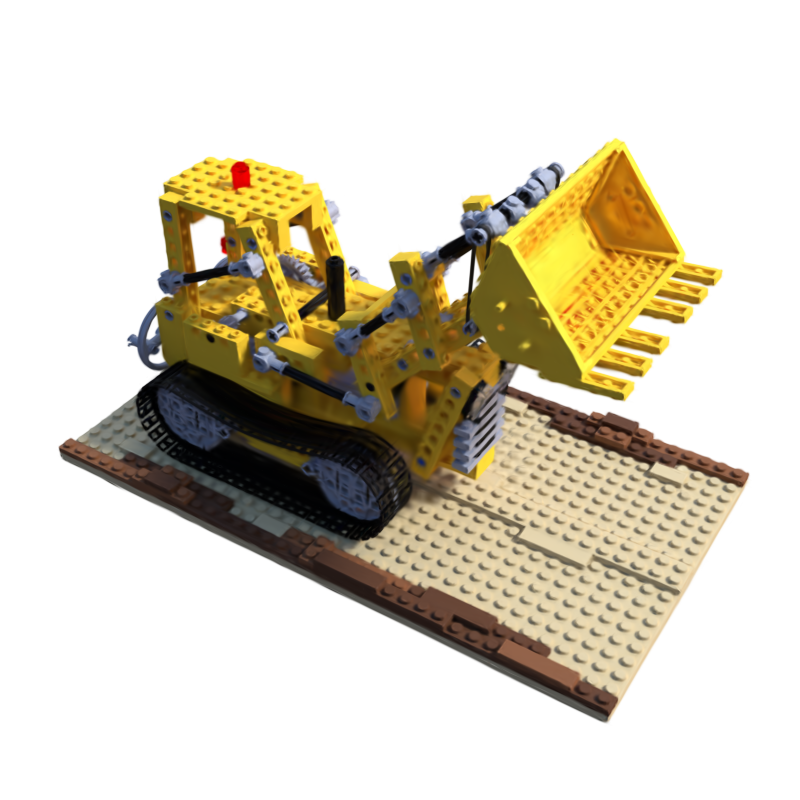} \\ PSNR 29.01} \\
        \midrule
        \makecell{\includegraphics[width=0.125\textwidth]{images/NeRF/3dgs/drums_T1_56.png} \\ PSNR 9.14} & \makecell{\includegraphics[width=0.125\textwidth]{images/NeRF/3dgs/drums_T2_56.png} \\ PSNR 10.63} & \makecell{\includegraphics[width=0.125\textwidth]{images/NeRF/3dgs/drums_T3_56.png} \\ PSNR 11.43} & \makecell{\includegraphics[width=0.125\textwidth]{images/NeRF/3dgs/drums_T5_56.png} \\ PSNR 14.81} & \makecell{\includegraphics[width=0.125\textwidth]{images/NeRF/3dgs/drums_T10_56.png} \\ PSNR 20.15} & \makecell{\includegraphics[width=0.125\textwidth]{images/NeRF/3dgs/drums_T20_56.png} \\ PSNR 22.88} & \makecell{\includegraphics[width=0.125\textwidth]{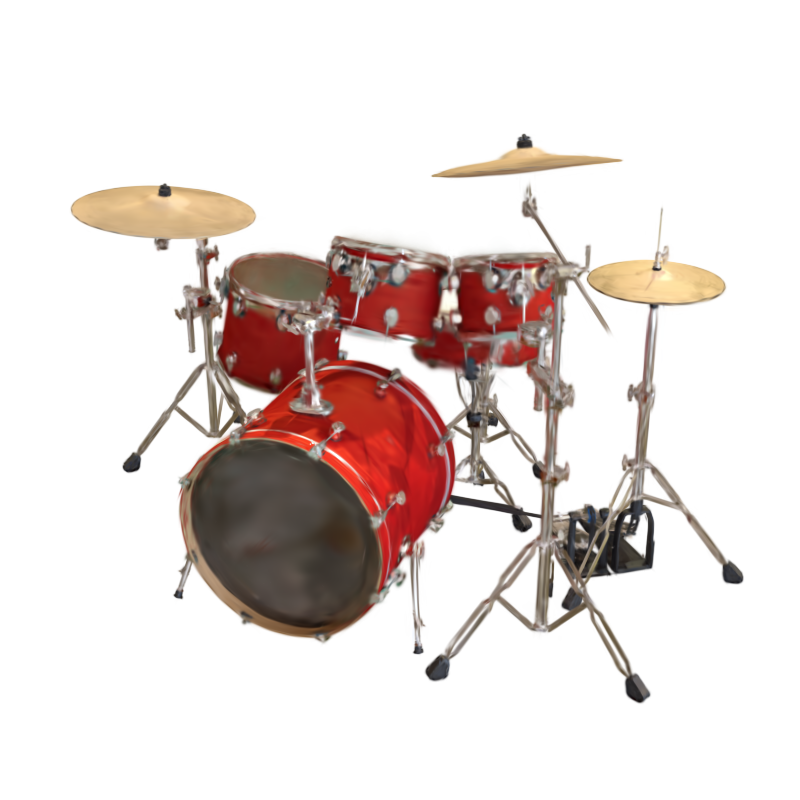} \\ PSNR 23.49} & \makecell{\includegraphics[width=0.125\textwidth]{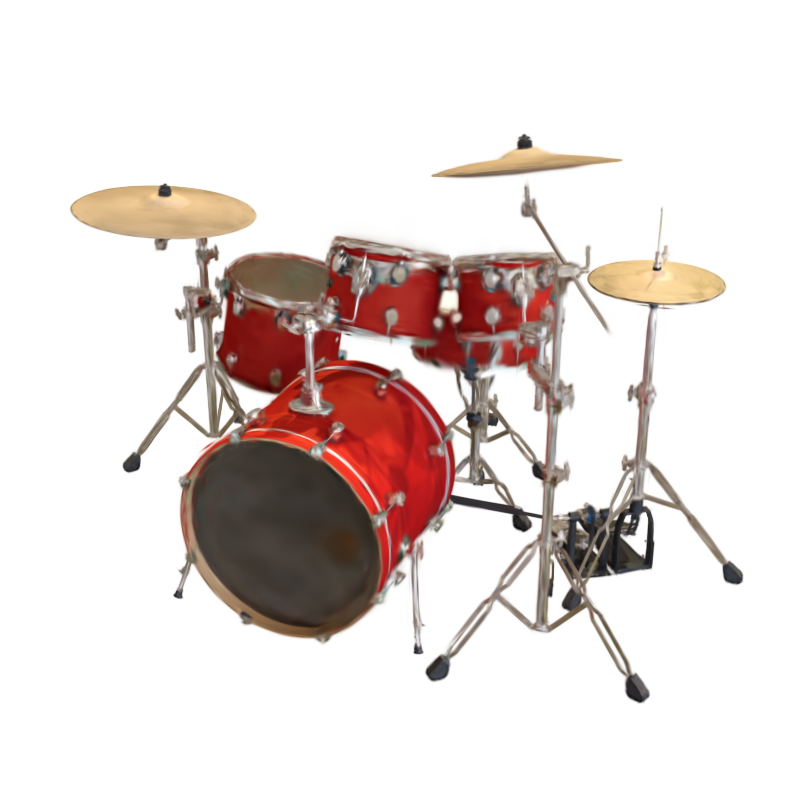} \\ PSNR 23.51} \\
        \midrule 
        \multicolumn{8}{l}{{\bf EscherNet (Zero Shot Inference)}} \\
        \makecell{\includegraphics[width=0.125\textwidth]{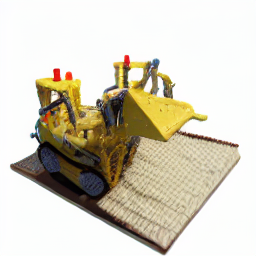} \\ PSNR 10.86} & \makecell{\includegraphics[width=0.125\textwidth]{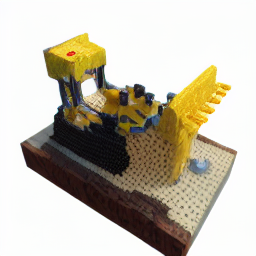} \\ PSNR 10.80} & \makecell{\includegraphics[width=0.125\textwidth]{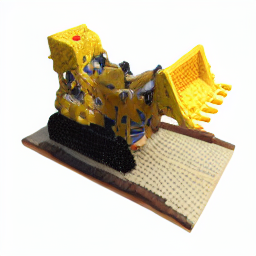} \\ PSNR 15.51} & \makecell{\includegraphics[width=0.125\textwidth]{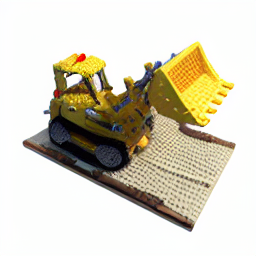} \\ PSNR 17.07} & \makecell{\includegraphics[width=0.125\textwidth]{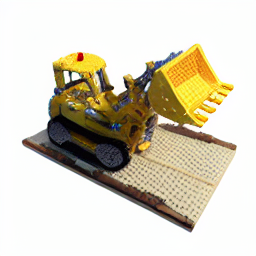} \\ PSNR 17.40} & \makecell{\includegraphics[width=0.125\textwidth]{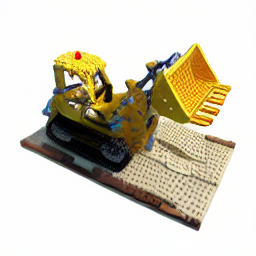} \\ PSNR 17.38} & \makecell{\includegraphics[width=0.125\textwidth]{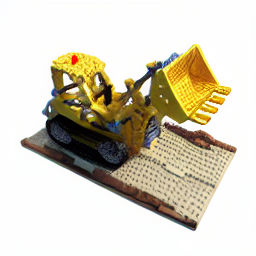} \\ PSNR 17.77} & \makecell{\includegraphics[width=0.125\textwidth]{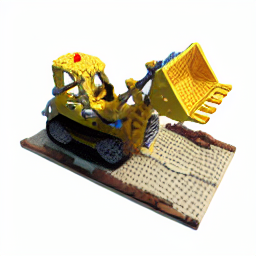} \\ PSNR 17.85} \\
        \midrule
        \makecell{\includegraphics[width=0.125\textwidth]{images/NeRF/eschernet/drums_T1_56.png} \\ PSNR 10.10} & \makecell{\includegraphics[width=0.125\textwidth]{images/NeRF/eschernet/drums_T2_56.png} \\ PSNR 13.25} & \makecell{\includegraphics[width=0.125\textwidth]{images/NeRF/eschernet/drums_T3_56.png} \\ PSNR 13.43} & \makecell{\includegraphics[width=0.125\textwidth]{images/NeRF/eschernet/drums_T5_56.png} \\ PSNR 14.33} & \makecell{\includegraphics[width=0.125\textwidth]{images/NeRF/eschernet/drums_T10_56.png} \\ PSNR 14.97} & \makecell{\includegraphics[width=0.125\textwidth]{images/NeRF/eschernet/drums_T20_56.png} \\ PSNR 15.65} & \makecell{\includegraphics[width=0.125\textwidth]{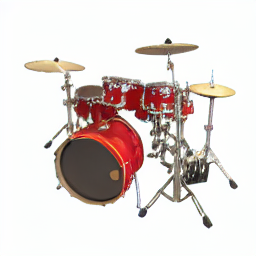} \\ PSNR 15.70} & \makecell{\includegraphics[width=0.125\textwidth]{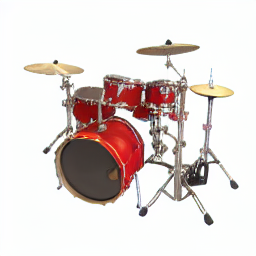} \\ PSNR 15.90} \\
        \bottomrule
       \end{tabular}
   \caption{Novel View Synthesis on NeRF Synthetic Dataset. We report the average PSNR per scene, conditioned on the respective number of reference views.}
   \label{tab:fig_nerf}
\end{table}

\newpage
\section{Additional Results on Text-to-3D}
\label{app:text23d}

We present additional visualisation on text-to-image-to-3D using EscherNet trained with 4 DoF CaPE.

\begin{table}[ht!]
   \centering
   \small
   \setlength{\tabcolsep}{0.2em}
       \begin{tabular}{C{0.29\linewidth}|C{0.7\linewidth}}
       \toprule
       \makecell{A robot made of vegetables. \vspace{0.2cm}\\ 
       \includegraphics[width=0.3\linewidth]{images/text23D/SDXL/veg_robot/veg_robot.png} 
       }
       &\includegraphics[width=\linewidth]{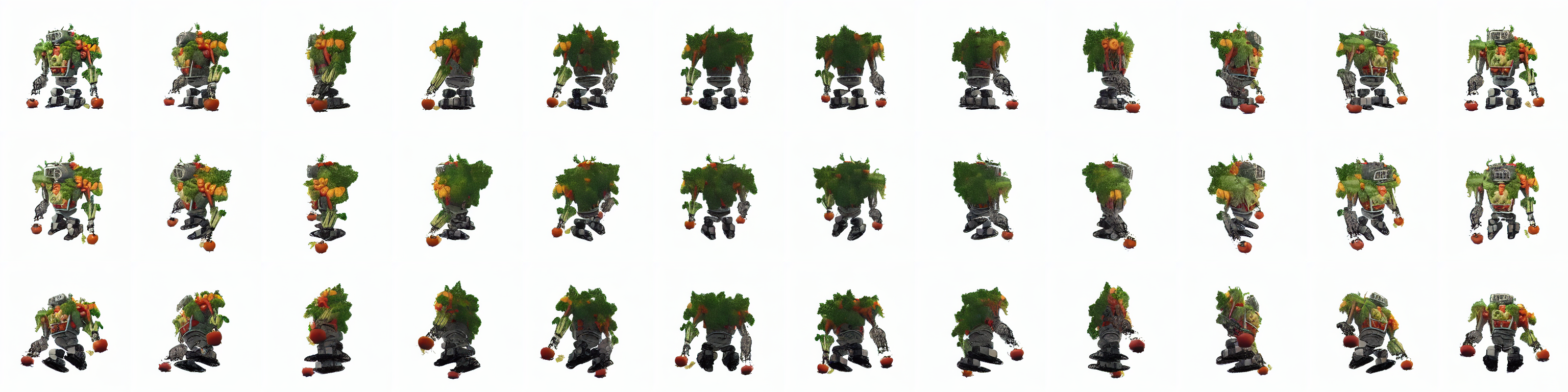}\\
       \midrule
        \makecell{A nurse corgi. \vspace{0.2cm} \\ \includegraphics[width=0.3\linewidth]{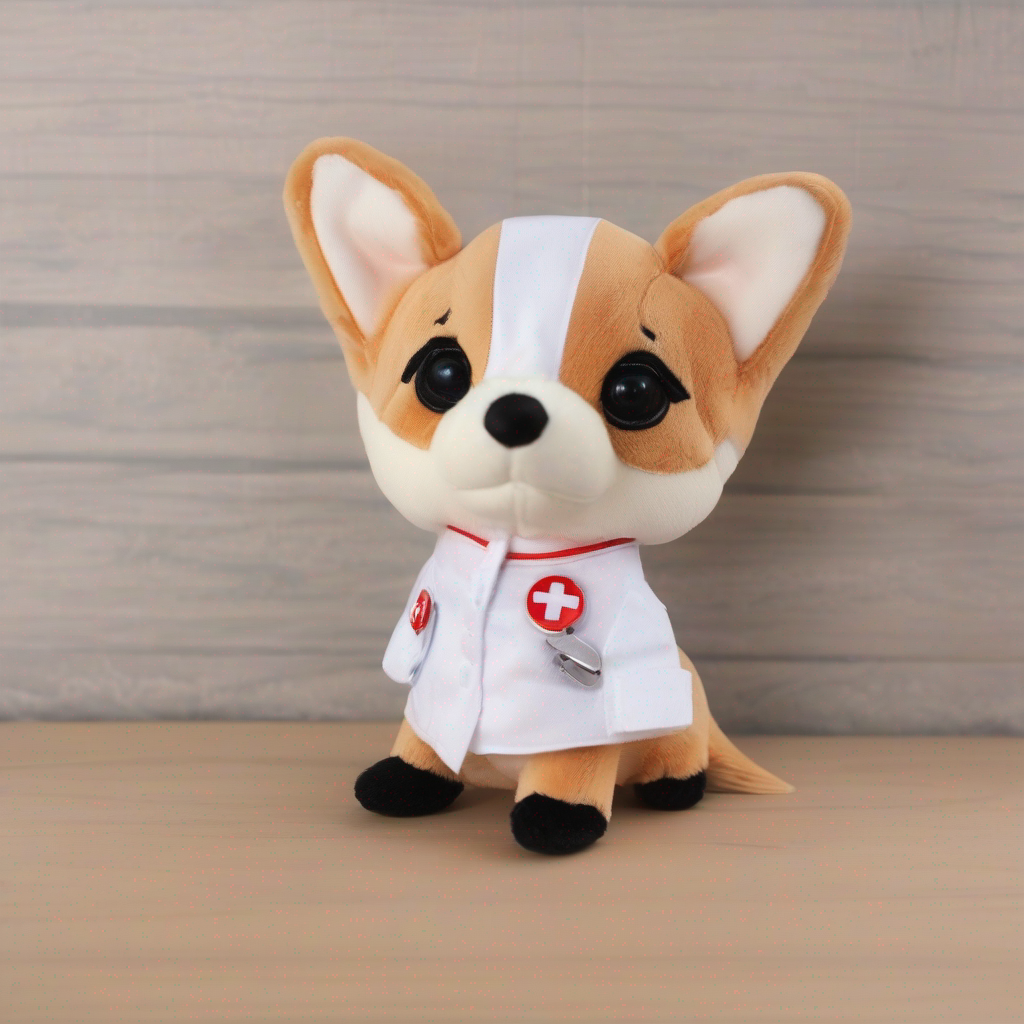}} & \includegraphics[width=\linewidth]{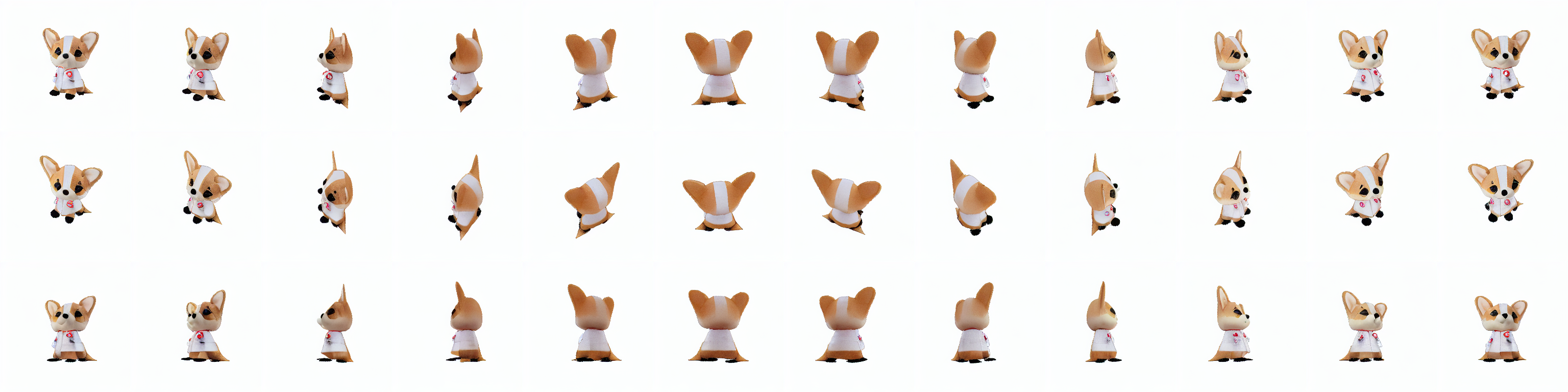}\\
       \midrule
       \makecell{A cute steampunk elephant. \vspace{0.2cm} \\ \includegraphics[width=0.3\linewidth]{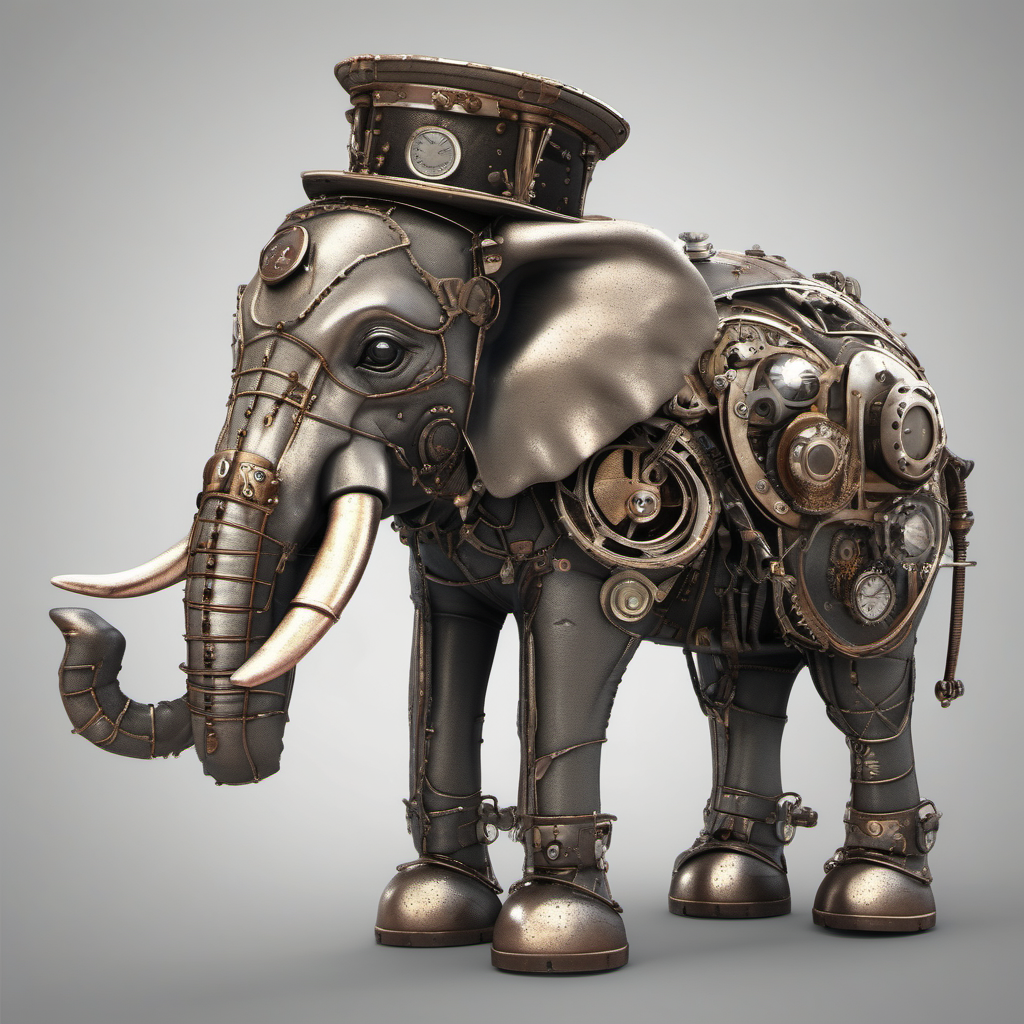}} &\includegraphics[width=\linewidth]{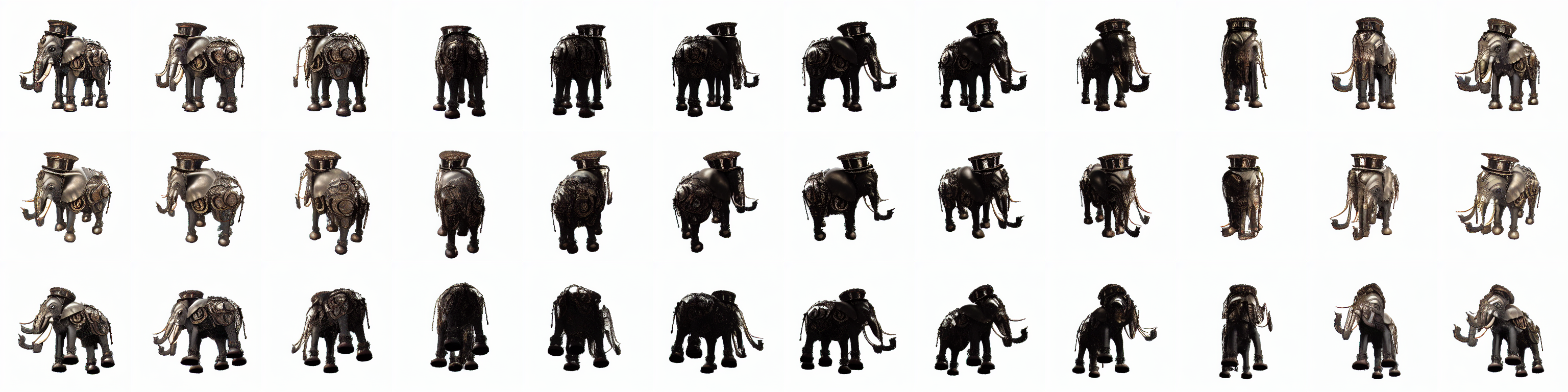}\\
       \midrule
       \makecell{A bull dog wearing a black pirate hat. \\ \includegraphics[width=\linewidth]{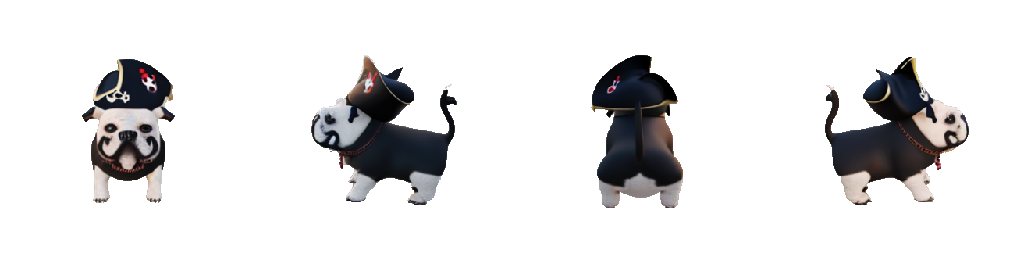}} &\includegraphics[width=\linewidth]{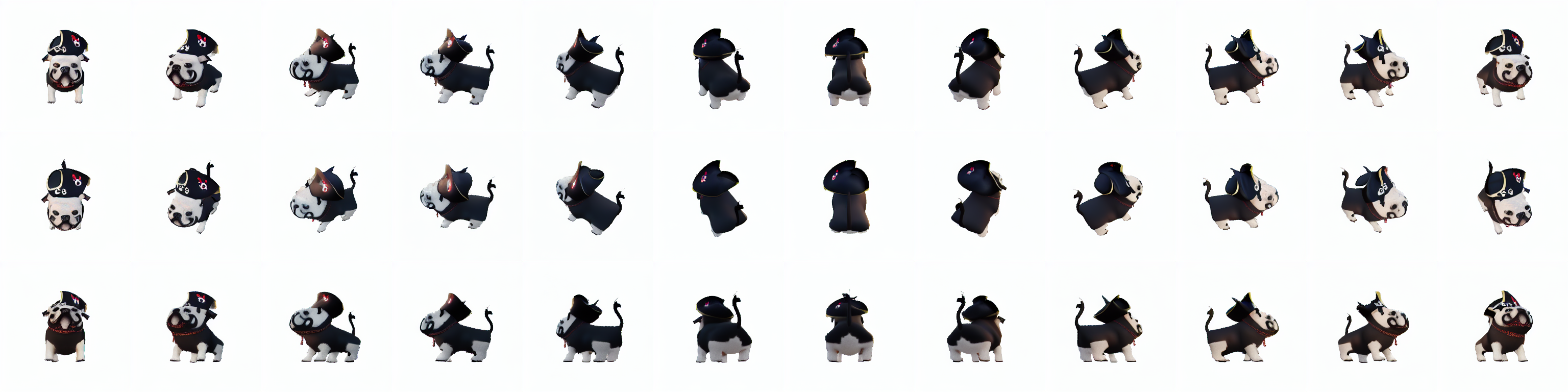}\\
       \midrule
       \makecell{An astronaut riding a horse. \\ \includegraphics[width=\linewidth]{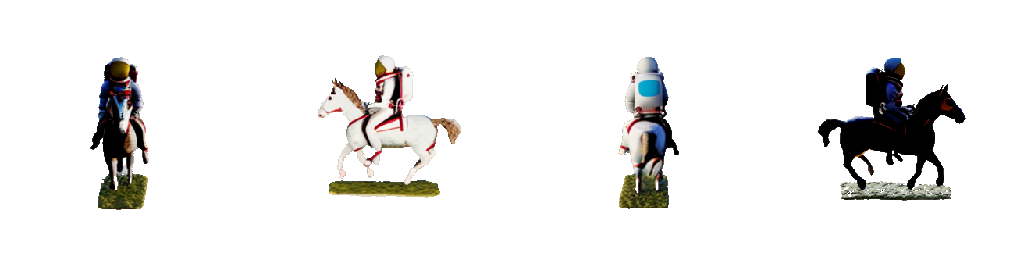}} &\includegraphics[width=\linewidth]{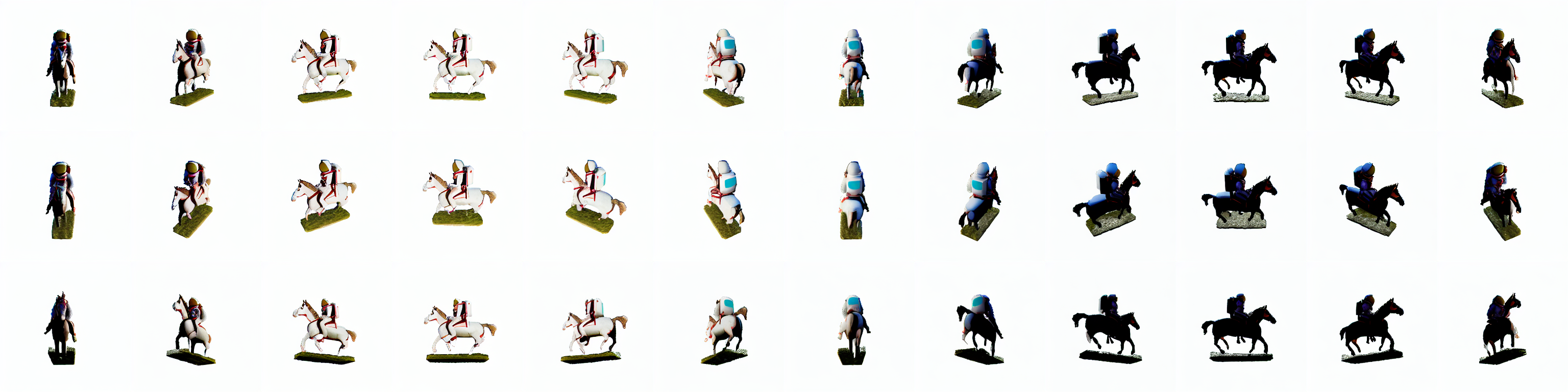}\\
       \midrule
       \makecell{Medieval House, grass, medieval,\\ medieval-decor, 3d asset. \\ \includegraphics[width=\linewidth]{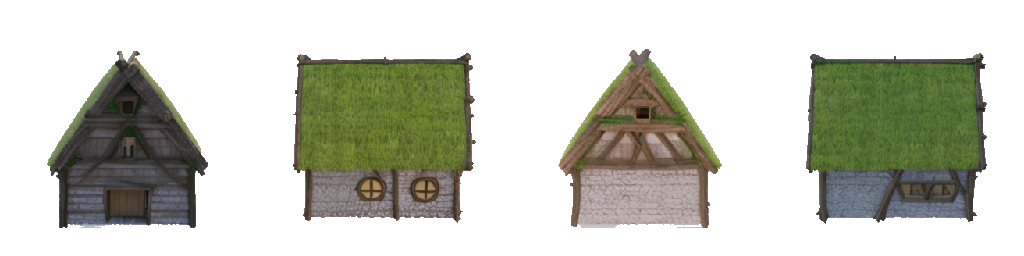}} &\includegraphics[width=\linewidth]{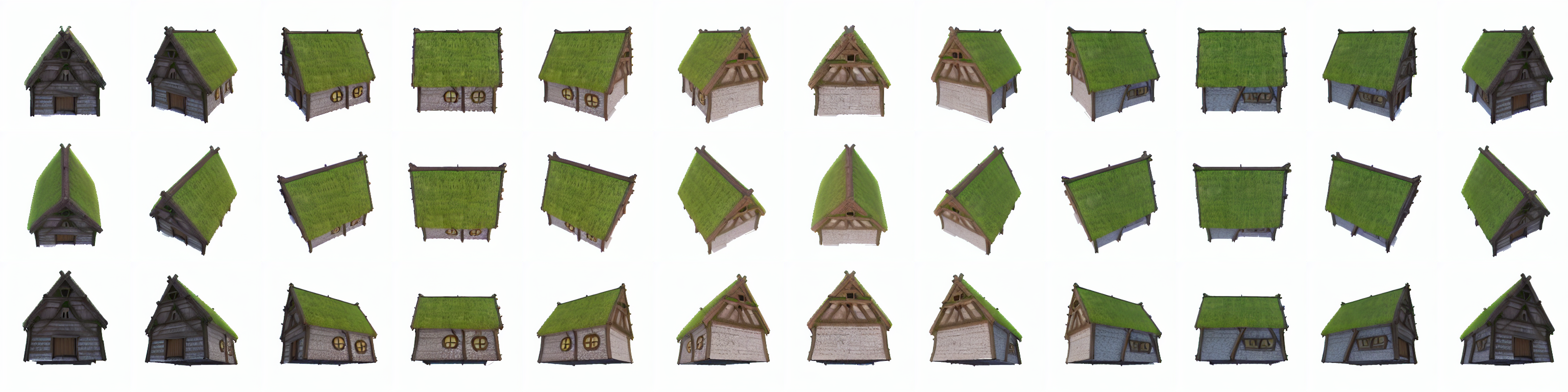}\\
        \bottomrule
       \end{tabular}
   \caption{Text-to-3D generation with SDXL (top 3) and MVDream (bottom 3).}
   \label{tab:fig_mvdream2}
\end{table}

\newpage
\section{Additional Discussions, Limitations and Future Work}
\label{app:discussions}

\paragraph{Direct v.s. Autoregressive Generation} 
EscherNet's flexibility in handling arbitrary numbers of reference and target views offers multiple choices for view synthesis. In our experiments, we employ the straightforward direct generation to jointly generate all target views. Additionally, an alternative approach is autoregressive generation, where target views are generated sequentially, similar to text generation with autoregressive language models.

For generating a large number of target views, autoregressive generation can be significantly faster than direct generation ({\it e.g.} more than $20\times$ faster for generating 200 views). This efficiency gain arises from converting a quadratic inference cost into a linear inference cost in each self-attention block. However, it's important to note that autoregressive generation may encounter a {\it content drifting problem} in our current design, where the generated quality gradually decreases as each newly generated view depends on previously non-perfect generated views. Autoregressive generation boasts many advantages in terms of inference efficiency and is well-suited for specific scenarios like SLAM (Simultaneous Localization and Mapping). As such, enhancing rendering quality in such a setting represents an essential avenue for future research.

\begin{figure}[ht!]
  \flushleft
  \captionsetup[subfigure]{font=footnotesize}
  \begin{subfigure}[b]{0.24\linewidth}
  \centering
    \resizebox{\linewidth}{!}{{\begin{tikzpicture}

\definecolor{color0}{rgb}{0.12156862745098,0.466666666666667,0.705882352941177}

\begin{axis}[
    every axis y label/.style={at={(current axis.north west)},above=2mm},
    width=6cm, height=6cm,
    legend cell align={left},
    legend style={fill opacity=1.0, draw opacity=1, text opacity=1, draw=none, column sep=2pt, very thick, anchor=south east, at={(0.98,0.02)}, legend columns=2, font=\footnotesize},
    tick align=outside,
    tick pos=left,
    xlabel={\# Target Views},
    xmin=0, xmax=31,
    xtick={1,3,5,10,15,20,25,30, 50, 100},
    ytick={13, 15, 17, 19, 21, 23, 25},
    ylabel={PSNR},
    ymin=12, ymax=26,
]
\addplot[line width=0.2pt, dashed, lightgray] coordinates {(-0.5,11)(31,11)};
\addplot[line width=0.2pt, dashed, lightgray] coordinates {(-0.5,13)(31,13)};
\addplot[line width=0.2pt, dashed, lightgray] coordinates {(-0.5,15)(31,15)};
\addplot[line width=0.2pt, dashed, lightgray] coordinates {(-0.5,17)(31,17)};
\addplot[line width=0.2pt, dashed, lightgray] coordinates {(-0.5,19)(31,19)};
\addplot[line width=0.2pt, dashed, lightgray] coordinates {(-0.5,21)(31,21)};
\addplot[line width=0.2pt, dashed, lightgray] coordinates {(-0.5,23)(31,23)};
\addplot[line width=0.2pt, dashed, lightgray] coordinates {(-0.5,25)(31,25)};

\addplot [draw=gray, fill=gray, mark=*, only marks, mark size=2pt]
table{%
x  y
1 14.35
3 16.32
5 17.25
10 17.49
20 17.42
25 16.71
30 16.54
50 16.73
100 17.12
};

\addplot [draw=orange, fill=orange, mark=*, only marks, mark size=2pt]
table{%
x  y
15 17.35
};
\node[above] at (axis cs:15, 17.35) {\footnotesize 17.35};
\end{axis}

\end{tikzpicture}}}
  \caption{1 Reference View}
  \end{subfigure}\hfill
  \begin{subfigure}[b]{0.24\linewidth}
  \centering
    \resizebox{\linewidth}{!}{{\begin{tikzpicture}

\definecolor{color0}{rgb}{0.12156862745098,0.466666666666667,0.705882352941177}

\begin{axis}[
    every axis y label/.style={at={(current axis.north west)},above=2mm},
    width=6cm, height=6cm,
    legend cell align={left},
    legend style={fill opacity=1.0, draw opacity=1, text opacity=1, draw=none, column sep=2pt, very thick, anchor=south east, at={(0.98,0.02)}, legend columns=2, font=\footnotesize},
    tick align=outside,
    tick pos=left,
    xlabel={\# Target Views},
    xmin=0, xmax=31,
    xtick={1,3,5,10,15,20,25,30},
    ytick={13, 15, 17, 19, 21, 23, 25},
    ylabel={PSNR},
    ymin=12, ymax=26,
]
\addplot[line width=0.2pt, dashed, lightgray] coordinates {(-0.5,11)(31,11)};
\addplot[line width=0.2pt, dashed, lightgray] coordinates {(-0.5,13)(31,13)};
\addplot[line width=0.2pt, dashed, lightgray] coordinates {(-0.5,15)(31,15)};
\addplot[line width=0.2pt, dashed, lightgray] coordinates {(-0.5,17)(31,17)};
\addplot[line width=0.2pt, dashed, lightgray] coordinates {(-0.5,19)(31,19)};
\addplot[line width=0.2pt, dashed, lightgray] coordinates {(-0.5,21)(31,21)};
\addplot[line width=0.2pt, dashed, lightgray] coordinates {(-0.5,23)(31,23)};
\addplot[line width=0.2pt, dashed, lightgray] coordinates {(-0.5,25)(31,25)};

\addplot [draw=gray, fill=gray, mark=*, only marks, mark size=2pt]
table{%
x  y
1 21.78
3 22.62
5 22.59
10 23.00
15 23.04
20 22.99
25 22.96
30 22.95
50 22.95
100 23.14
};

\addplot [draw=orange, fill=orange, mark=*, only marks, mark size=2pt]
table{%
x  y
15 23.04
};
\node[above] at (axis cs:15, 23.04) {\footnotesize 23.04};
\end{axis}

\end{tikzpicture}}}
  \caption{5 Reference Views}
  \end{subfigure}\hfill
    \begin{subfigure}[b]{0.24\linewidth}
  \centering
    \resizebox{\linewidth}{!}{{\begin{tikzpicture}

\definecolor{color0}{rgb}{0.12156862745098,0.466666666666667,0.705882352941177}

\begin{axis}[
    every axis y label/.style={at={(current axis.north west)},above=2mm},
    width=6cm, height=6cm,
    legend cell align={left},
    legend style={fill opacity=1.0, draw opacity=1, text opacity=1, draw=none, column sep=2pt, very thick, anchor=south east, at={(0.98,0.02)}, legend columns=2, font=\footnotesize},
    tick align=outside,
    tick pos=left,
    xlabel={\# Target Views},
    xmin=0, xmax=31,
    xtick={1,3,5,10,15,20,25,30},
    ytick={13, 15, 17, 19, 21, 23, 25},
    ylabel={PSNR},
    ymin=12, ymax=26,
]
\addplot[line width=0.2pt, dashed, lightgray] coordinates {(-0.5,11)(31,11)};
\addplot[line width=0.2pt, dashed, lightgray] coordinates {(-0.5,13)(31,13)};
\addplot[line width=0.2pt, dashed, lightgray] coordinates {(-0.5,15)(31,15)};
\addplot[line width=0.2pt, dashed, lightgray] coordinates {(-0.5,17)(31,17)};
\addplot[line width=0.2pt, dashed, lightgray] coordinates {(-0.5,19)(31,19)};
\addplot[line width=0.2pt, dashed, lightgray] coordinates {(-0.5,21)(31,21)};
\addplot[line width=0.2pt, dashed, lightgray] coordinates {(-0.5,23)(31,23)};
\addplot[line width=0.2pt, dashed, lightgray] coordinates {(-0.5,25)(31,25)};

\addplot [draw=gray, fill=gray, mark=*, only marks, mark size=2pt]
table{%
x  y
1 22.55
3 22.94
5 23.13
10 23.45
15 23.53
20 23.53
25 23.48
30 23.43
50 23.42
100 23.59
};

\addplot [draw=orange, fill=orange, mark=*, only marks, mark size=2pt]
table{%
x  y
15 23.53
};
\node[above] at (axis cs:15, 23.53) {\footnotesize 23.53};

\end{axis}

\end{tikzpicture}}}
  \caption{10 Reference Views}
  \end{subfigure}
  \hfill
    \begin{subfigure}[b]{0.24\linewidth}
  \centering
    \resizebox{\linewidth}{!}{{\begin{tikzpicture}

\definecolor{color0}{rgb}{0.12156862745098,0.466666666666667,0.705882352941177}

\begin{axis}[
    every axis y label/.style={at={(current axis.north west)},above=2mm},
    width=6cm, height=6cm,
    legend cell align={left},
    legend style={fill opacity=1.0, draw opacity=1, text opacity=1, draw=none, column sep=2pt, very thick, anchor=south east, at={(0.98,0.02)}, legend columns=2, font=\footnotesize},
    tick align=outside,
    tick pos=left,
    xlabel={\# Target Views},
    xmin=0, xmax=31,
    xtick={1,3,5,10,15,20,25,30},
    ytick={13, 15, 17, 19, 21, 23, 25},
    ylabel={PSNR},
    ymin=12, ymax=26,
]
\addplot[line width=0.2pt, dashed, lightgray] coordinates {(-0.5,11)(31,11)};
\addplot[line width=0.2pt, dashed, lightgray] coordinates {(-0.5,13)(31,13)};
\addplot[line width=0.2pt, dashed, lightgray] coordinates {(-0.5,15)(31,15)};
\addplot[line width=0.2pt, dashed, lightgray] coordinates {(-0.5,17)(31,17)};
\addplot[line width=0.2pt, dashed, lightgray] coordinates {(-0.5,19)(31,19)};
\addplot[line width=0.2pt, dashed, lightgray] coordinates {(-0.5,21)(31,21)};
\addplot[line width=0.2pt, dashed, lightgray] coordinates {(-0.5,23)(31,23)};
\addplot[line width=0.2pt, dashed, lightgray] coordinates {(-0.5,25)(31,25)};

\addplot [draw=gray, fill=gray, mark=*, only marks, mark size=2pt]
table{%
x  y
1 23.06
3 23.63
5 23.60
10 23.59
15 23.87
20 23.94
25 23.84
30 23.83
50 23.89
100 24.04
};

\addplot [draw=orange, fill=orange, mark=*, only marks, mark size=2pt]
table{%
x  y
15 23.87
};
\node[above] at (axis cs:15, 23.87) {\footnotesize 23.87};

\end{axis}

\end{tikzpicture}}}
  \caption{20 Reference Views}
  \end{subfigure}
  \caption{{\bf Novel view synthesis with a different number of reference and target views.} We present the averaged performance of EscherNet on {\it one} pre-selected target view across objects in the GSO dataset. We observe a clear improvement in view synthesis quality as the number of both reference and target views increases. In this scenario, the multiple target views are essentially multiple duplicates of the initially chosen single pre-selected view, a strategy we find effective in enhancing view synthesis quality.}
  \label{fig:analysis}
\end{figure}
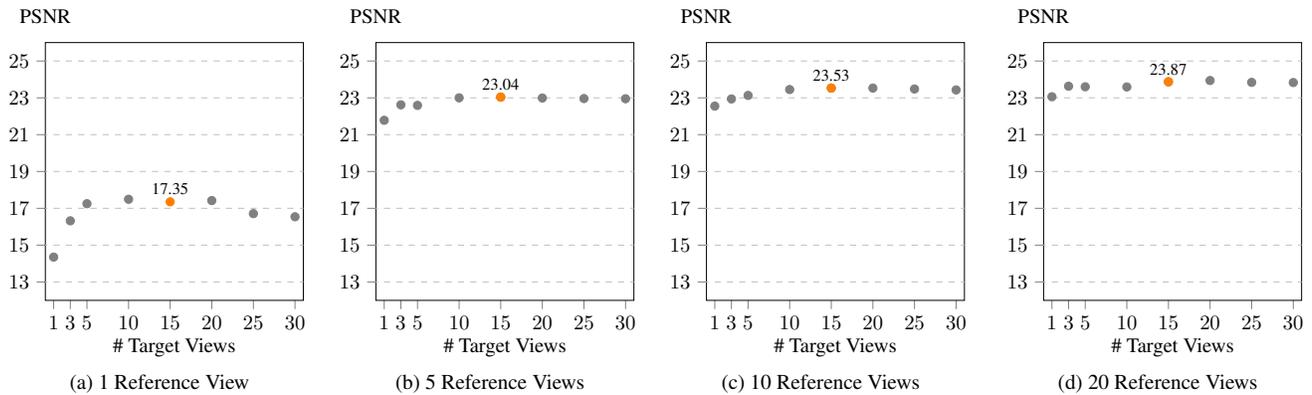

\paragraph{Stochasticity and Consistency in Multi-View Generation} 
We also observe that to enhance the target view synthesis quality, especially when conditioning on a limited number of reference views, introducing additional target views can be highly beneficial. These supplementary target views can either be randomly defined or duplicates with the identical target camera poses. Simultaneously generating multiple target views serves to implicitly reduce the inherent stochasticity in the diffusion process, resulting in improved generation quality and consistency. Through empirical investigations, we determine that the optimal configuration ensures a minimum of 15 target views, as highlighted in orange in Fig.~\ref{fig:analysis}. Beyond this threshold, any additional views yield marginal performance improvements.

\paragraph{Training Data Sampling Strategy}
We have explored various combinations of $N\in\{1,2,3,4,5\}$ reference views and $M\in\{1,2,3,4,5\}$ target views during EscherNet training. Empirically, a larger number of views demand more GPU memory and slow down training speed, while a smaller number of views may restrict the model's ability to learn multi-view correspondences. To balance training efficiency and performance, we set our training views to $N=3$ reference views and $M=3$ target views for each object, a configuration that has proven effective in practice. Additionally, we adopt a random sampling approach with replacement for these 6 views, introducing the possibility of repeated images in the training views. This sampling strategy has demonstrated a slight improvement in performance compared to sampling without replacement.

\paragraph{Scaling with Multi-view Video}
EscherNet's flexibility sets it apart from other multi-view diffusion models \cite{liu2023syncdreamer,long2023wonder3d} that require a set of fixed-view rendered images from 3D datasets for training. EscherNet can efficiently construct training samples using just a pair of posed images. While it can benefit from large-scale 3D datasets like~\cite{deitke2023objaverse, deitke2023objaversexl}, EscherNet's adaptability extends to a broader range of posed image sources, including those directly derived from videos. Scaling EscherNet to accommodate multiple data sources is an important direction for future research.

\end{document}